\documentclass{article}

\PassOptionsToPackage{numbers, compress}{natbib}

\usepackage[final]{neurips_2021}
\usepackage{algorithm}
\usepackage{algpseudocode}

\usepackage{amsmath,amsfonts,bm}

\def\eqref#1{equation~\ref{#1}}
\def\1{\bm{1}}

\DeclareMathAlphabet{\mathsfit}{\encodingdefault}{\sfdefault}{m}{sl}
\SetMathAlphabet{\mathsfit}{bold}{\encodingdefault}{\sfdefault}{bx}{n}

\usepackage[utf8]{inputenc} % allow utf-8 input
\usepackage[T1]{fontenc}    % use 8-bit T1 fonts
\usepackage{hyperref}       % hyperlinks
\usepackage{url}            % simple URL typesetting
\usepackage{booktabs}       % professional-quality tables
\usepackage{amsfonts}       % blackboard math symbols
\usepackage{amsmath,mathtools}
\usepackage{nicefrac}       % compact symbols for 1/2, etc.
\usepackage{microtype}      % microtypography
\usepackage{subcaption,siunitx,booktabs}
\usepackage{pifont}
\usepackage{bm}
\usepackage{lipsum}
\usepackage{xspace}
\usepackage{fancyvrb}
\usepackage{colortbl}
\usepackage[table]{xcolor}
\usepackage{wrapfig}
\usepackage{tablefootnote,footnotehyper}
\usepackage{mwe,pdflscape}
\usepackage{listings}% http://ctan.org/pkg/listings
\usepackage{graphicx, multirow, graphbox}
\usepackage{enumitem}
\usepackage{atbegshi,etoolbox}
\usepackage[toc,page,header]{appendix}
\usepackage{tocloft}
\usepackage{setspace}
\usepackage{minitoc}
\usepackage{stfloats}
\fnbelowfloat % puts footnotes below the bottom floats
\makeatletter
\def\blfootnote{\xdef\@thefnmark{}\@footnotetext}
\makeatother

\newcommand{\bx}{\mathbf{x}}

\graphicspath{ {figs/} }

\title{Parameter Prediction for Unseen Deep Architectures}

\author{\hspace{-5pt}Boris Knyazev$^{1,2}$\thanks{Part of the work was done while interning at Facebook AI Research.
	}%
	\quad
	Michal Drozdzal$^{4,\dagger}$
	\quad
	Graham W.~Taylor$^{1,2,3,\dagger}$%
	\quad
	Adriana Romero-Soriano$^{4,5,\dagger}$%
	\vspace{5pt} 
	\\
	$^1$ University of Guelph \:\:\:
	$^2$ Vector Institute for Artificial Intelligence \:\:\:
	\\
	$^3$ Canada CIFAR AI Chair \:\:\:
	$^4$ Facebook AI Research \:\:\:
	$^5$ McGill University \\
	$^\dagger$equal advising \\
}

\newcommand{\std}[1]{{\scriptsize{$\pm$#1}}}

\newcommand{\sem}[1]{{\scriptsize{$\pm$#1}}}

\newcommand\Tstrut{\rule{0pt}{2.6ex}}         % = `top' strut
\newcommand\Bstrut{\rule[-0.9ex]{0pt}{0pt}}   % = `bottom' strut

\newcommand{\f}{a}

\newcommand{\nets}{\mathcal{F}}
\newcommand{\domain}{\mathcal{D}}
\newcommand{\neigh}{\mathcal{N}}
\newcommand{\h}{\mathbf{h}}
\newcommand{\loss}{{\cal L}}
\newcommand{\w}{\mathbf{w}}

\newcommand{\params}{parameters\xspace}

\newcommand{\PLH}{{\mkern-0.1mu\times\mkern-0.1mu}}

\newcommand{\ghnbase}{\textsc{GHN-1}\xspace}
\newcommand{\ghnours}{\textsc{GHN-2}\xspace}
\newcommand{\dataset}{\textsc{DeepNets-1M}\xspace}

\newcommand{\iid}{\textsc{ID}\xspace}
\newcommand{\iidtrain}{\textsc{Train}\xspace}
\newcommand{\iidval}{\textsc{Val}\xspace}
\newcommand{\iidtest}{\textsc{Test}\xspace}
\newcommand{\ood}{\textsc{OOD}\xspace}
\newcommand{\wide}{\textsc{Wide}\xspace}
\newcommand{\deep}{\textsc{Deep}\xspace}
\newcommand{\dense}{\textsc{Dense}\xspace}
\newcommand{\bnfree}{\textsc{BN-free}\xspace}

\newcommand{\cmark}{\scriptsize\ding{51}}%
\newcommand{\xmark}{\scriptsize\ding{55}}%

\begin{document}
	
	\doparttoc % Tell to minitoc to generate a toc for the parts
	\faketableofcontents % Run a fake tableofcontents command for the partocs
	
	%\part{} % Start the document part
	\parttoc % Insert the document TOC
	
	\maketitle
	
	\begin{center}
		\vspace{-20pt}
		\textcolor{violet}{\url{https://github.com/facebookresearch/ppuda}}
		\vspace{0.08in}
	\end{center}
	
	\begin{abstract}
		Deep learning has been successful in automating the design of features in machine learning pipelines. However, the algorithms optimizing neural network parameters remain largely hand-designed and computationally inefficient. We study if we can use deep learning to directly predict these parameters by exploiting the past knowledge of training other networks.
		We introduce a large-scale dataset of diverse computational graphs of neural architectures -- \dataset -- and use it to explore parameter prediction on CIFAR-10 and ImageNet. By leveraging advances in graph neural networks, we propose a hypernetwork that can predict performant parameters in a \textit{single forward pass} taking a fraction of a second, even on a CPU.
		The proposed model achieves surprisingly good performance on \textit{unseen} and \textit{diverse} networks.
		For example, it is able to predict all 24 million parameters of a ResNet-50 achieving a 60\% accuracy on CIFAR-10. On ImageNet, top-5 accuracy of some of our networks approaches 50\%.
		Our task along with the model and results can potentially lead to a new, more computationally efficient paradigm of training networks.
		Our model also learns a strong representation of neural architectures enabling their analysis.\looseness-1
	\end{abstract}
	
	\setlength{\belowdisplayskip}{2pt}
	\setlength{\abovedisplayskip}{2pt}
	
	\section{Introduction}

	Consider the problem of training deep neural networks on large annotated datasets, 
	such as ImageNet~\cite{russakovsky2015imagenet}. This problem can be formalized as finding optimal parameters for a given neural network $a$, parameterized by $\w$, w.r.t. a loss function $\loss$ on the dataset $\domain=\{\bx_i, y_i\}_{i=1}^N$ of inputs $\bx_i$ and targets $y_i$:\looseness-1
	\begin{equation}
	\label{eq:optim1b}
	\underset{\w}{\text{arg\,min }}\sum\nolimits_{i=1}^N \loss(f(\bx_i; \f, \w), y_i),
	\end{equation}
	where $f(\bx_i; a, \w)$ represents a forward pass.
	Equation \ref{eq:optim1b} is usually minimized by iterative optimization algorithms -- e.g. SGD~\cite{ruder2016overview} and Adam~\cite{kingma2014adam}
	-- that converge to performant parameters $\w_p$ of the architecture $\f$. Despite the progress in improving the training speed and convergence~\cite{huang2016deep,brock2017freezeout,choi2019faster,ioffe2015batch}, obtaining $\w_p$ remains a bottleneck in large-scale machine learning pipelines. For example, training a ResNet-50~\cite{he2016deep} on ImageNet can take many GPU hours~\cite{nvidia}. With the ever growing size of networks~\cite{brown2020language} and necessity of training the networks repeatedly (e.g.~for hyperparameter or architecture search), the classical process of obtaining $\w_p$ is becoming computationally unsustainable~\cite{strubell2019energy,cai2019onceforall,thompson2020computational}. 
	
	\textbf{A new parameter prediction task.}
	When optimizing the parameters for a \textit{new} architecture $\f$, typical optimizers disregard past experience gained by optimizing other nets. However, leveraging this past experience can be the key to reduce the reliance on iterative optimization and, hence the high computational demands.
	To progress in that direction, we propose a new task where iterative optimization is replaced with a \textit{single forward pass} of a hypernetwork~\cite{ha2016hypernetworks} $H_\domain$.
	To tackle the task, $H_\domain$ is expected to leverage the knowledge of how to optimize \textit{other}	networks $\nets$.
	Formally, the task is to predict the parameters of an \textit{unseen} architecture $\f \notin \nets$ using $H_\domain$, parameterized by $\theta_p$: $\hat{\w}_p=H_{\domain}(\f; \theta_p)$.
	The task is constrained to a dataset $\domain$, so $\hat{\w}_p$ are the predicted parameters for which the test set performance of $f(\bx; \f, \hat{\w}_p)$ is similar to the one of $f(\bx; \f, \w_p)$.
	For example, we consider CIFAR-10~\cite{krizhevsky2009learning} and ImageNet image classification datasets $\domain$, where the test set performance is classification accuracy on test images.\looseness-1
	
	\textbf{Approaching our task.}
	A straightforward approach to expose $H_\domain$ to the knowledge of how to optimize other networks is to train it on a large training set of $\{a_i, \w_{p,i}\}$ pairs, however, that is prohibitive\footnote{Training a single network $a_i$ can take several GPU days and thousands of trained networks may be required.\looseness-1}. Instead, we follow the bi-level optimization paradigm common in meta-learning~\cite{hospedales2020meta,andrychowicz2016learning,ravi2016optimization}, but rather than iterating over $M$ tasks, we iterate over $M$ training architectures $\nets=\{a_i\}_{i=1}^M$:\looseness-1
	\begin{equation}
	\label{eq:solution}
	\underset{\theta}{\text{arg\,min }} \sum\nolimits_{j=1}^N \sum\nolimits_{i=1}^{M}\loss\Big(f\Big( \bx_j; a_i,  H_\domain(a_i;{\theta})\Big), y_j\Big).
	\end{equation}

\begin{figure}[tbhp]
		\centering
		\small 
		\setlength{\tabcolsep}{2pt}
		\vspace{-7pt}
		\begin{tabular}{cc}
			\multirow{2}{*}{\includegraphics[width=0.75\textwidth,align=c,trim={0 0 0 0}, clip]{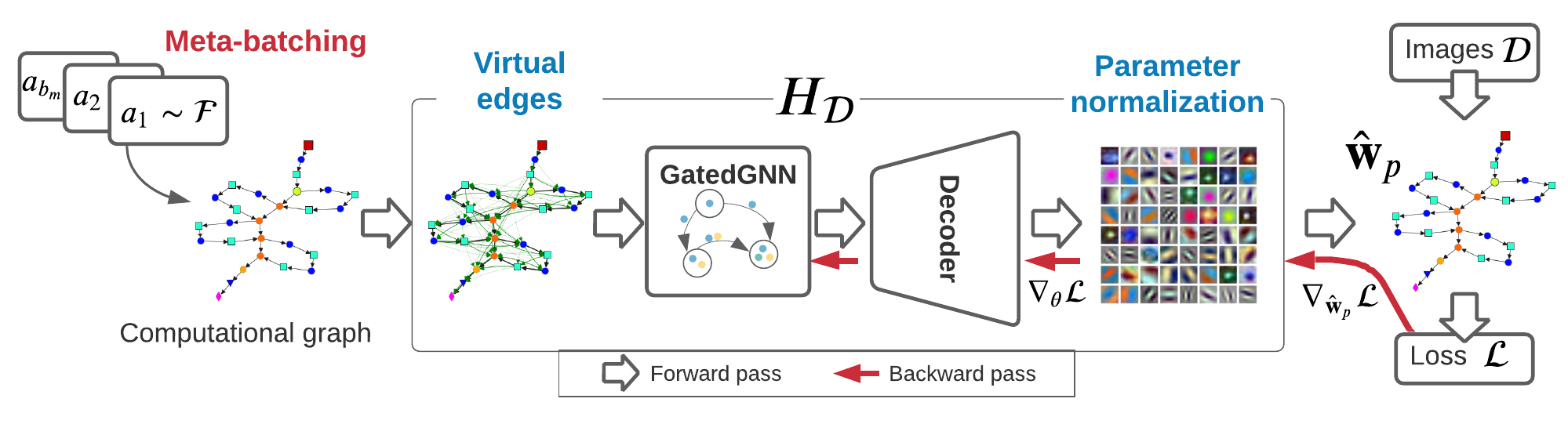}} & \parbox{3cm}{\vspace{10pt} \scriptsize \centering Example of evaluating on an unseen architecture $a \notin \nets$ (ResNet-50)}\\
			& \includegraphics[width=0.22\textwidth,align=c,trim={0 0 0 0}, clip]{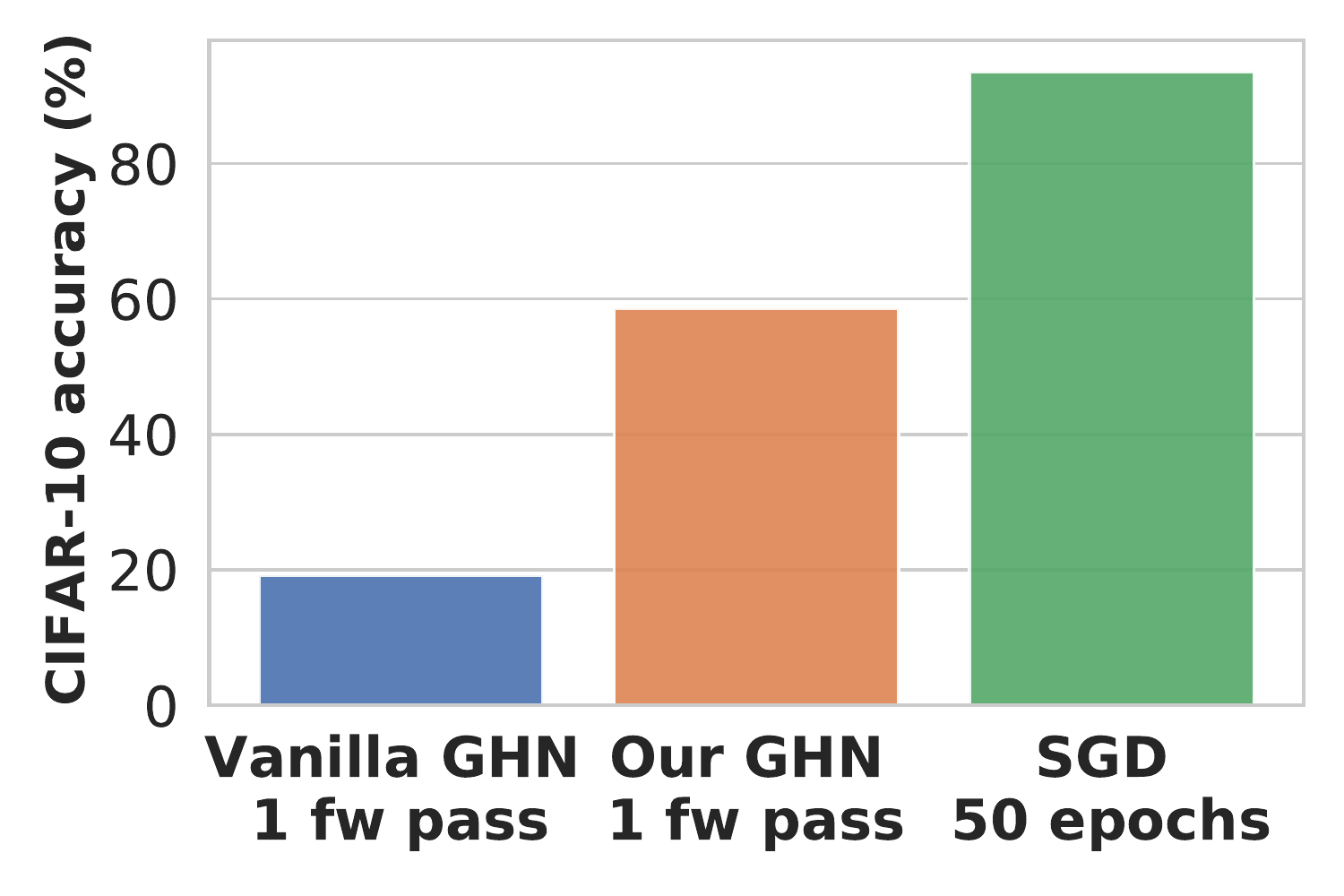} \vspace{5pt}\\
			(a) & (b) \\
		\end{tabular}
		\vspace{-3pt}
		\caption{\textbf{(a)} Overview of our GHN model (\S~\ref{sec:model}) trained by backpropagation through the predicted parameters ($\hat{\w}_p$) on a given image dataset and our \dataset dataset of architectures. Colored captions show our key improvements to vanilla GHNs (\S~\ref{sec:bg_ghn}). The red one is used only during training GHNs, while the blue ones are used both at training and testing time. The computational graph of $a_1$ is visualized as described in Table~\ref{tab:graphs}. \textbf{(b)} Comparing classification accuracies when all the parameters of a ResNet-50 are predicted by GHNs versus when its parameters are trained with SGD (see full results in \S~\ref{sec:exper}).
		}
		\label{fig:overview}
		\vspace{-15pt}
	\end{figure}

	By optimizing Equation~\ref{eq:solution}, the hypernetwork $H_{\domain}$ gradually gains knowledge of how to predict performant parameters for training architectures. It can then leverage this knowledge at test time -- when predicting parameters for \textit{unseen} architectures. 
	To approach the problem in Equation~\ref{eq:solution}, we need to design the network space $\nets$ and $H_{\domain}$.
	For $\nets$, we rely on the previous design spaces for neural architectures~\cite{liu2018darts} that we extend
	in two ways: the ability to sample distinct architectures and an expanded design space that includes diverse architectures, such as ResNets and Visual Transformers~\cite{dosovitskiy2020image}. 
	Such architectures can be fully described in the form of computational graphs (Fig.~\ref{fig:overview}). So, to design the hypernetwork $H_{\domain}$, we rely on recent advances in machine learning on graph-structured data~\cite{KipfW17, velickovic2018graph,dwivedi2020benchmarking,zhang2018graph}.
	In particular, we build on the Graph HyperNetworks method (GHNs)~\cite{zhang2018graph} that also optimizes Equation~\ref{eq:solution}. However, GHNs do not aim to predict large-scale performant parameters as we do in this work, which motivates us to improve on their approach.\looseness-1
	
	By designing our diverse space $\nets$ and improving on GHNs, we boost the accuracy achieved by the predicted parameters on \textit{unseen} architectures to 77\% (top-1) and 48\% (top-5) on CIFAR-10~\cite{krizhevsky2009learning} and ImageNet~\cite{russakovsky2015imagenet}, respectively. Surprisingly, our GHN shows good out-of-distribution generalization and predicts good \params for architectures that are much larger and deeper compared to the ones seen in training. For example, we can predict all 24 million parameters of ResNet-50 in less than a second either on a GPU or CPU achieving $\sim$60\% on CIFAR-10 without any gradient updates (Fig~\ref{fig:overview}, (b)).\looseness-1
	
	Overall, our framework and results pave the road toward a new and significantly more efficient paradigm for training networks.
	Our \textbf{contributions} are as follows: (\textbf{a}) we introduce the novel task of predicting performant \params for diverse feedforward neural networks with a single hypernetwork forward pass;
	(\textbf{b}) we introduce \dataset~-- a standardized benchmark with in-distribution and out-of-distribution architectures to track progress on the task (\S~\ref{sec:dataset}); (\textbf{c}) we define several baselines and propose a GHN model (\S~\ref{sec:model}) that performs surprisingly well on CIFAR-10 and ImageNet (\S~\ref{sec:our_task}); (\textbf{d}) we show that our model learns a strong representation of neural network architectures (\S~\ref{sec:prop_pred}), and our model is useful for initializing neural networks (\S~\ref{sec:finetune}).
	Our \dataset dataset, trained GHNs and code is available at \textcolor{violet}{\url{https://github.com/facebookresearch/ppuda}}.
	
	\section{Background\label{sec:problem}}
	
	We start by providing a brief background about the network design spaces leveraged in the creation of our \dataset dataset of neural architectures described in \S~\ref{sec:dataset}. We then cover elements of graph hypernetworks that we leverage when designing our specific GHN $H_\domain$ in \S~\ref{sec:model}.
	
	\subsection{Network Design Space of DARTS\label{sec:bg_darts}}

	DARTS~\cite{liu2018darts} is a differentiable NAS framework. For image classification tasks such as those considered in this work, its networks are defined by four types of building blocks: \emph{stems}, \emph{normal cells},  \emph{reduction cells}, and \emph{classification heads}. Stems are fixed blocks of convolutional operations that process input images. 
	The normal and reduction cells are the main blocks of architectures and are composed of: 
	3$\PLH$3 and 5$\PLH$5 separable convolutions,
	3$\PLH$3 and 5$\PLH$5 dilated separable convolutions, 
	3$\PLH$3 max pooling,  3$\PLH$3 average pooling, identity and zero (to indicate the absence of connectivity between two operations). Finally, the classification head defines the network output and is built with a global pooling followed by a single fully connected layer.
	
	Typically, DARTS networks have one stem block, 14-20 cells, and one classification head, altogether forming a deep computational graph. The reduction cells, placed only at 1/3 and 2/3 of the total depth, decrease the spatial resolution and increase the channel dimensionality by a factor of 2. Summation and concatenation are used to aggregate outputs from multiple operations within each cell. To make the channel dimensionalities match, 1$\PLH$1 convolutions are used as needed. All convolutional operations use the ReLU-Conv-Batch Norm (BN)~\cite{ioffe2015batch} order. Overall, DARTS enables defining strong architectures that combine many principles of manual~\cite{simonyan2014very,he2016deep,xie2017aggregated,huang2017densely} and automatic~\cite{zhang2018graph,zoph2016neural,zoph2018learning,liu2018progressive,real2019regularized,chen2019progressive,howard2019searching} design of neural architectures. While DARTS learns the optimal task-specific cells, the framework can be modified to permit sampling randomly-structured cells. We leverage this possibility for the \dataset construction in \S~\ref{sec:dataset}.
	Please see \S~\ref{apdx:darts_bg} for further details on DARTS.
	\looseness-1

	\subsection{Graph HyperNetwork: \ghnbase\label{sec:bg_ghn}}
	
	\paragraph{Representation of architectures.} GHNs~\cite{zhang2018graph} directly operate on the computational graph of a neural architecture $\f$. Specifically, $\f$ is a directed acyclic graph (DAG), where nodes $V =\{v_i\}_{i=1}^{|V|}$ are operations (e.g. convolutions, fully-connected layers, summations, etc.) and their connectivity is described by a binary adjacency matrix $\mathbf{A}\in \{0,1\}^{|V|\times |V|}$. Nodes are further characterized by a matrix of initial node features $\mathbf{H}^{0}=[\h_1^{0}, \h_2^{0}, ..., \h_{|V|}^{0}]$, where each $\h_v^{0}$ is a one-hot vector representing the operation performed by the node. 
	We also use such a one-hot representation for $\mathbf{H}^{0}$, but in addition encode the shape of parameters associated with nodes as described in detail in \S~\ref{apdx:ghn_1}.\looseness-1
	
	\paragraph{Design of the graph hypernetwork.} In~\cite{zhang2018graph}, the graph hypernetwork $H_\domain$ consists of three key modules. The first module takes the input node features $\mathbf{H}^{0}$ and transforms them into $d$-dimensional node features $\mathbf{H}^{1} \in \mathbb{R}^{|V| \times d}$ through an embedding layer. The second module takes $\mathbf{H}^{1}$ together with $\mathbf{A}$ and feeds them into a specific variant of the gated graph neural network (GatedGNN)~\cite{li2015gated}. In particular, their GatedGNN mimics the canonical order $\pi$ of node execution in the forward (fw) and backward (bw) passes through a computational graph.
	To do so, it sequentially traverses the graph and performs iterative message passing operations and node feature updates as follows: 
	\begin{align}
	\label{eq:ghn_prop}
	\forall t \in [1,...,T]:  \Big[ \forall \pi \in [\text{fw},\text{bw}]: \Big( \forall v \in \pi: \mathbf{m}^t_v = \sum\limits_{u \in \neigh_{v}^{\pi}} \text{MLP}(\mathbf{h}^{t}_u), \ \ \mathbf{\mathbf{h}}^{t}_v = \text{GRU}(\mathbf{h}_v^t, \mathbf{m}_v^t) \Big) \Big],
	\end{align}
	where $T$ denotes the total number of forward-backward passes; $\mathbf{h}_v^t$ corresponds to the features of node $v$ in the $t$-th graph traversal; $\text{MLP}(\cdot)$ is a multi-layer perceptron; and $\text{GRU}(\cdot)$ is the update function of the Gated Recurrent Unit~\cite{cho2014learning}. In the forward propagation ($\pi=\text{fw}$), $\neigh_v^{\pi}$ corresponds to the incoming neighbors of the node defined by $\mathbf{A}$, then in the backward propagation ($\pi=\text{bw}$) it similarly corresponds to the outgoing neighbors of the node. The last module uses the GatedGNN output hidden states $\mathbf{h}_v^T$ to condition a decoder that produces the parameters $\hat{\w}_{p}^{v}$ (e.g. convolutional weights) associated with each node. 
	In practice, to handle different parameter dimensionalities per operation type, the output of the hypernetwork is reshaped and sliced according to the shape of parameters in each node. We refer to the model described above as \ghnbase~(Fig.~\ref{fig:overview}). Further subtleties of implementing this model in the context of our task are discussed in \S~\ref{apdx:ghn_1}.

	\vspace{-5pt}
	\section{\dataset\label{sec:dataset}}
	\vspace{-5pt}
	
	The network design space of DARTS is limited by the number of unique operations that compose cells, and the low variety of stems and classification heads. Thus, many architectures are not realizable within this design space, including: VGG~\cite{simonyan2014very}, ResNets~\cite{he2016deep}, MobileNet~\cite{howard2019searching} or more recent ones such as Visual Transformer (ViT)~\cite{dosovitskiy2020image} and Normalization-free networks~\cite{brock2021characterizing,brock2021high}.
	Furthermore, DARTS does not define a procedure to sample random architectures.	
	By addressing these two limitations we aim to expose our hypernetwork to diverse training architectures and permit its evaluation on common architectures, such as ResNet-50. We hypothesize that increased training diversity can improve hypernetworks' generalization to unseen architectures making it more competitive to iterative optimizers.\looseness-1
	
	\textbf{Extending the network design space.} We extend the set of possible operations with non-separable 2D convolutions\footnote{Non-separable convolutions have weights of e.g. shape 3$\PLH$3$\PLH$512$\PLH$512 as in ResNet-50. NAS works, such as DARTS and GHN, avoid such convolutions, since the separable ones~\cite{sifre2014rigid} are more efficient. Non-separable convolutions are nevertheless common in practice and can often boost the downstream performance.}, Squeeze\&Excite\footnote{The Squeeze\&Excite operation is common in many efficient networks~\cite{howard2017mobilenets,cai2019onceforall}.} (SE)~\cite{hu2018squeeze} and Transformer-based operations~\cite{vaswani2017attention,dosovitskiy2020image}: multihead self-attention (MSA), positional encoding and layer norm (LN)~\cite{ba2016layer}. 
	Each node (operation) in our graphs has two attributes: \emph{primitive type} (e.g. convolution) and \emph{shape} (e.g. 3$\PLH$3$\PLH$512$\PLH$512). Overall, our extended set consists of 15 primitive types (Table~\ref{tab:graphs}).
	We also extend the diversity of the generated architectures by introducing VGG-style classification heads and ViT stems. 
	Finally, to further increase architectural diversity, we allow the operations to not include batch norm (BN)~\cite{ioffe2015batch} and permit networks without channel width expansion (e.g. as in~\cite{dosovitskiy2020image}).\looseness-1
	
	\textbf{Architecture generation process.} We generate different subsets of architectures (see the description of each subset in the next two paragraphs and in Table~\ref{tab:graphs}). For each subset depending on its purpose, we predefine a range of possible model depths (number of cells), widths and number of nodes per cell. Then, we sample a stem, a normal and reduction cell and a classification head. The internal structure of the normal and reduction cells is defined by uniformly sampling from all available operations. 
	Due to a diverse design space it is extremely unlikely to sample the same architecture multiple times, but we ran a sanity check using the Hungarian algorithm~\cite{kuhn1955hungarian} to confirm that (see Figure~\ref{fig:vis_stats} in \S~\ref{apdx:stats} for details).\looseness-1
	
	\begin{table}[t!]
		\centering
		\vspace{-10pt}
		\caption{\small Examples of computational graphs (visualized using NetworkX~\cite{hagberg2008exploring}) in each split and their key statistics, to which we add the average degree and average shortest path length often used to measure local and global graph properties respectively~\cite{barrat2004architecture,you2020graph}. In the visualized graphs, a node is one of the 15 primitives coded with markers shown at the bottom, where they are sorted by the frequency in the training set. For visualization purposes, a blue triangle marker differentiates a 1$\PLH$1 convolution (equivalent to a fully-connected layer over channels) from other convolutions, but its primitive type is still just convolution. \textsuperscript{*}Computed based on CIFAR-10.\looseness-1			
		}
		\vspace{2pt}
		\label{tab:graphs}
		\scriptsize
		\newcommand{\width}{0.135\textwidth}
		\setlength{\tabcolsep}{0pt}
		\begin{tabular}{p{1.4cm}ccp{0.2cm}ccccc}
			& \multicolumn{2}{c}{{\small \textbf{\textsc{In-Distribution}}}} & &
			\multicolumn{5}{c}{{\small \textbf{\textsc{Out-of-Distribution}}}}
			\Bstrut\\
			\cline{2-3}\cline{5-9} \\[-2ex]
			& \multicolumn{2}{c}{{\includegraphics[width=\width,align=c,trim={3cm 3cm 3cm 3cm},clip]{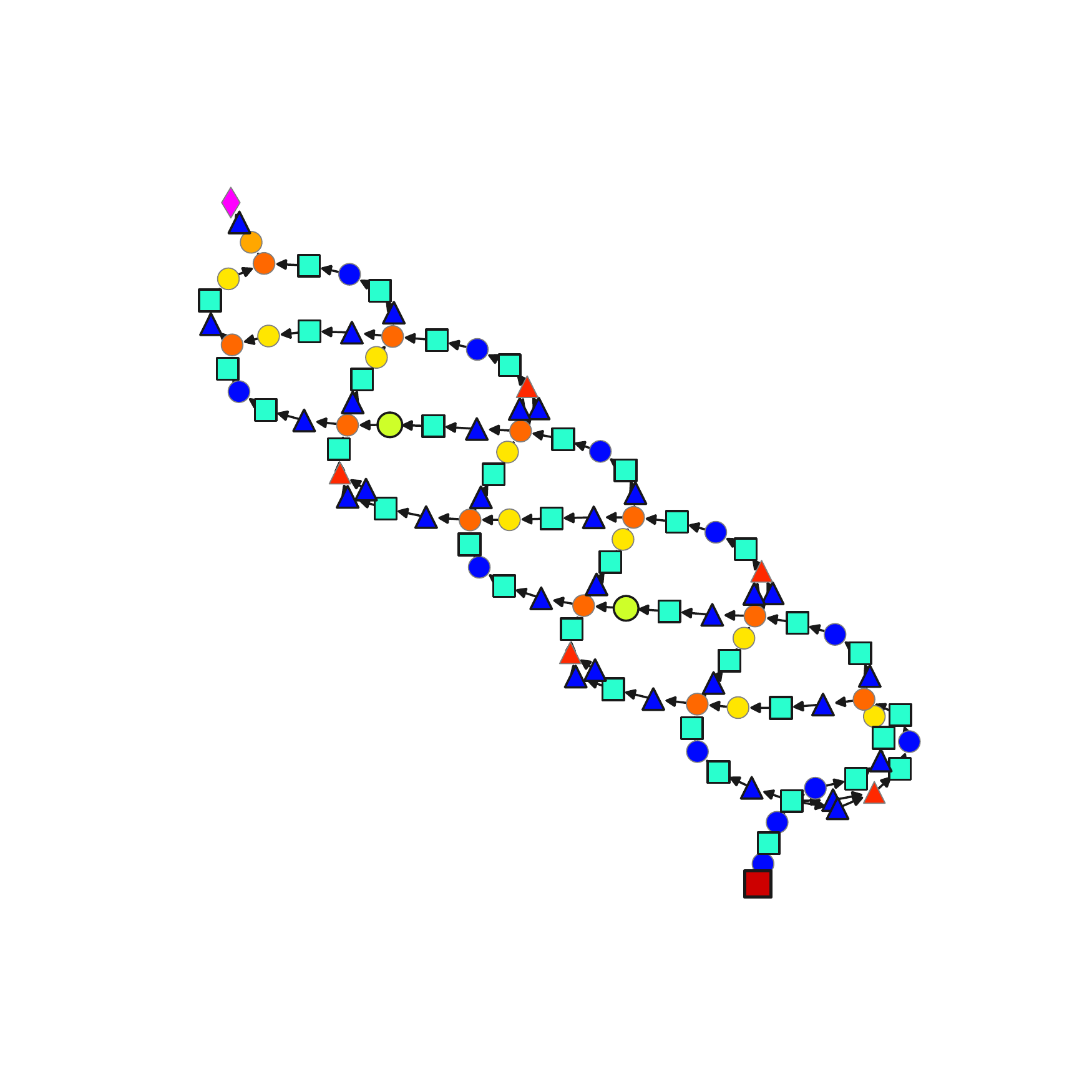}}} & & {\includegraphics[width=\width,align=c,trim={2.3cm 3cm 2.3cm 3cm},clip]{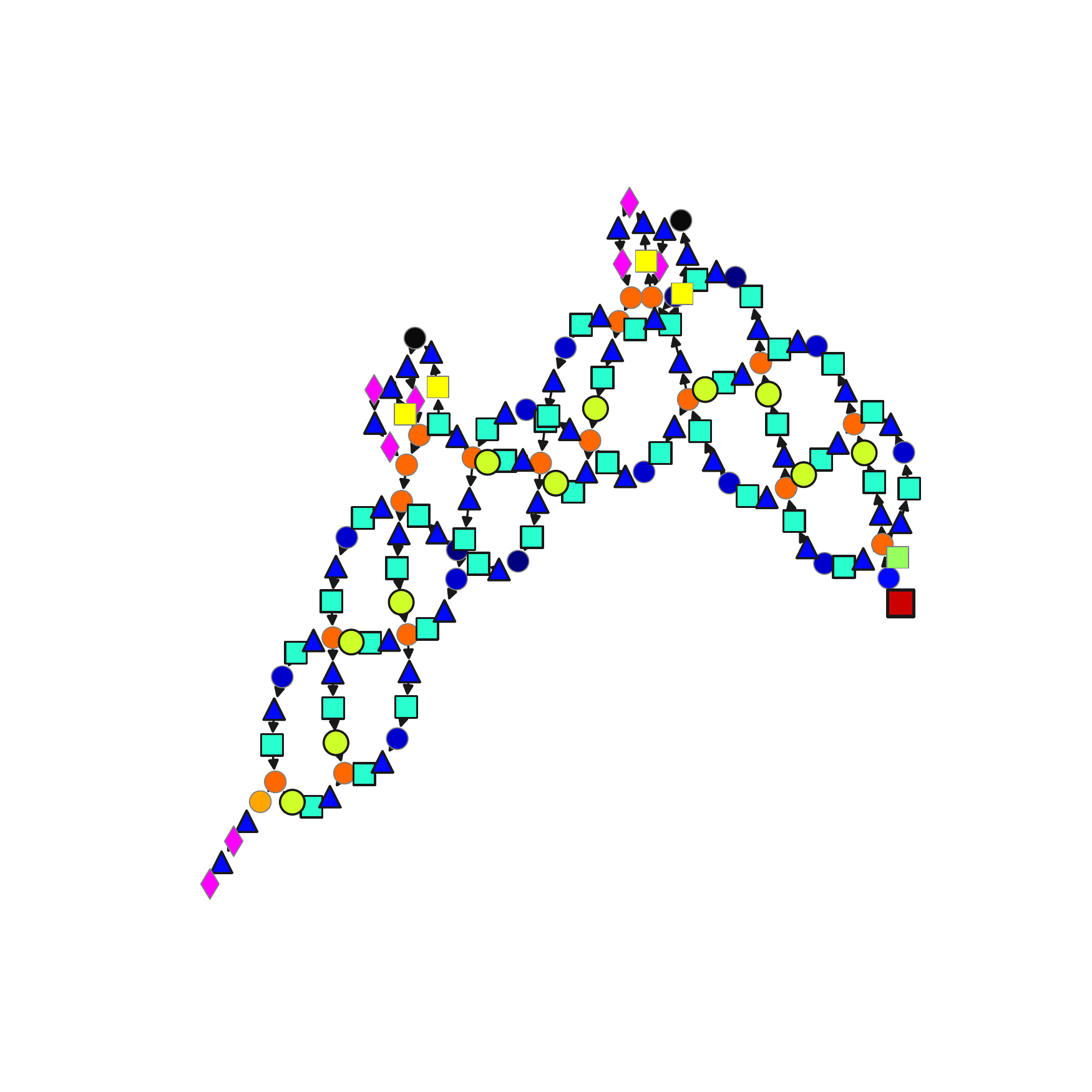}} & 
			{\includegraphics[width=\width,align=c,trim={2.3cm 3cm 2.3cm 3cm},clip]{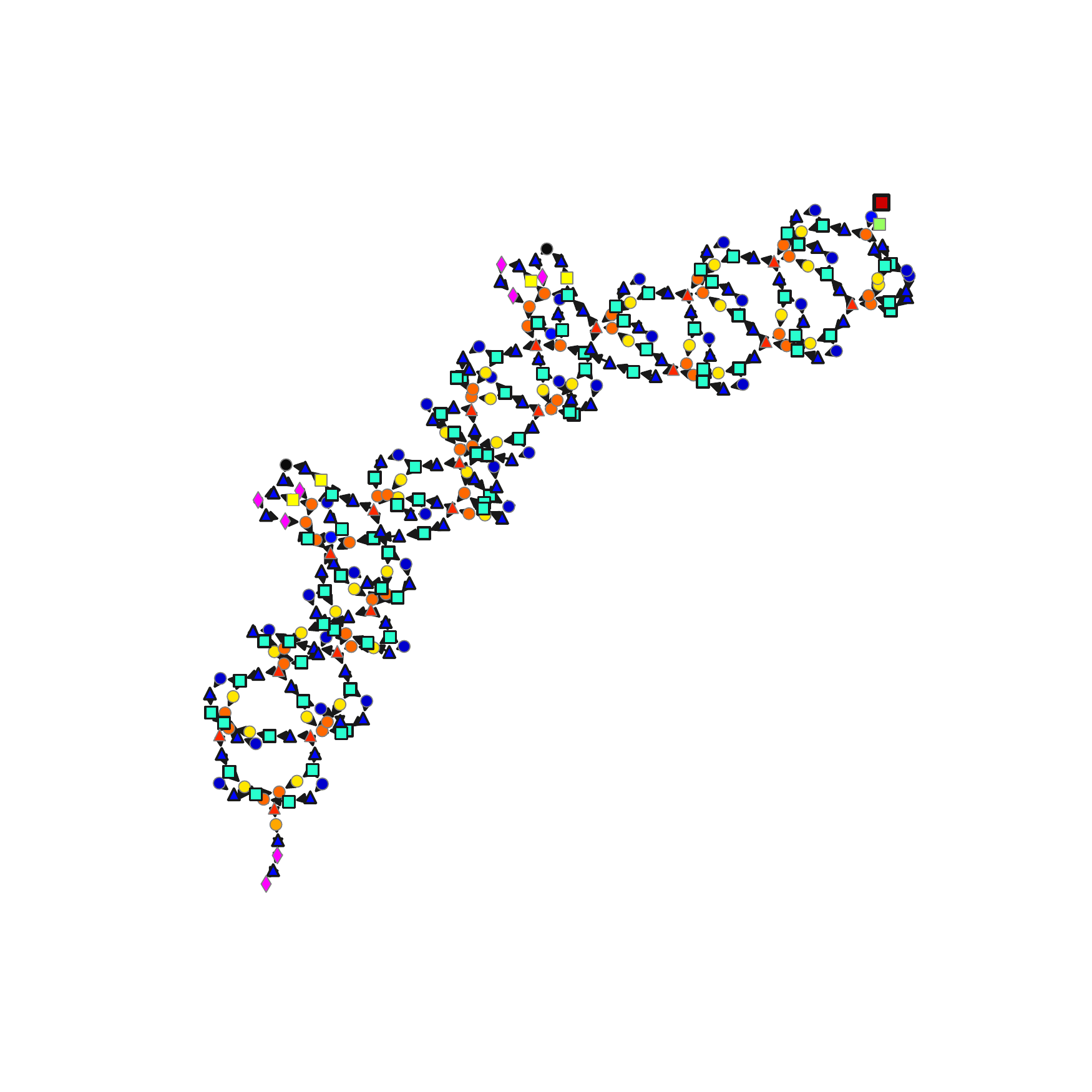}} & 
			\includegraphics[width=\width,align=c,trim={2.3cm 3cm 2.3cm 3cm},clip]{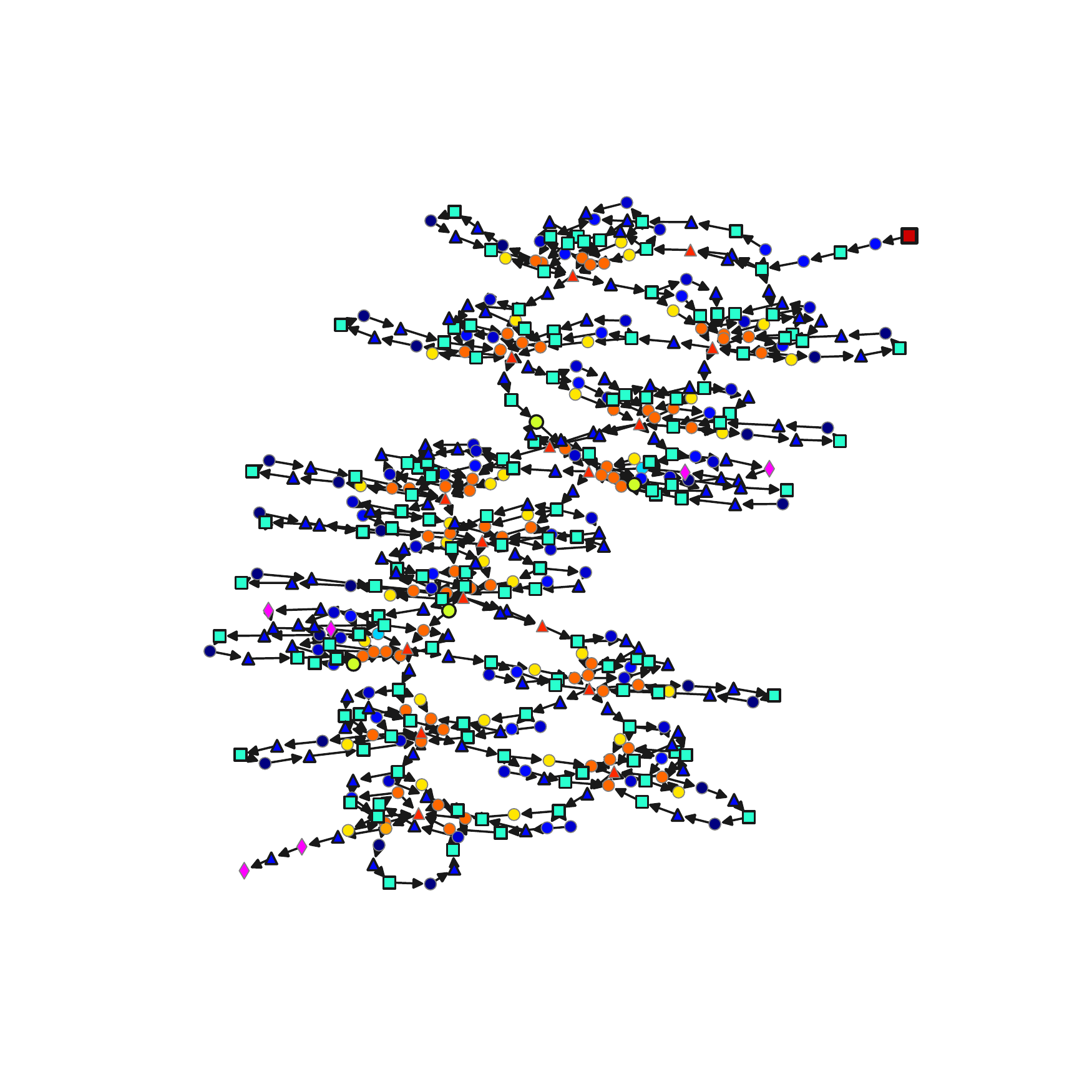} & 
			{\includegraphics[width=\width,align=c,trim={2.3cm 3cm 2.3cm 3cm},clip]{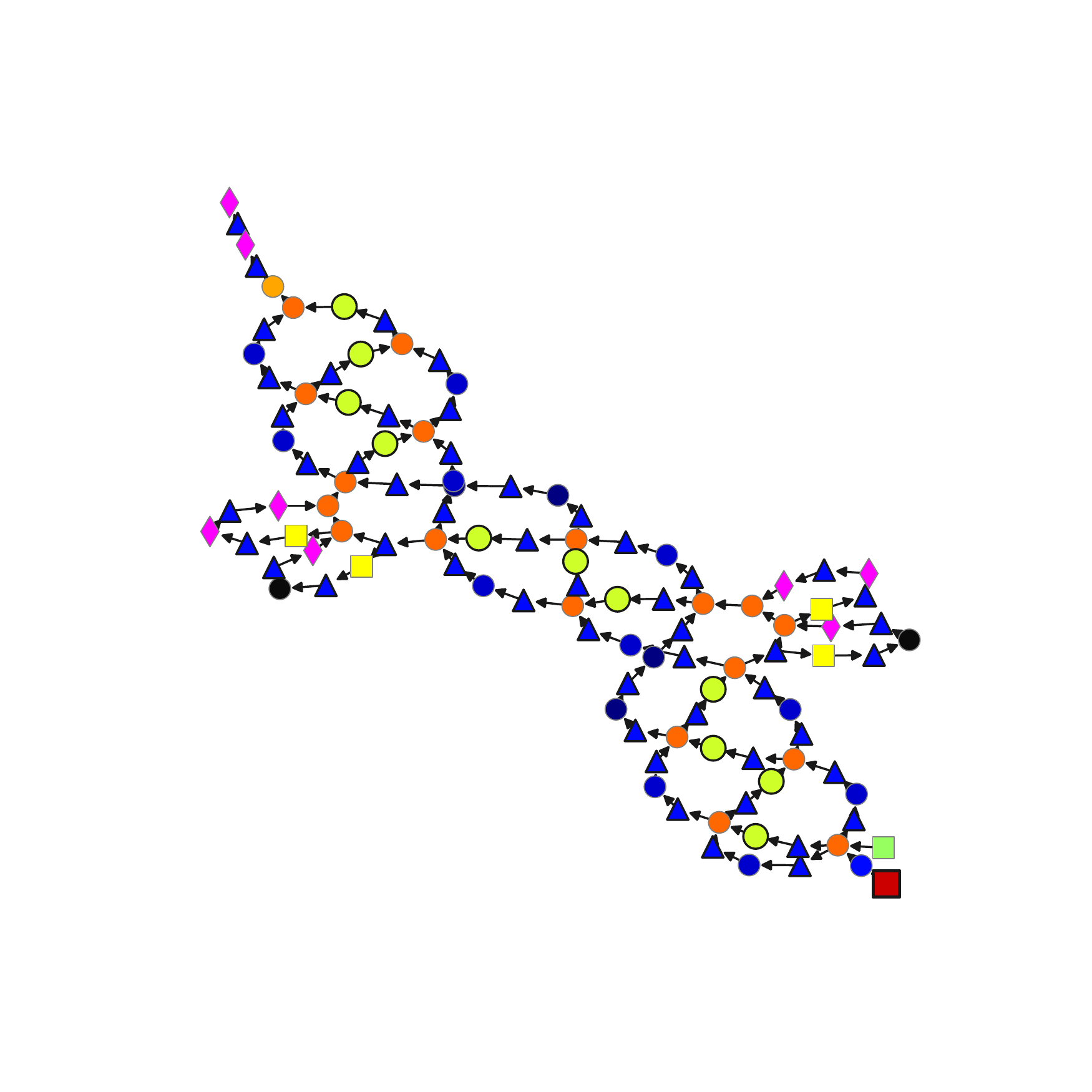}} & 
			{\includegraphics[width=\width,align=c,trim={2.3cm 3cm 2.3cm 3cm},clip]{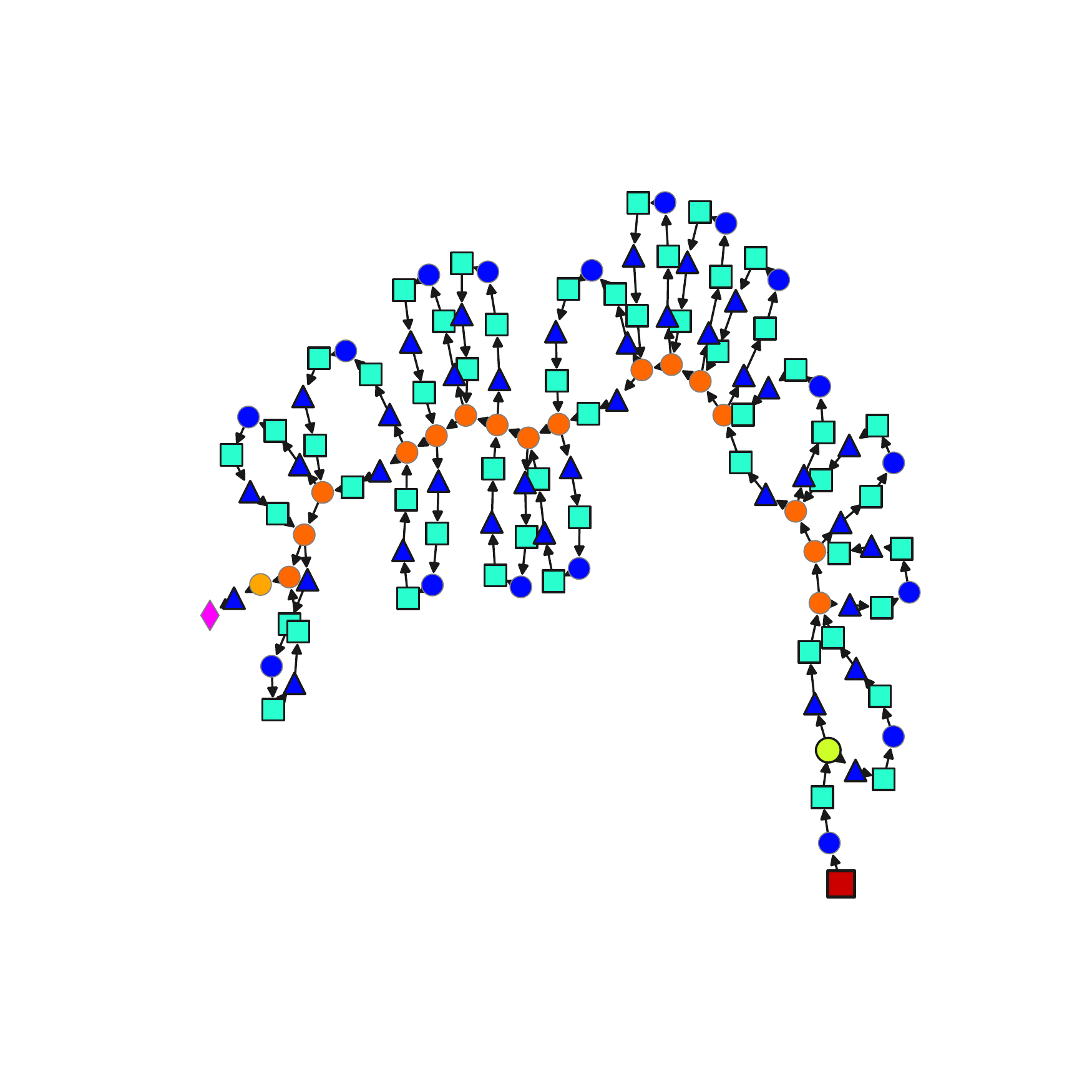}} 
			\vspace{0pt}\Bstrut\\
			& {\small \textbf{\iidtrain}} & {\small \textbf{\iidval/\iidtest}} & & {\small \textbf{\wide}} & {\small \textbf{\deep}} & {\small \textbf{\dense}} & {\small \textbf{\bnfree}} & \scriptsize \textbf{\textsc{ResNet/ViT}} \Bstrut\Tstrut\\
			\cline{2-3}\cline{5-9}
			\#graphs & $10^6$ & 500/500 & & 100 & 100 & 100 & 100 & 1/1\Tstrut\\
			\#cells & 4-18 & 4-18 &  & 4-18 & \textbf{10-36} & 4-18 & 4-18 & 16/12 \\
			\#channels & 16-128 & 32-128 & &  \textbf{128-1216} & 32-208 & 32-240 & 32-336 & 64/128 \\
			\#nodes ($|V|$) & 21-827 & 33-579 &  & 33-579 & \textbf{74-1017} & \textbf{57-993} & 33-503 & 161/114\\
			\% w/o BN & 3.5\% & 4.1\% &  & 4.1\% & 2.0\% & 5.0\% & \textbf{100\%} & 0\%/\textbf{100\%} \\
			\#params(M)* & 0.01-3.1 & 2.5-35 & & \textbf{39-101} & 2.5-15.3 & 2.5-8.8 & 2.5-7.7 & \textbf{23.5}/1.0 \\
			
			avg degree & 2.3\std{0.1} & 2.3\std{0.1}  & & 2.3\std{0.1} & 2.3\std{0.1} & \textbf{2.4}\std{0.1} & \textbf{2.4}\std{0.1} & 2.2/2.3\\
			
			avg path & 14.5\std{4.8} & 14.5\std{4.9}  & & 14.7\std{4.9} & \textbf{26.2}\std{9.3} & 15.1\std{4.1} & 10.0\std{2.8} & 11.2/10.7\\			
		\end{tabular}
		\newcommand{\primwidth}{0.02\textwidth}
		\newcolumntype{x}{>{\centering\arraybackslash\hspace{0pt}}p{0.7cm}}
		\setlength{\tabcolsep}{0.7pt}
		\begin{tabular}{lxxxxxxxxxxxxxxx}
			\toprule
			marker & \includegraphics[width=\primwidth,align=c,trim={0.2cm 0.2cm 0.2cm 0.2cm},clip]{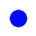} & \includegraphics[width=\primwidth,align=c,trim={0.2cm 0.2cm 0.2cm 0.2cm},clip]{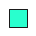} & \includegraphics[width=\primwidth,align=c,trim={0.2cm 0.2cm 0.2cm 0.2cm},clip]{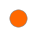} & \includegraphics[width=\primwidth,align=c,trim={0.2cm 0.2cm 0.2cm 0.2cm},clip]{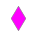} & \includegraphics[width=\primwidth,align=c,trim={0.2cm 0.2cm 0.2cm 0.2cm},clip]{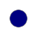} & \includegraphics[width=\primwidth,align=c,trim={0.2cm 0.2cm 0.2cm 0.2cm},clip]{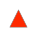} & \includegraphics[width=\primwidth,align=c,trim={0.2cm 0.2cm 0.2cm 0.2cm},clip]{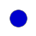} & \includegraphics[width=\primwidth,align=c,trim={0.2cm 0.2cm 0.2cm 0.2cm},clip]{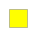} & \includegraphics[width=\primwidth,align=c,trim={0.2cm 0.2cm 0.2cm 0.2cm},clip]{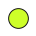} & \includegraphics[width=\primwidth,align=c,trim={0.2cm 0.2cm 0.2cm 0.2cm},clip]{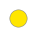} & \includegraphics[width=\primwidth,align=c,trim={0.2cm 0.2cm 0.2cm 0.2cm},clip]{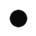} & \includegraphics[width=\primwidth,align=c,trim={0.2cm 0.2cm 0.2cm 0.2cm},clip]{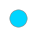} & \includegraphics[width=\primwidth,align=c,trim={0.2cm 0.2cm 0.2cm 0.2cm},clip]{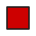} & \includegraphics[width=\primwidth,align=c,trim={0.2cm 0.2cm 0.2cm 0.2cm},clip]{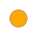} & \includegraphics[width=\primwidth,align=c,trim={0.2cm 0.2cm 0.2cm 0.2cm},clip]{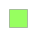}\\
			
			primitive & conv & BN & sum & bias & group conv & concat & \tiny dilated gr. conv & LN & max pool & avg pool & MSA & SE & input & glob avg & pos enc \\
			fraction in \iidtrain (\%) & 36.3 & 25.5 & 11.1 & 6.5 & 5.1 & 3.8 & 2.5 & 2.5 & 1.8 & 1.7 & 1.2 & 1.0 & 0.5 & 0.5 & 0.2 \\
			\bottomrule
		\end{tabular}
	\end{table}
	
	\textbf{In-distribution (\iid) architectures.} We generate a training set of $|\nets|=10^{6}$ architectures and validation/test sets of 500/500 architectures that follow the same generation rules and are considered to be \iid samples. 
	However, training on large architectures can be prohibitive, e.g.~in terms of GPU memory. Thus, in the training set we allow the number of channels and, hence the total number of parameters, to be stochastically defined given computational resources. For example, to train our models we upper bound the number of parameters in the training architectures to around 3M by sampling fewer channels if necessary. In the evaluation sets, the number of channels is fixed. Therefore, this pre-processing step prior to training results in some distribution shift between the training and the validation/test sets. However, the shift is not imposed by our dataset.\looseness-1
	
	\textbf{Out-of-distribution (\ood) architectures.}
	We generate five \ood test sets that follow different generation rules.
	In particular, we define \wide and \deep sets that are of interest due the stronger downstream performance of such nets in large-scale tasks~\cite{golubeva2020wider,zagoruyko2016wide,brown2020language}. These nets are often more challenging to train for fundamental~\cite{nguyen2017loss,srivastava2015training} or computational~\cite{hooker2020hardware} reasons, so predicting their parameters might ease their subsequent optimization.
	We also define the \dense set, since networks with many operations per cell and complex connectivity are underexplored in the literature despite their potential~\cite{huang2017densely}.
	Next, we define the \bnfree set that is of interest due to BN's potential negative side-effects~\cite{galloway2019batch,hendrycks2019benchmarking} and the difficulty or unnecessity of using it in some cases~\cite{wu2018group,qiao2019micro,zhang2019fixup,brock2021characterizing,brock2021high}. 
	We finally add the \textsc{ResNet/ViT} set with two predefined image classification architectures: commonly-used ResNet-50~\cite{he2016deep} and a smaller 12-layer version of the Visual Transformer (ViT)~\cite{dosovitskiy2020image} that has recently received a lot of attention in the vision community.
	Please see \S~\ref{apdx:darts_bg} and \S~\ref{apdx:stats} for further details and statistics of our \dataset dataset.
		
	\vspace{-5pt}
	\section{Improved Graph HyperNetworks: \ghnours\label{sec:model}}
	\vspace{-5pt}
	
	In this section, we introduce our three key improvements to the baseline \ghnbase~described in \S~\ref{sec:bg_ghn} (Fig.~\ref{fig:overview}).
	These components are essential to predict stronger parameters on our task. For the empirical validation of the effectiveness of these components see ablation studies in \S~\ref{sec:our_task} and \S~\ref{apdx:ablations}.\looseness-1
	
	\vspace{-3pt}
	\subsection{Differentiable Normalization of Predicted Parameters\label{sec:renorm}}
	\vspace{-5pt}

	\begin{wraptable}{r}{5.2cm}
		\centering
		\vspace{-13pt}
		\scriptsize
		\caption{Parameter normalizations.\looseness-1}
		\vspace{-1pt}
		\setlength{\tabcolsep}{8pt}
		\label{tab:norm}
		\begin{tabular}{l|l}
			\toprule
			Type of node $v$ & Normalization \\
			\midrule 
			Conv./fully-conn. &  \(\displaystyle \hat{\w}_{p}^{v}\sqrt{{\beta}/{(C_{in}\mathcal{HW})}}  \) \\
			Norm. weights & \(\displaystyle 2 \times
			\text{sigmoid}(\hat{\w}_{p}^{v}/T) \) \\
			Biases & \(\displaystyle \text{ tanh}(\hat{\w}_{p}^{v} / T) \) \\
			\bottomrule
		\end{tabular}
		\vspace{-3pt}
	\end{wraptable}
	
	When training the parameters of a given network from scratch using iterative optimization methods, the initialization of parameters is crucial. A common approach is to use He~\cite{he2015delving} or Glorot~\cite{glorot2010understanding} initialization to stabilize the variance of activations across layers of the network.
	\citet{chang2019principled} showed that when the \params of the network are instead predicted by a hypernetwork, the activations in the network tend to explode or vanish.
	To address the issue of unstable network activations especially for the case of predicting \params of diverse architectures, we apply \emph{operation-dependent normalizations} (Table~\ref{tab:norm}).
	We normalize convolutional and fully-connected weights by following the \emph{fan-in} scheme of~\cite{he2015delving} (see the comparison to \emph{fan-out} in \S~\ref{apdx:ablations}): $\hat{\w}_{p}^{v}\sqrt{{\beta}/{(C_{in}\mathcal{HW})}}$, where $C_{in},\mathcal{H,W}$ are the number of input channels and spatial dimensions of weights $\hat{\w}_{p}^{v}$, respectively; and $\beta$ is a nonlinearity specific constant following the analysis in~\cite{he2015delving}.
	The parameters of normalization layers such as BN and LN, as well as biases typically initialized with constants, are normalized by applying a squashing function with temperature $T$ to imitate the empirical distributions of models trained with SGD (see Table~\ref{tab:norm}).	
	These are differentiable normalizations, so that they are applied at training (and testing) time.
	Further analysis of our normalization and its stabilizing effect on activations is presented in \S~\ref{apdx:renorm}.\looseness-1
	
	\vspace{-3pt}
	\subsection{Enhancing Long-range Message Propagation\label{sec:sp_edges}}
	\vspace{-5pt}
	
	\begin{wrapfigure}{r}{3.2cm}
		\centering
		\vspace{-30pt}
		\small
		\includegraphics[width=0.14\textwidth,align=c,trim={2.8cm 3cm 2.7cm 3cm}, clip]{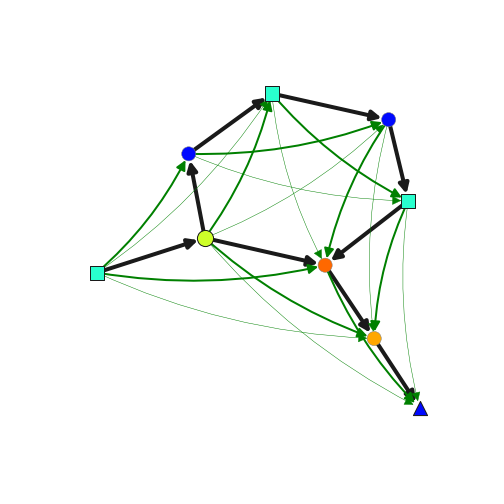} 
		\vspace{-5pt}
		\caption{\small Virtual edges (in green) allow for better capture of global context.}\label{fig:long_range}
		\vspace{-5pt}
	\end{wrapfigure}

	Computational graphs often take the form of long chains (Table~\ref{tab:graphs}) with only a few incoming/outcoming edges per node. This structure might hinder long-range propagation of information between nodes~\cite{alon2020bottleneck}.	
	Different approaches to alleviate the long-range propagation problem exist~\cite{el1996hierarchical,liu2020non,pei2020geom}, including stacking GHNs in~\cite{zhang2018graph}.
	Instead we adopt simple graph-based heuristics in line with recent works~\cite{you2019position,yang2021spagan}. In particular, we add \emph{virtual edges} between two nodes $v$ and $u$ and weight them based on the shortest path $s_{vu}$ between them (Fig.~\ref{fig:long_range}). To avoid interference with the \emph{real} edges in the computational graph, we introduce a separate MLP\textsubscript{sp} to transform the features of the nodes connected through these virtual edges, and redefine the message passing of Equation \ref{eq:ghn_prop} as:\looseness-1
	\setlength{\belowdisplayskip}{1pt}
	\setlength{\abovedisplayskip}{1pt}
	\begin{equation}
	\label{eq:ghn_sp}
	\mathbf{m}_v^t = \sum\nolimits_{u \in \neigh_v^{\pi}} \text{MLP}(\mathbf{h}_u^t) + \sum\nolimits_{u \in \neigh_{v}^{(\text{sp})}} \frac{1}{s_{vu}} \text{MLP}_{\text{sp}}(\mathbf{h}_u^t),
	\end{equation}
	\noindent where $\neigh_{v}^{(sp)}$ are neighbors satisfying $1 < s_{vu} \leq s^{(\max)}$, and $s^{(\max)}$ is a hyperparameter.
	To maintain the same number of trainable parameters as in \ghnbase, we decrease MLPs' sizes appropriately. 
	Despite its simplicity, this approach is effective (see the comparison to stacking GHNs in \S~\ref{apdx:ablations}).
	
	\vspace{-2pt}
	\subsection{Meta-batching Architectures During Training\label{sec:meta_batch}}
	\vspace{-3pt}
	
	\ghnbase~updates its parameters $\theta$ based on a single architecture sampled for each batch of images (Equation~\ref{eq:solution}).
	In vanilla SGD training, larger batches of images often speed up convergence by reducing gradient noise and improve model's performance~\cite{radiuk2017impact}. Therefore, we define a meta-batch $b_m$ as the number of architectures sampled per batch of images. Both the parameter prediction and the forward/backward passes through the architectures in a meta-batch can be done in parallel. We then average the gradients across $b_m$ to update the parameters $\theta$ of $H_\domain$: $\nabla_\theta \loss = 1/{b_m} \sum_{i=1}^{b_m} \nabla_\theta \loss_i$. Further analysis of the meta-batching effect on the training loss and convergence speed is presented in \S~\ref{apdx:meta}.\looseness-1	
	
	\vspace{-3pt}
	\section{Experiments\label{sec:exper}}
	\vspace{-5pt}
	
	We focus the evaluation of \ghnours~on our parameter prediction task (\S~\ref{sec:our_task}). In addition, we show beneficial side-effects of i) learning a stronger neural architecture representation using \ghnours in analyzing networks (\S~\ref{sec:prop_pred}) and ii) predicting parameters for fine-tuning (\S~\ref{sec:finetune}). We provide further experimental and implementation details, as well as more results supporting our arguments in \S~\ref{apdx:exper}. 
	
	\textbf{Datasets.} We use the \dataset dataset of architectures (\S~\ref{sec:dataset}) as well as two image classification datasets  $\domain_1$ (CIFAR-10~\cite{krizhevsky2009learning}) and $\domain_2$ (ImageNet~\cite{russakovsky2015imagenet}). CIFAR-10 consists of 50k training and 10k test images of size 32$\PLH$32$\PLH$3 and 10 object categories.
	ImageNet is a larger scale dataset with 1.28M training and 50k test images of variable size and 1000 fine-grained object categories. We resize ImageNet images to 224$\PLH$224$\PLH$3 following~\cite{liu2018darts,zhang2018graph}. We use 5k/50k training images as a validation set in CIFAR-10/ImageNet and 500 validation architectures of \dataset for hyperparameter tuning.\looseness-1
	
	\textbf{Baselines.} Our baselines include \ghnbase and a simple MLP that only has access to operations, but not to the connections between them.
	This MLP baseline is obtained by replacing the GatedGNN with an MLP in our \ghnours.
	Since GHNs were originally introduced for small architectures of $\sim50$ nodes and only trained on CIFAR-10, we reimplement\footnote{While source code for GHNs~\cite{zhang2018graph} is unavailable, we appreciate the authors' help in implementing some steps.\looseness=-1} them and scale them up by introducing minor modifications to their decoder that enable their training on ImageNet and on larger architectures of up to 1000 nodes (see \S~\ref{apdx:ghn_1} for details). 
	We use the same hyperparameters to train the baselines and \ghnours.\looseness-1
	
	\textbf{Iterative optimizers.}
	In the parameter prediction experiments, we also compare our model to standard optimization methods: SGD and Adam~\cite{kingma2014adam}.
	We use off-the-shelf hyperparameters common in the literature~\cite{zhang2018graph,liu2018darts,chen2019progressive,yang2020cars,he2020milenas,li2020sgas}. On CIFAR-10, we train evaluation architectures with SGD/Adam, initial learning rate $\eta=0.025$ / $\eta=0.001$, batch size $b=96$ and up to 50 epochs. With Adam, we train only 300 evaluation architectures as a rough estimation of an average performance.
	On ImageNet, we train them with SGD, $\eta=0.1$ and $b=128$,
	and, for computational reasons (given 1402 evaluation architectures in total), we limit training with SGD to 1 epoch.
	We have also considered meta-optimizers, such as~\cite{andrychowicz2016learning,ravi2016optimization}. However, we were unable to scale them to diverse and large architectures of our \dataset, since their LSTM requires a separate hidden state for every trainable parameter in the architecture. The scalable variants exist~\cite{wichrowska2017learned,metz2020tasks}, but are hard to reproduce without open source code.\looseness-1
	
	\textbf{Additional experimental details.}
	We follow~\cite{zhang2018graph} and train GHNs with Adam, $\eta=0.001$ and batch size of 64 images for CIFAR-10 and 256 for ImageNet. We train for up to 300 epochs, except for one experiment in the ablation studies,
	where we train one GHN with $b_m = 1$ eight times longer, i.e. for 2400 epochs.
	All GHNs in our experiments use $T=1$ propagation (Equation~\ref{eq:ghn_prop}), as we found the original $T=5$ of~\cite{zhang2018graph} to be inefficient and it did not improve the accuracies in our task.
	\ghnours~uses $s^{(\max)}=50$ and $b_m=8$ and additionally uses LN that slightly further improves results (see these ablations in \S~\ref{apdx:ablations}). Model selection is performed on the validation sets, but the results in our paper are reported on the test sets to enable their direct comparison.\looseness-1
	
	\vspace{-2pt}
	\subsection{Parameter Prediction\label{sec:our_task}}
	\vspace{-2pt}

	\textbf{Experimental setup.} We trained our \ghnours and baselines on the training architectures and training images, i.e.~a separate model is trained for CIFAR-10 and ImageNet. According to our \dataset benchmark, we assess whether these models can generalize to unseen in-distribution (ID) and out-of-distribution (OOD) test architectures from our \dataset. We measure this generalization by predicting \params for the test architectures and computing their classification accuracies on the test images of CIFAR-10 (Table~\ref{tab:bench_c10}) and ImageNet (Table~\ref{tab:bench_imagenet}). The evaluation architectures with batch norm (BN) have running statistics, which are not learned by gradient descent~\cite{ioffe2015batch}, and hence are not predicted by our GHNs. To alleviate that, we follow~\cite{zhang2018graph} and evaluate the networks with BN by computing per batch statistics with batch size of 64 images. This is further discussed in \S~\ref{apdx:details}.\looseness-1
	
	\textbf{Results.}
	Despite \ghnours never observed the test architectures, \ghnours predicts good parameters for them making the test networks perform surprisingly well on both image datasets (Tables~\ref{tab:bench_c10} and~\ref{tab:bench_imagenet}). Our results are especially strong on CIFAR-10, where some architectures with predicted parameters achieve up to 77.1\%, while the best accuracy of training with SGD for 50 epochs is around 15\% more. We even show good results on ImageNet, where for some architectures we achieve a top-5 accuracy of up to 48.3\%. While these results are low for direct downstream applications, they are remarkable for three main reasons. First, to train GHNs by optimizing Equation~\ref{eq:solution}, we do not rely on the prohibitively expensive procedure of training the architectures $\nets$ by SGD. Second, GHNs rely on a single forward pass to predict all parameters. Third, these results are obtained for unseen architectures, including the OOD ones. Even in the case of severe distribution shifts (e.g.~ResNet-50\footnote{Large architectures with bottleneck layers such as ResNet-50 do not appear during training.}) and underrepresented networks (e.g. ViT\footnote{Architectures such as ViT do not include BN and, except for the first layer, convolutions -- the two most frequent operations in the training set.}), our model still predicts \params that perform better than random ones. On CIFAR-10, generalization of \ghnours is particularly strong with a 58.6\% accuracy on ResNet-50.	
	
	\begin{table}[b!]
		\centering
		\vspace{-10pt}
		\caption{CIFAR-10 results of predicted parameters for unseen ID and OOD architectures of \dataset. Mean (\sem{}standard error of the mean) accuracies are reported (random chance $\approx$10\%). $^\dagger$The number of parameter updates.\looseness-1}
		\label{tab:bench_c10}
		\vspace{2pt}
		\small
		\centering
		\setlength{\tabcolsep}{3pt}
		\begin{tabular}{llp{0.1cm}llp{0.5cm}llllc}
			\toprule
			
			\textbf{\textsc{Method}} & \textbf{\#upd}$^\dagger$ & &
			\multicolumn{2}{c}{\textbf{\textsc{\iid-test}}} &
			& 
			\multicolumn{5}{c}{\textbf{\textsc{OOD-test}}} \\
			
			& & & \multicolumn{1}{c}{avg} & max & & \wide & \deep & \dense & \bnfree & \scriptsize \textsc{ResNet/ViT} \\ 
			\cline{1-2}\cline{4-5}\cline{7-11}
			
			MLP & 1 & & 42.2\sem{0.6} & 60.2 & & 22.3\sem{0.9} & 37.9\sem{1.2} & 44.8\sem{1.1} & 23.9\sem{0.7} & 17.7/10.0 \Tstrut \\
			
			\ghnbase & 1 & & 51.4\sem{0.4} & 59.9 &  & 43.1\sem{1.7} & 48.3\sem{0.8} & 51.8\sem{0.9} & 13.7\sem{0.3} & 19.2/\textbf{18.2} \\
			
			\ghnours & 1 & & \textbf{66.9}\sem{0.3} & \textbf{77.1} & & \textbf{64.0}\sem{1.1} & \textbf{60.5}\sem{1.2} & \textbf{65.8}\sem{0.7} & \textbf{36.8}\sem{1.5} & \textbf{58.6}/11.4 \\
			
			\hline\hline
			
			\multicolumn{10}{l}{\textbf{Iterative optimizers (all architectures are \iid in this case)}} \Tstrut \\
			
			SGD (1 epoch) & \scriptsize $0.5 \PLH 10^3$ & & 46.1\sem{0.4} & 66.5 & & 47.2\sem{1.1} & 34.2\sem{1.1} & 45.3\sem{0.7} & 18.0\sem{1.1} & 61.8/34.5 \\
			
			SGD (5 epochs) & \scriptsize $2.5\PLH 10^3$ & & 69.2\sem{0.4} & 82.4 & & 71.2\sem{0.3} & 56.7\sem{1.6} & 67.8\sem{0.9} & 29.0\sem{2.0} & 78.2/52.5\\
	
			SGD (50 epochs) & \scriptsize $25\PLH 10^3$ & & 88.5\sem{0.3} & 93.1 & & 88.9\sem{1.2} & 84.5\sem{1.2} & 87.3\sem{0.8} & 45.6\sem{3.6} & 93.5/75.7 \\
			
			Adam (50 epochs) & \scriptsize $25\PLH 10^3$ & & 84.0\sem{0.8} & 89.5 & & 82.0\sem{1.6} & 76.2\sem{2.6} & 84.8\sem{0.4} & 38.8\sem{4.8} & 91.5/79.4 \\
			
			\bottomrule
		\end{tabular}
		\vspace{-5pt}
	\end{table}
	
	\begin{table}[b!]
		\centering
		\vspace{-5pt}
		\caption{ImageNet results on \dataset.
			Mean (\sem{}standard error of the mean) top-5 accuracies are reported (random chance $\approx$0.5\%).
			$^*$Estimated on ResNet-50 with batch size 128.
		}
		\label{tab:bench_imagenet}
		\vspace{2pt}
		\small
		\centering
		\setlength{\tabcolsep}{1.5pt}
		\begin{tabular}{llccllp{0.1cm}llllc}
			\toprule
			
			\textbf{\textsc{Method}} & \textbf{\#upd} & \scriptsize \textbf{GPU sec.} & \scriptsize \textbf{CPU sec.} & \multicolumn{2}{c}{\textbf{\textsc{\iid-test}}} &
			& 
			\multicolumn{5}{c}{\textbf{\textsc{OOD-test}}} \\
			
			& & \multicolumn{1}{c}{avg} & \multicolumn{1}{c}{avg} & \multicolumn{1}{c}{avg} & max & & \wide & \deep & \dense & \bnfree & \scriptsize \textsc{ResNet/ViT} \\
			\cline{1-4}\cline{5-6}\cline{8-12}
			
			\ghnbase & \scriptsize 1 & \scriptsize 0.3 & \scriptsize 0.5 & 17.2\sem{0.4} & 32.1 &  & 15.8\sem{0.9} & 15.9\sem{0.8} & 15.1\sem{0.7} & 0.5\sem{0.0} & \textbf{6.9}/0.9 \Tstrut\\ 
			
			\ghnours & \scriptsize 1 & \scriptsize 0.3 & \scriptsize 0.7 & \textbf{27.2}\sem{0.6} & \textbf{48.3} & & \textbf{19.4}\sem{1.4} & \textbf{24.7}\sem{1.4} & \textbf{26.4}\sem{1.2} & \textbf{7.2}\sem{0.6} & 5.3/\textbf{4.4} \Bstrut\\
			
			\hline\hline
			
			\multicolumn{10}{l}{\textbf{Iterative optimizers (all architectures are \iid in this case)}} \Tstrut \\
			
			SGD (1 step) & \scriptsize 1 & \scriptsize 0.4 & \scriptsize 6.0 & 0.5\sem{0.0} & 0.7 & & 0.5\sem{0.0} & 0.5\sem{0.0} & 0.5\sem{0.0} & 0.5\sem{0.0} & 0.5/0.5\\ 
			SGD (5000 steps) & \scriptsize $5$k & \scriptsize $2\PLH 10^{3}$ & \scriptsize $3\PLH 10^{4}$ & 25.6\sem{0.3} & 50.7 & & 26.2\std{1.4} & 13.2\sem{1.1} & 25.4\sem{1.1} & 4.8\sem{0.8} & 34.8/24.3 \\
			SGD (10000 steps) & \scriptsize $10$k & \scriptsize $4\PLH 10^{3}$ & \scriptsize $6\PLH 10^{4}$ & 37.7\sem{0.6} & 62.0 & & 38.7\sem{1.6} & 22.1\sem{1.4} & 36.3\sem{1.2} & 8.0\sem{1.2} & 49.0/33.4 \\
			SGD (100 epochs) & \scriptsize $1000$k & \scriptsize $6\PLH 10^{5*}$ & \scriptsize $6\PLH 10^{7*}$ & $-$ &  & & $-$ & $-$ & $-$ & $-$ & 92.9/72.2 \\
			
			\bottomrule
		\end{tabular}
		\vspace{-2pt}
	\end{table}
	
	On both image datasets, our \ghnours significantly outperforms \ghnbase on all test subsets of \dataset with more than a 20\% absolute gain in certain cases, e.g.~36.8\% vs 13.7\% on the \bnfree networks (Table~\ref{tab:bench_c10}). Exploiting the structure of computational graphs is a critical property of GHNs with the accuracy dropping from 66.9\% to 42.2\% on \iid (and even more on \ood) architectures when we replace the GatedGNN of \ghnours~with an MLP.
	Compared to iterative optimization methods, \ghnours predicts parameters achieving an accuracy similar to $\sim$2500 and $\sim$5000 iterations of SGD on CIFAR-10 and ImageNet respectively.
	In contrast, \ghnbase performs similarly to only $\sim$500 and $\sim$2000 (not shown in Table~\ref{tab:bench_imagenet}) iterations respectively. 
	Comparing SGD to Adam, the latter performs worse in general except for the ViT architectures similar to~\cite{zhang2019adam,dosovitskiy2020image}.

	To report speeds on ImageNet in Table~\ref{tab:bench_imagenet}, we use a dedicated machine with a single NVIDIA V100-32GB and Intel Xeon CPU E5-1620 v4@ 3.50GHz. So for SGD these numbers can be reduced by using faster computing infrastructure and more optimal hyperparameters~\cite{goyal2017accurate}.
	Using our setup, SGD requires on average $10^4 \PLH$ more time on a GPU ($10^5 \PLH$ on a CPU) to obtain \params that yield performance similar to \ghnours.
	As a concrete example, AlexNet~\cite{krizhevsky2012imagenet} requires around 50 GPU hours (on our setup) to achieve a 81.8\% top-5 accuracy, while on some architectures we achieve $\geq$48.0\% in just 0.3 GPU seconds.\looseness-1

	\begin{minipage}[tbhp]{\textwidth}
		\vspace{5pt}
		\begin{minipage}[b]{0.49\textwidth}
			\captionof{table}{Ablating \ghnours on CIFAR-10. An average rank of the model is computed across all \iid and \ood test architectures.}
			\label{tab:ablations}
			\vspace{-3pt}
			\centering
			\scriptsize
			\setlength{\tabcolsep}{2pt}
			\begin{tabular}{lcc|c}
				\toprule
				\textbf{\textsc{Model}} & \multicolumn{1}{c}{\textbf{\textsc{\iid-test}}} & \multicolumn{1}{c|}{\textbf{\textsc{OOD-test}}} & \textbf{\textsc{Avg. rank}} \\ 
				\midrule
				
				\ghnours & \textbf{66.9}\sem{0.3} & \textbf{56.8}\sem{0.8} & \textbf{1.9}\Bstrut\\
				\hline 
				
				1000 training architectures & 65.1\sem{0.5} & 52.5\sem{1.0} & 2.6\Tstrut\\
				
				No normalization (\S~\ref{sec:renorm}) & 62.6\sem{0.6} & 47.1\sem{1.2} & 3.9\\
				
				No virtual edges (\S~\ref{sec:sp_edges}) & 61.5\sem{0.4} & 53.9\sem{0.6} & 4.1\\
				
				No meta-batch ($b_m=1$, \S~\ref{sec:meta_batch}) & 54.3\sem{0.3} & 47.5\sem{0.6} & 5.5 \\
				
				$b_m=1$, train 8$\PLH$ longer & 62.4\sem{0.5} & 51.9\sem{1.0} & 3.7\\
				
				No GatedGNN (MLP) & 42.2\sem{0.6} & 32.2\sem{0.7} & 7.4 \\

				\hline

				\ghnbase & 51.4\sem{0.4} & 39.2\sem{0.9} & 6.8\Tstrut\\
				
				\bottomrule
			\end{tabular}
		\end{minipage}
		\hfill
		\begin{minipage}[b]{0.45\textwidth}
			\centering
			\hspace{-9pt}
			{\includegraphics[width=1.03\textwidth,trim={0.5cm 0.5cm 5.5cm 0.5cm},clip,align=c]{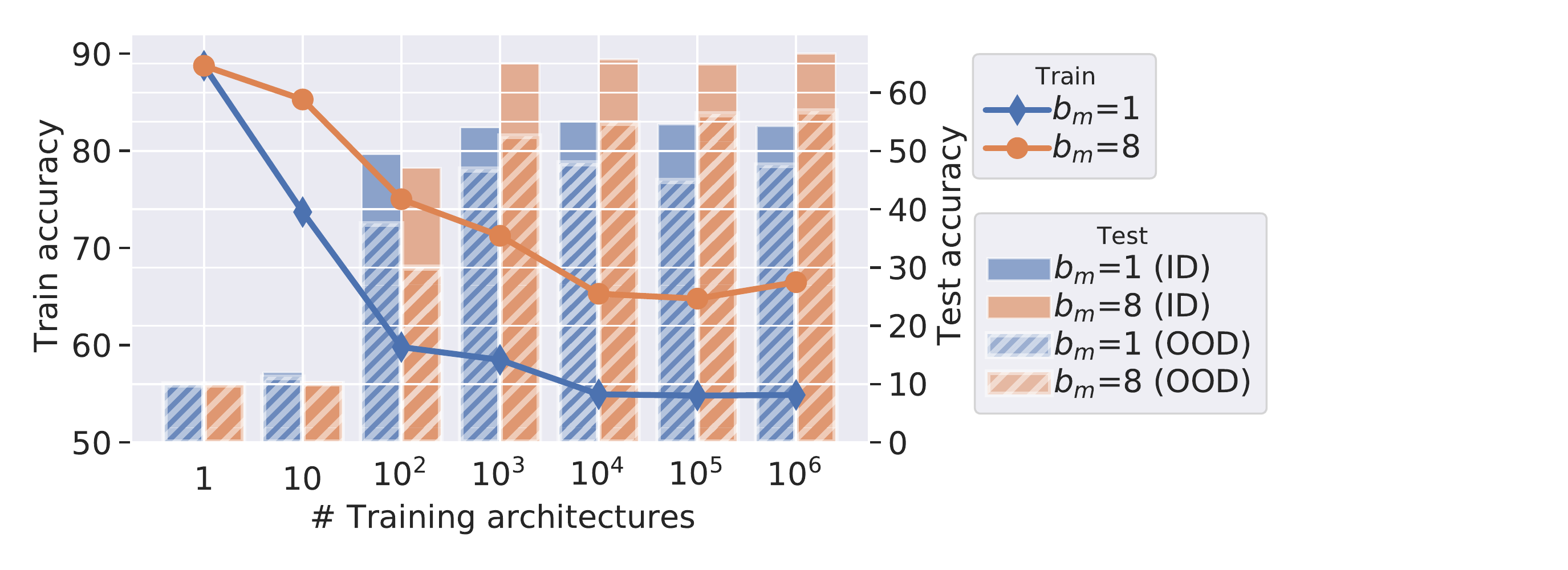}}
			\vspace{-2pt}
			\captionof{figure}{\hspace{5pt}\ghnours~with meta batch $b_m = 8$ versus $b_m = 1$ for different numbers of training architectures on CIFAR-10.}
			\label{fig:acc_arch}
			\vspace{-40pt}
		\end{minipage}
		\vspace{5pt}
	\end{minipage}

	Ablations (Table~\ref{tab:ablations}) show that all three components proposed in \S~\ref{sec:model} are important. Normalization is particularly important for OOD generalization with the largest drops on the \wide and \bnfree networks (see \S~\ref{apdx:ablations}). Using meta-batching ($b_m = 8$) is also essential and helps stabilize training and accelerate convergence (see \S~\ref{apdx:ghn_2}).
	We also confirm that the performance gap between $b_m = 1$ and $b_m = 8$ is not primarily due to the observation of more architectures, since the ablated \ghnours with $b_m = 1$ trained eight times longer is still inferior.
	The gap between $b_m = 8$ and $b_m = 1$ becomes pronounced with \emph{at least} 1k training architectures (Fig.~\ref{fig:acc_arch}).
	When training with fewer architectures (e.g.~100), the GHN with meta-batching starts to overfit to the training architectures.
	Given our challenging setup with unseen evaluation architectures, it is surprising that using 1k training architectures already gives strong results. However, OOD generalization degrades in this case compared to using all 1M architectures, especially on the \bnfree networks (see \S~\ref{apdx:ghn_2}).
	When training GHNs on just a few architectures, the training accuracy soars to the level of training them with SGD. With more architectures, it generally decreases indicating classic overfitting and underfitting cases.\looseness-1

	\vspace{-2pt}
	\subsection{Property Prediction\label{sec:prop_pred}}
	\vspace{-3pt}
	
	Representing computational graphs of neural architectures is a challenging problem~\cite{li2020neural,wen2019neural,jin2019auto,kriege2020survey,makarov2021survey}.
	We verify if GHNs are capable of doing that out-of-the-box in the property prediction experiments. We also experiment with architecture comparison in \S~\ref{apdx:graph_compare}. Our hypothesis is that by better solving our parameter prediction task, GHNs should also better solve graph representation tasks.
	
	\textbf{Experimental setup.}
	We predict the properties of architectures given their graph embeddings obtained by averaging node features\footnote{A fixed size graph embedding for the architecture $\f$ can be computed by averaging the output node features: $\mathbf{h}_{a}=\frac{1}{|V|} \sum_{v \in V} \mathbf{h}_v^T$, where $\mathbf{h}_a \in \mathbb{R}^d$ and $d$ is the dimensionality of node features.
	}.
	We consider four such properties (see \S~\ref{apdx:prop} for details):
	\vspace{-5pt}
	\begin{itemize}[leftmargin=5mm]
		\setlength\itemsep{0em}
		\item Accuracy on the ``clean'' (original) validation set of images;
		\item Accuracy on a corrupted set (obtained by adding the Gaussian noise to images following~\cite{hendrycks2019benchmarking});
		\item Inference speed (latency or GPU seconds per a batch of images);\looseness-1
		\item Convergence speed (the number of SGD iterations to achieve a certain training accuracy).
	\end{itemize}
	\vspace{-3pt}
	
	Estimating these properties accurately can have direct practical benefits. Clean and corrupted accuracies can be used to search for the best performing architectures (e.g. for the NAS task); inference speed can be used to choose the fastest network, so by estimating these properties we can trade-off accurate, robust and fast networks~\cite{cai2019onceforall}. Convergence speed can be used to find networks that are easier to optimize.
	These properties correlate poorly with each other and between CIFAR-10 and ImageNet (\S~\ref{apdx:prop}), so they require the model to capture different regularities of graphs. 
	While specialized methods to estimate some of these properties exist, often as a NAS task~\cite{wen2020neural,lukasik2020neural,baker2017accelerating,liu2018progressive,li2020neural}, our GHNs provide a generic representation that can be easily used for many such properties.
	For each property, we train a simple regression model using graph embeddings and ground truth property values. We use 500 validation architectures of \dataset for training the regression model and tuning its hyperparameters (see \S~\ref{apdx:prop} for details).
	We then use 500 testing architectures of \dataset to measure Kendall's Tau rank correlation between the predicted and ground truth property values similar to~\cite{wen2020neural}.\looseness-1
	
	\textbf{Additional baseline.} 
	We compare to the Neural Predictor (NeuPred)~\cite{wen2020neural}. NeuPred is based on directed graph convolution and is developed for accuracy prediction achieving strong NAS results. We train a separate such NeuPred for each property from scratch following their hyperparameters.

	\begin{wrapfigure}{r}{7cm}
		\vspace{-15pt}
		{\includegraphics[align=c,width=0.5\textwidth]{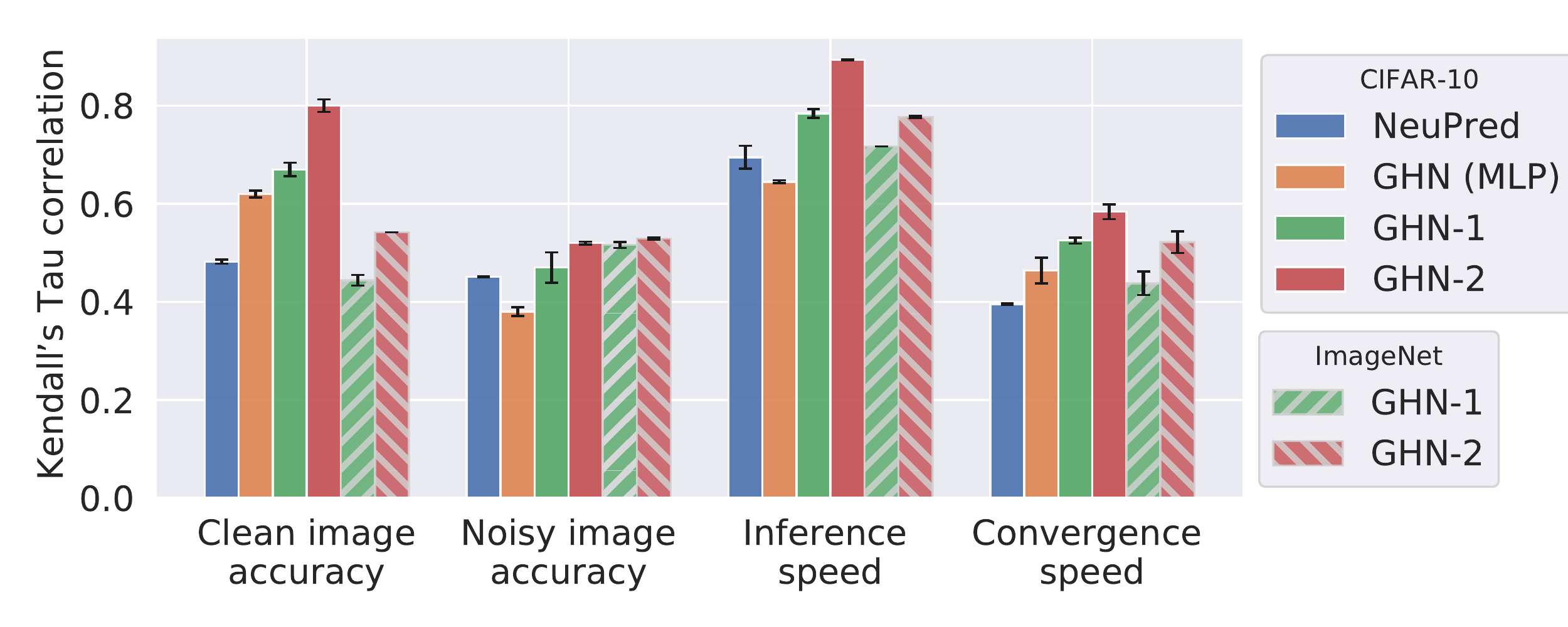}}
		\vspace{-5pt}
		\caption{\small Property prediction of neural networks in terms of correlation (higher is better). Error bars denote the standard deviation across 5 runs.}\label{fig:properties}
		\vspace{-5pt}
	\end{wrapfigure}
	
	\textbf{Results.}
	\ghnours consistently outperforms the \ghnbase and MLP baselines as well as NeuPred (Fig.~\ref{fig:properties}). In \S~\ref{apdx:prop}, we also provide results verifying if higher correlations translate to downstream gains. For example, on CIFAR-10 by choosing the most accurate architecture according to the regression model and training it from scratch following~\cite{liu2018darts,zhang2018graph}, we obtained a 97.26\%(\std{0.09}) accuracy, which is competitive with leading NAS approaches, e.g.~\cite{liu2018darts,zhang2018graph,chen2019progressive,yang2020cars,he2020milenas,li2020sgas}. In contrast, the network chosen by the regression model trained on the \ghnbase~embeddings achieves 95.90\%(\std{0.08}).\looseness-1

	\vspace{-3pt}
	
	\vspace{-2pt}
	\subsection{Fine-tuning Predicted Parameters\label{sec:finetune}}
	\vspace{-2pt}
	
	Neural networks trained on ImageNet and other large datasets have proven useful in diverse visual tasks in the transfer learning setup~\cite{kornblith2019better,huh2016makes,neyshabur2020being,raghu2019transfusion,zhai2019large,dosovitskiy2020image}. 
	Therefore, we explore how predicting parameters on ImageNet with GHNs compares to pretraining them on ImageNet with SGD in such a setup. 
	We consider low-data tasks as they often benefit more from transfer learning~\cite{raghu2019transfusion,zhai2019large}.
	
	\textbf{Experimental setup.}
	We perform two transfer-learning experiments. The first experiment is fine-tuning the predicted parameters on 1,000 training samples (100 labels per class) of CIFAR-10. 
	We fine-tune ResNet-50, Visual Transformer (ViT) and a 14-cell architecture based on the DARTS best cell~\cite{liu2018darts}. The hyperparameters of fine-tuning (initial learning rate and weight decay) are tuned on 200 validation samples held-out of the 1,000 training samples. The number of epochs is fixed to 50 as in \S~\ref{sec:our_task} for simplicity.
	In the second experiment, we fine-tune the predicted parameters on the object detection task. We closely follow the experimental protocol and hyperparameters from~\cite{pytorchdetection} and train the networks on the Penn-Fudan dataset~\cite{wang2007object}. The dataset contains only 170 images and the task is to detect pedestrians. Therefore this task is also well suited for transfer learning. Following \cite{pytorchdetection}, we replace the backbone of a Faster R-CNN with one of the three architectures.
	To perform transfer learning with GHNs, in both experiments we predict the parameters of a given architecture using GHNs trained on ImageNet. 
	We then replace the ImageNet classification layer with the target task-specific layers and fine-tune the entire network on the target task.
	We compare the results of GHNs to He’s initialization~\cite{he2015delving} and the initialization based on pretraining the parameters on ImageNet with SGD.\looseness-1

	\begin{table}[htbp]
		\vspace{-5pt}
		\centering
		\scriptsize
		\caption{\small CIFAR-10 test set accuracies and Penn-Fudan object detection average precision (at IoU=0.50) after fine-tuning the networks using SGD initialized with different methods. Average results and standard deviations for 3 runs with different random seeds are shown. For each architecture, similar GHN-2-based and ImageNet-based results are bolded.\textsuperscript{*}Estimated on ResNet-50.}
		\label{tab:finetune}
		\vspace{3pt}
		\setlength{\tabcolsep}{4.1pt}
		\begin{tabular}{lcccc|ccc}
			\toprule
			
			\multirow{2}{*}{\textbf{\textsc{\parbox{2cm}{Initialization Method}}}} & \multirow{2}{*}{\textbf{{\parbox{0.9cm}{GPU sec. to init.\textsuperscript{*}}}}} & \multicolumn{3}{c|}{\textbf{\textsc{100-Shot Cifar-10}}} &
			\multicolumn{3}{c}{\textbf{\textsc{Penn-Fudan Object Detection}}}\Bstrut\\
			
			\cline{3-5}\cline{6-8}
			
			& & {\textsc{ResNet-50}} & {\textsc{ViT}} & {\textsc{Darts}} & {\textsc{ResNet-50}} & {\textsc{ViT}} & {\textsc{Darts}}\Tstrut\\
			
			\midrule
			
			He’s \cite{he2015delving} & 0.003 & 41.0\sem{0.4} & 33.2\sem{0.3} &	45.4\sem{0.4} & 0.197\sem{0.042} & 0.144\sem{0.010} &	0.486\sem{0.035}\Tstrut\\
			
			GHN-1 (trained on ImageNet) & 0.6 &	46.6\sem{0.0} &	23.3\sem{0.1} & 49.2\sem{0.1} & 0.433\sem{0.013} &	0.0\sem{0.0} & 0.468\sem{0.024} \\
			
			GHN-2 (trained on ImageNet) & 0.7 & \textbf{56.4}\sem{0.1} & \textbf{41.4}\sem{0.6} & \textbf{60.7}\sem{0.3} & \textbf{0.560}\sem{0.019} & \textbf{0.436}\sem{0.032} &	\textbf{0.785}\sem{0.032} \\
			
			\midrule
			
			ImageNet (1k pretraining steps) & $6\PLH 10^{2}$ & 45.4\sem{0.3} &	\textbf{44.3}\sem{0.1} & \textbf{62.4}\sem{0.3} & 0.302\sem{0.022} & 0.182\sem{0.046} &	\textbf{0.814}\sem{0.033}\Tstrut\\
			
			ImageNet (2.5k pretraining steps) & $1.5\PLH 10^{3}$ &	\textbf{55.4}\sem{0.2} &	50.4\sem{0.3} &	70.4\sem{0.2} & \textbf{0.571}\sem{0.056} &	0.322\sem{0.073} &	0.823\sem{0.022}\\
			
			ImageNet (5 pretraining epochs) & $3\PLH 10^{4}$ & 84.6\sem{0.2} & 70.2\sem{0.5} &	83.9\sem{0.1} & 0.723\sem{0.045} &	{0.391}\sem{0.024} &	0.827\sem{0.053} \\
			
			ImageNet (final epoch) & $6\PLH 10^{5}$ &	89.2\sem{0.2} &	74.5\sem{0.2} &	85.6\sem{0.2} & 0.876\sem{0.011} &	\textbf{0.468}\sem{0.023} &	0.881\sem{0.023}\\
			
			\bottomrule
		\end{tabular}
		\vspace{-5pt}
	\end{table}
	
	\textbf{Results.} The CIFAR-10 image classification results of fine-tuning the parameters predicted by our GHN-2 are $\geq$10 percentage points better (in absolute terms) than fine-tuning the parameters predicted by GHN-1 or training the parameters initialized using He’s method (Table~\ref{tab:finetune}).
	Similarly, the object detection results of GHN-2-based initialization are consistently better than both GHN-1 and He’s initializations. The GHN-2 results are a factor of 1.5-3 improvement over He’s for all the three architectures. Overall, the two experiments clearly demonstrate the practical value of predicting parameters using our GHN-2.
	Using GHN-1 for initialization provides relatively small gains or hurts convergence (for ViT).
	Compared to pretraining on ImageNet with SGD, initialization using GHN-2 leads to performance similar to 1k-2.5k steps of pretraining on ImageNet depending on the architecture in the case of CIFAR-10. In the case of Penn-Fudan, GHN-2’s performance is similar to $\geq$1k steps of pretraining with SGD. In both experiments, pretraining on ImageNet for just 5 epochs provides strong transfer learning performance and the final ImageNet checkpoints are only slightly better, which aligns with previous works~\cite{neyshabur2020being}. 
	Therefore, further improvements in the parameter prediction models appear promising.\looseness-1

	\vspace{-10pt}
	\section{Related Work\label{sec:related}}
	\vspace{-10pt}
	
	Our proposed parameter prediction task,  objective in Equation~\ref{eq:solution}  and improved GHN are related to a wide range of machine learning frameworks, in particular meta-learning and neural architecture search (NAS). Meta-learning is a general framework~\cite{hospedales2020meta,schmidhubermetalearning} that includes meta-optimizers and meta-models, among others. Related NAS works include differentiable~\cite{liu2018darts} and one-shot methods~\cite{cai2019onceforall}. See additional related work in \S~\ref{apdx:related_work}.\looseness-1
	
	\textbf{Meta-optimizers.} Meta-optimizers~\cite{andrychowicz2016learning,ravi2016optimization,metz2020tasks,kirsch2020meta,gomes2021meta} define a problem similar to our task, but where $H_\domain$ is an RNN-based model predicting the gradients $\nabla \w$, mimicking the behavior of iterative optimizers. Therefore, the objective of meta-optimizers may be phrased as \emph{learning to optimize} as opposed to our \emph{learning to predict} \params.
	Such meta-optimizers can have their own hyperparameters that need to be tuned for a given architecture $\f$ and need to be run expensively (on the GPU) for many iterations following Equation~\ref{eq:optim1b}.\looseness-1

	\textbf{Meta-models.} Meta-models include methods based on MAML~\cite{finn2017model}, ProtoNets~\cite{snell2017prototypical} and auxiliary nets predicting task-specific parameters~\cite{romero2016diet,requeima2019fast,li2019lgm,bertinetto2016learning}. These methods are tied to a particular architecture and need to be trained from scratch if it is changed.
	Several recent methods attempt to relax the choice of architecture in meta-learning. T-NAS~\cite{lian2020towards} combines MAML with DARTS~\cite{liu2018darts} to learn both the optimal architecture and its parameters for a given task. However, the best network, $\f$, needs to be trained using MAML from scratch. Meta-NAS~\cite{elsken2020meta} takes a step further and only requires fine-tuning of $\f$ on a given task. However, the $\f$ is obtained from a single meta-architecture and so its choice is limited, preventing parameter prediction for arbitrary $\f$. CATCH~\cite{chen2020catch} follows a similar idea, but uses reinforcement learning to quickly search for the best $\f$ on the specific task.
	Overall meta-learning mainly aims at generalization \textit{across tasks}, often motivated by the few-shot learning problem. In contrast, our parameter prediction problem assumes a single task (here an image dataset), but aims at generalization \textit{across architectures} $\f$ with the ability to predict parameters in a single forward pass.\looseness-1

	\textbf{One-shot NAS.} One-shot NAS aims to learn a single ``supernet''~\cite{yu2020bignas,cai2019onceforall,he2021automl} that can be used to estimate the performance of smaller nets (subnets) obtained by some kind of pruning the supernet, followed by training the best chosen $\f$ from scratch with SGD.
	Recent models, in particular BigNAS~\cite{cai2019onceforall} and OnceForAll (OFA)~\cite{yu2020bignas}, eliminate the need to train subnets. However, the fundamental limitation of one-shot NAS is poor scaling with the number of possible computational operations~\cite{zhang2018graph}. This limits the diversity of architectures for which \params can be obtained. For example, all subnets in OFA are based on MobileNet-v3~\cite{howard2019searching}, which does not allow to solve our more general parameter prediction task.
	To mitigate this, SMASH~\cite{brock2017smash} proposed to predict some of the \params using hypernetworks~\cite{ha2016hypernetworks} by encoding architectures as a 3D tensor.
	Graph HyperNetworks (GHNs)~\cite{zhang2018graph} further generalized this approach to ``arbitrary'' computational graphs (DAGs), which allowed them to improve NAS results.
	GHNs focused on obtaining reliable subnetwork rankings for NAS and did not aim to predict large-scale performant parameters.
	We show that the vanilla GHNs perform poorly on
	our parameter prediction task mainly due to the inappropriate scale of predicted parameters, lack of long-range interactions in the graphs, gradient noise and slow convergence when optimizing Equation~\ref{eq:solution}.
	Conventionally to NAS, GHNs were also trained in a quite constrained architecture space~\cite{bender2018understanding}. We expand the architecture space adopting GHNs for a more general problem.\looseness-1

	\vspace{-5pt}
	\section{ Conclusion}
	\vspace{-10pt}
	We propose a novel framework and benchmark to learn and evaluate neural parameter prediction models. Our model (\ghnours) is able to predict \params for very diverse and large-scale architectures in a single forward pass in a fraction of a second. The networks with predicted \params yield surprisingly high image classification accuracy given the extremely challenging nature of our parameter prediction task. However, the accuracy is still far from networks trained with handcrafted optimization methods. Bridging the gap is a promising future direction. As a beneficial side-effect, \ghnours learns a strong representation of neural architectures as evidenced by our property prediction evaluation. Finally, parameters predicted using \ghnours trained on ImageNet benefit transfer learning in the low-data regime. This motivates further research towards solving our task.\looseness-1

	\section*{Acknowledgments}
	BK is thankful to Facebook AI Research for funding the initial phase of this research during his internship and to NSERC and the Ontario Graduate Scholarship used to fund the other phases of this research.
	GWT and BK also acknowledge support from CIFAR and the Canada Foundation for Innovation.
	Resources used in preparing this research were provided, in part, by the Province of Ontario, the Government of Canada through CIFAR, and companies sponsoring the Vector Institute: \url{http://www.vectorinstitute.ai/#partners}.
	We are thankful to Magdalena Sobol for editorial help. We are thankful to the Vector AI Engineering team (Gerald Shen, Maria Koshkina and Deval Pandya) for code review. We are also thankful to the reviewers for their constructive feedback.\looseness-1

	\normalsize
	\medskip
	
	{\small
		\begin{spacing}{0.99}
			\bibliographystyle{unsrtnat}
			\bibliography{ref}
		\end{spacing}	
	}

	\vfill
			
	\newpage
	\normalsize

\appendix
\setlength\cftbeforesubsecskip{-1pt}
\renewcommand\cftsecafterpnum{\vskip0pt}

\part{Appendix}\label{sec:apdx}

\vspace{-10pt}
\begin{spacing}{0.2}
	\parttoc % Insert the appendix TOC
\end{spacing}

\vspace{-3pt}
\section{\dataset Details \label{apdx:dataset}}
\vspace{-5pt}

\subsection{Generating \dataset using DARTS\label{apdx:darts_bg}}
\vspace{-3pt}

\begin{figure}[tbhp]
	\vspace{-10pt}
	\centering
	\begin{tabular}{ccc}
		\includegraphics[width=0.12\textwidth,trim={0 0 0 0},clip,align=c]{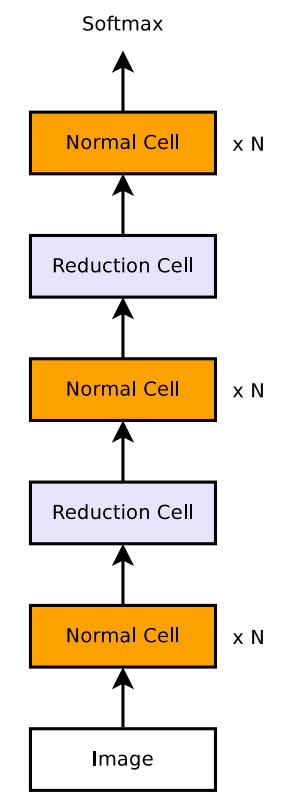}
		& 
		\includegraphics[width=0.3\textwidth,trim={0 0 0 0},clip,align=c]{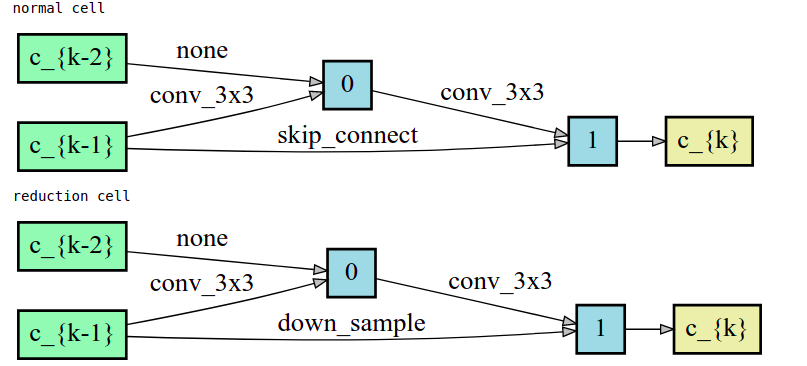} 
		& 
		\includegraphics[width=0.25\textwidth,trim={2cm 2cm 2cm 2cm},clip,align=c]{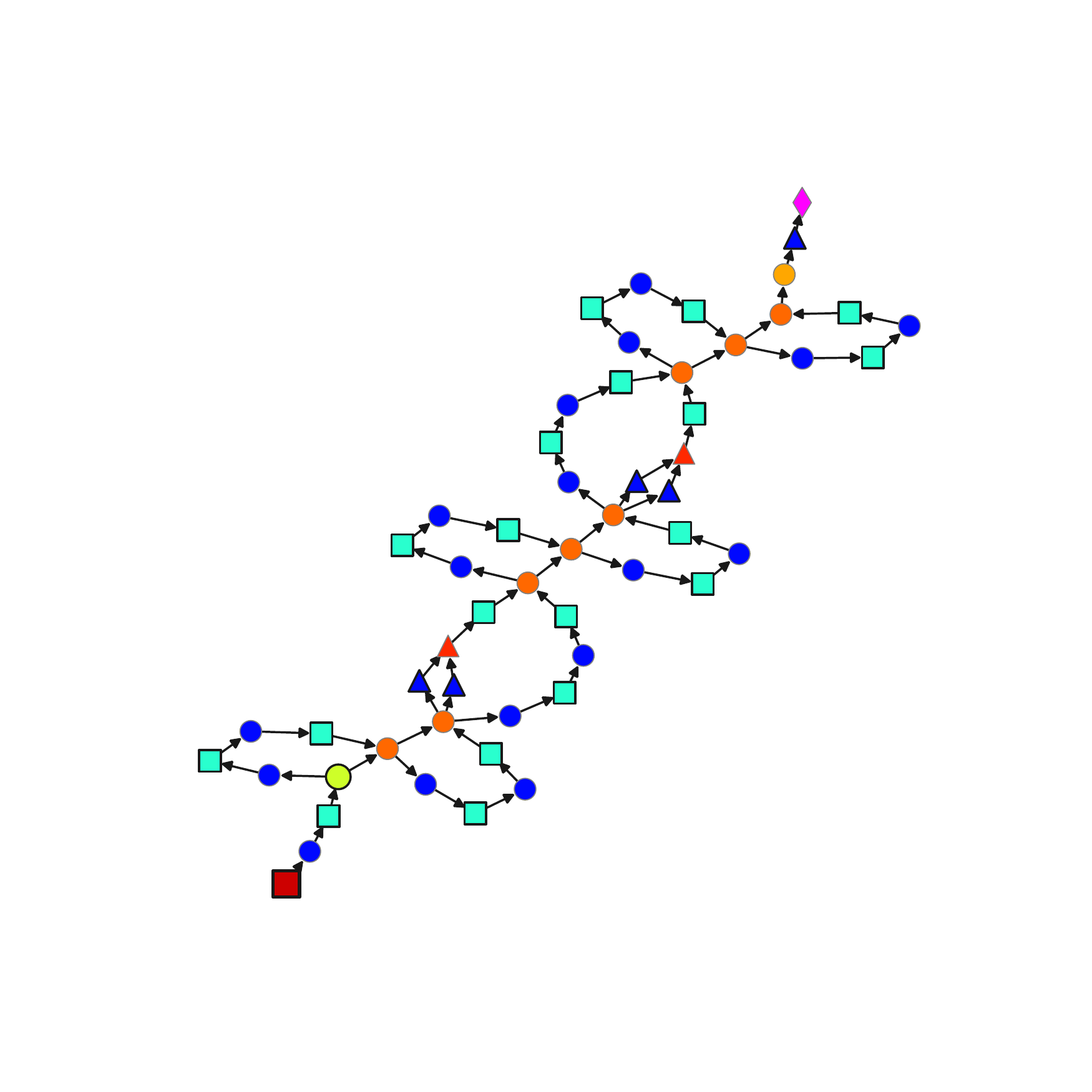} \\
		(a) & (b) & (c) \\
		& \includegraphics[width=0.3\textwidth,trim={0 0 0 0},clip,align=c]{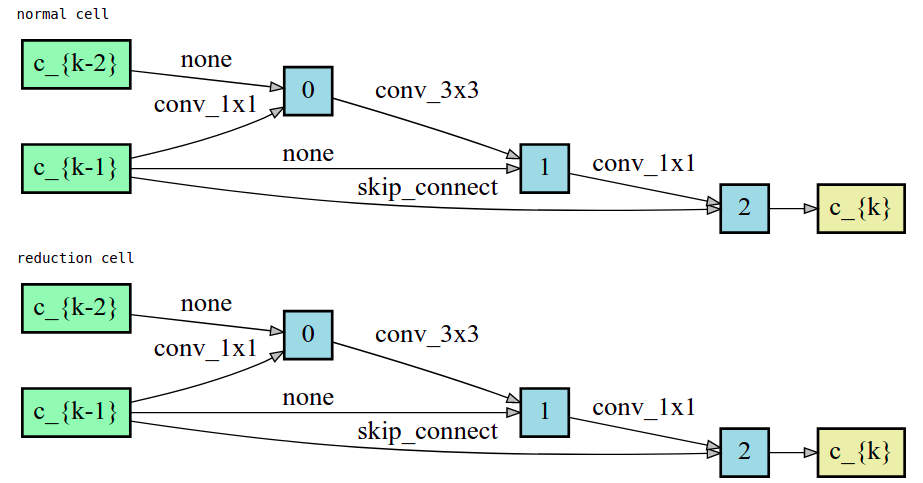} 
		& \includegraphics[width=0.25\textwidth,trim={2cm 2cm 2cm 2cm},clip,align=c]{dag_resnet_50.pdf} \vspace{-15pt}\\
		& (d) & (e) \\
	\end{tabular}
	\vspace{-8pt}
	\caption{\small \textbf{(a)} Network's high-level structure introduced in \cite{zoph2018learning} and employed by many following papers on network design, including DARTS~\cite{liu2018darts} and ours, where N$\geq 1$; \textbf{(b)} A residual block~\cite{he2016deep} in terms of DARTS normal and reduction cells, where green nodes denote outputs from the two previous cells, blue nodes denote summation, a yellow node denotes concatenation$^\dagger$; edges denote operations, `none' indicates dropping the edge$^\ddagger$; the reduction cell has the same structure in ResNets, but decreases spatial resolution by 2 using a downsample operation and stride 2 in operations, at the same time, optionally increasing the channel dimensionality by 2. \textbf{(c)} The result of combining (a) and (b) for 8 cells using our code to build an analogous of the ResNet-18 architecture$^\star$.
		\textbf{(d)} A residual block of ResNet-50 with $1 \PLH 1$ bottleneck layers defined using DARTS and \textbf{(e)} the graph built using our code, where 3 reduction cells are placed as in the original ResNet-50 architecture. }
	\label{fig:darts_bg}
	\vspace{-20pt}
\end{figure}

In this section, we elaborate on our description in \S~\ref{sec:bg_darts} about how DARTS~\cite{liu2018darts} defines networks. We also elaborate on our discussion in \S~\ref{sec:dataset} about how we modify the DARTS framework to generate our \dataset dataset of architectures and summarize these modifications here, in Table~\ref{tab:darts_diff}.
We visualize examples of architectures defined using DARTS~\cite{liu2018darts} and corresponding computational graphs obtained using our code (Fig.~\ref{fig:darts_bg}). 

\textbf{Overall architecture structure.} At a high level, all our \iid and \ood networks are composed of a stem, repeated normal and reduction cells, global average pooling and a classification head (Fig.~\ref{fig:darts_bg}, (a)). We optionally sample fully-connected layers between the global pooling and the last classification layer and/or replace global pooling with fully connected layers, e.g. as in VGG~\cite{simonyan2014very}. The stem in DARTS, and in other NAS works, is predefined and fixed for each image dataset. We uniformly sample either a CIFAR-10 style or ImageNet style stem, so that our network space is unified for both image datasets. To prevent extreme GPU memory consumption when using a non-ImageNet stem for ImageNet images, we additionally use a larger stride in the stem that does not affect the graph structure.
\textbf{At test time}, however, we can predict \params for networks without these constraints, but the performance of the predicted \params might degrade accordingly. For example, we can successfully predict \params for ResNet-50 (Fig.~\ref{fig:darts_bg}, (e)), which has 13 normal and 3 reduction cells placed after the 3rd, 7th and 13th cells. ResNet-50's cells (Fig.~\ref{fig:darts_bg}, (d)) are similar to those of ResNet-18 (Fig.~\ref{fig:darts_bg}, (b)), but have $1\PLH 1$ bottleneck layers.

\textbf{Within and between cell connectivity.}
Within each cell and between them, there is a certain pattern to create connections in DARTS (Fig.~\ref{fig:darts_bg}, (b,d)): each cell receives features from the two previous cells, each summation node can only receive features from two nodes, the last concatenation node can receive features from an arbitrary number of nodes. But due to the presence of the Zero (`none') and Identity (`skip connection') operations, we can enable any connectivity. 
We represent operations as nodes\footnote{In DARTS, the operations are placed on the edges, except for inputs, summations and concatenation (Fig.~\ref{fig:darts_bg}).\looseness-1} and drop redundant edges and nodes. For example, if the node performs the Zero operation, we remove the edges connected to that node. This can lead to a small fraction of disconnected graphs, which we remove from the training/testing sets. If the node performs the Identity operation, we remove the node, but keep corresponding edges. 
We also omit ReLU operations and other nonlinearities in a graph representation to avoid significantly enlarging graphs, since the position of nonlinearities w.r.t. other operations is generally consistent across architectures (e.g. all convolutions are preceded by ReLUs except the first layer).
This leads to graphs visualized in Fig.~\ref{fig:darts_bg}, (c,e).

\textbf{Operations.} The initial choice of operations in DARTS is quite standard in NAS. In normal cells, each operation returns the tensor of the same shape as it receives. So any differentiable operation that can preserve the shape of the input tensor can be used, therefore extending the set of operations is relatively trivial. In reduction cells, spatial resolution is reduced by a factor of 2, so some operations can have stride 2 there. In both cells, there are summation and concatenation nodes that aggregate features from several operations into one tensor. Concatenation (across channels) is used as the final node in a cell. To preserve channel dimensions after concatenating many features, $1\PLH 1$ convolutions are used as needed.
For the Squeeze\&Exicte (SE) and Visual Transformer (ViT) operations, we use open source implementations\footnote{SE: \url{https://github.com/ai-med/squeeze_and_excitation/blob/master/squeeze_and_excitation/squeeze_and_excitation.py}, ViT: \url{https://github.com/lucidrains/vit-pytorch/blob/main/vit_pytorch/vit.py}} with default configurations, e.g. 8 heads in the multihead self-attention of ViT.\looseness-1

\blfootnote{$^\dagger$Concatenation is redundant in ResNets and removed from our graphs due to only one input node in cells.\looseness-1}
\blfootnote{$^\ddagger$In ResNets~\cite{he2016deep}, there is no skip connection between the input of a given cell and the output of the cell before the previous one.}
\blfootnote{$^\star$Note that ResNets of~\cite{he2016deep} commonly employed in practice have 3 reduction cells instead of 2 and have other minor differences (e.g. the order of convolution, BN and ReLU, down sampling convolution type, bottleneck layers, etc.). We still can predict \params for them, but such architectures would be further away from the training distribution, so the predicted \params might have significantly lower performance.\vspace{-15pt}}

\begin{table}[thbp]
	\centering
	\caption{\small Summary of differences between the DARTS design space and ours. \textsuperscript{*}Means implementation-wise possibility of predicting \params given a trained GHN, and does not mean our testing \iid architectures, which follow the training design protocol. Overall, from the implementation point of view, our trained GHNs allow to predict \params for arbitrary DAGs composed of our 15 primitives with the parameters of arbitrary shapes$^{\mathparagraph}$. We place ResNet-50 in a separate column even though it is one of the evaluation architectures of \dataset, because it has different properties as can be seen in the table.}
	\label{tab:darts_diff}
	\vspace{3pt}
	\setlength{\tabcolsep}{2.5pt}
	\small
	\begin{tabular}{lcccc}
		\toprule
		
		\textbf{\textsc{Property}} & \textbf{\textsc{DARTS}} & \textbf{\textsc{\dataset}} & \textbf{\textsc{ResNet-50}} & \textbf{\textsc{Testing GHN}}\textsuperscript{*} \\
		\midrule
		
		Unified style across image datasets & \xmark & \cmark & \xmark & \cmark \\
		
		VGG style classification heads~\cite{simonyan2014very} & \xmark & \cmark & \xmark & \cmark \\
		
		Visual Transformer stem~\cite{dosovitskiy2020image} & \xmark & \cmark & \xmark & \cmark \\
		
		Channel expansion ratio & 2 & 1 or 2 & 2 & arbitrary \\
		
		Bottleneck layers (e.g. in ResNet-50~\cite{he2016deep}) & \xmark & \xmark & \cmark & \cmark \\
		
		Reduction cells position (w.r.t. total depth) & 1/3, 2/3 & 1/3, 2/3 & 3,7,17 cells & arbitrary \\
		
		Networks w/o $1\PLH 1$ preprocessing layers in cells & \xmark & \cmark & \cmark & \cmark \\
		
		Networks w/o batch norm & \xmark & \cmark & \xmark & \cmark \\
		
		\bottomrule
	\end{tabular}
\end{table}

\blfootnote{$^{\mathparagraph}$While we predefine a wide range of possible shapes in our GHNs according to \S~\ref{apdx:ghn_1}, in the rare case of using the shape that is not one of the predefined values, we use the closest values, which worked reasonably well in many cases.}

\subsection{\dataset Statistics\label{apdx:stats}}

\begin{figure}[tbhp]
	\centering
	\small
	\setlength{\tabcolsep}{0pt}
	\begin{tabular}{cccc}
		\includegraphics[width=0.25\textwidth,align=c,trim={0 0 0 0},clip]{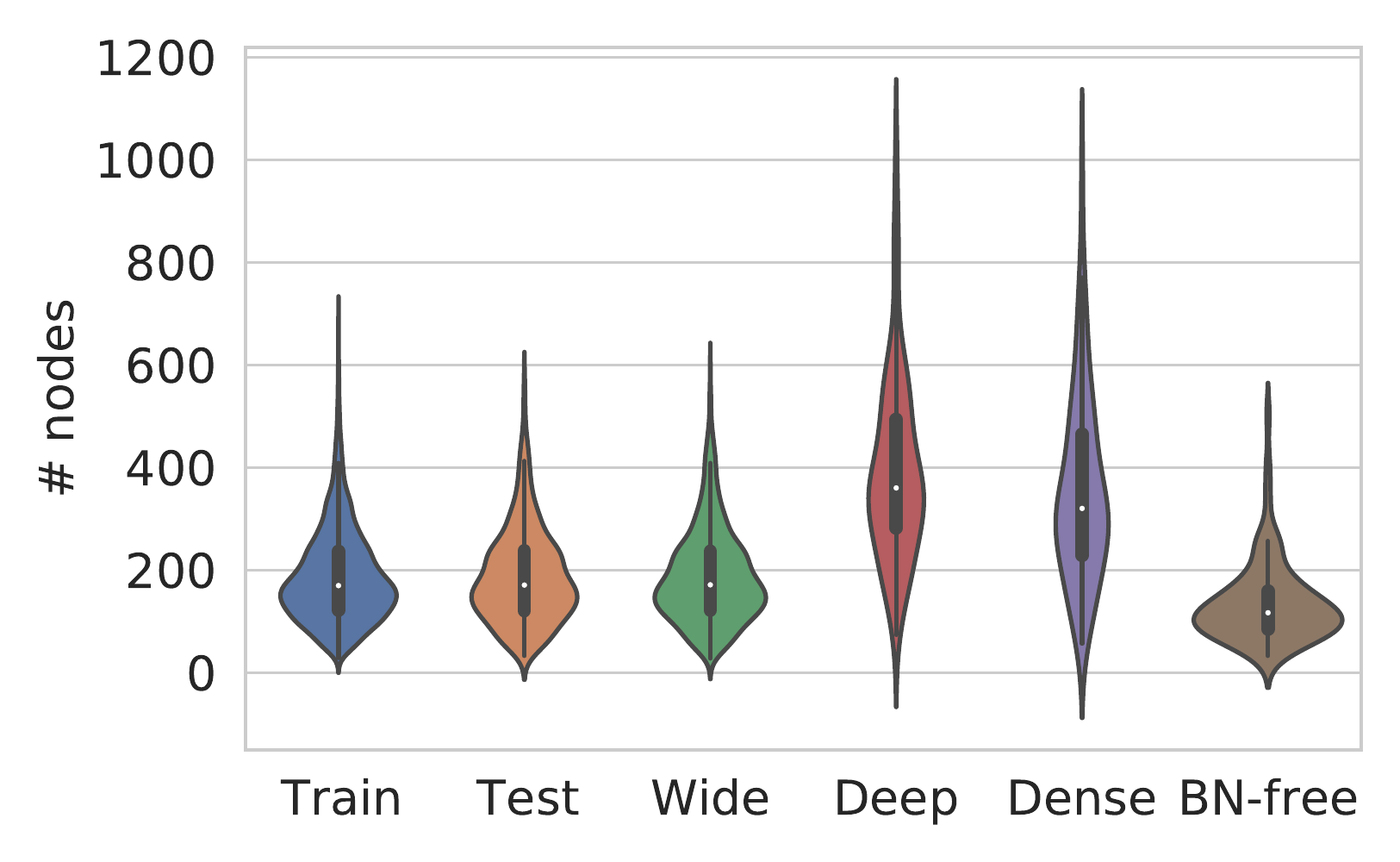} & \includegraphics[width=0.25\textwidth,align=c,trim={0 0 0 0},clip]{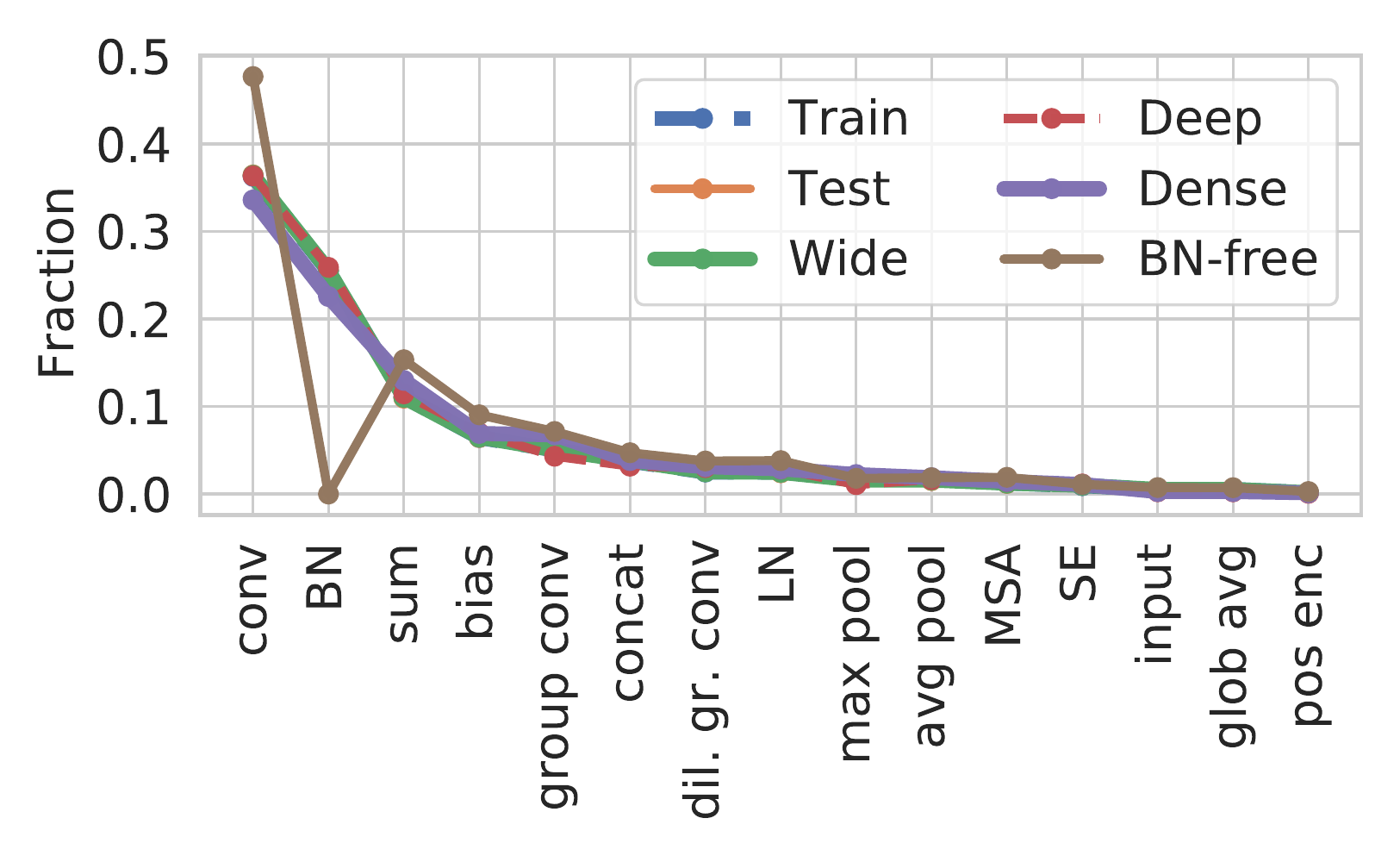} &
		\includegraphics[width=0.25\textwidth,align=c,trim={0 0 0 0},clip]{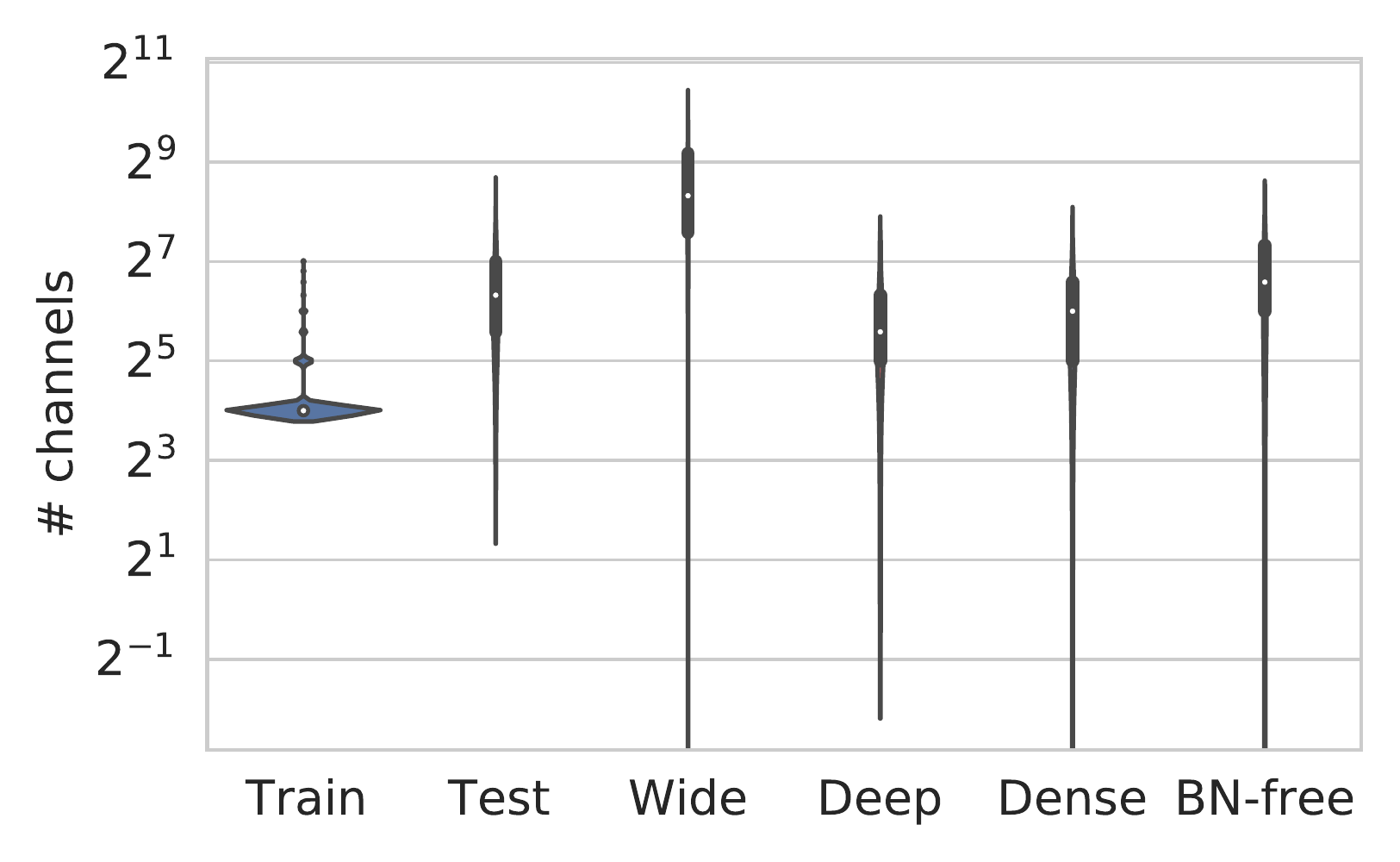} &
		\includegraphics[width=0.25\textwidth,align=c,trim={0 0 0 0},clip]{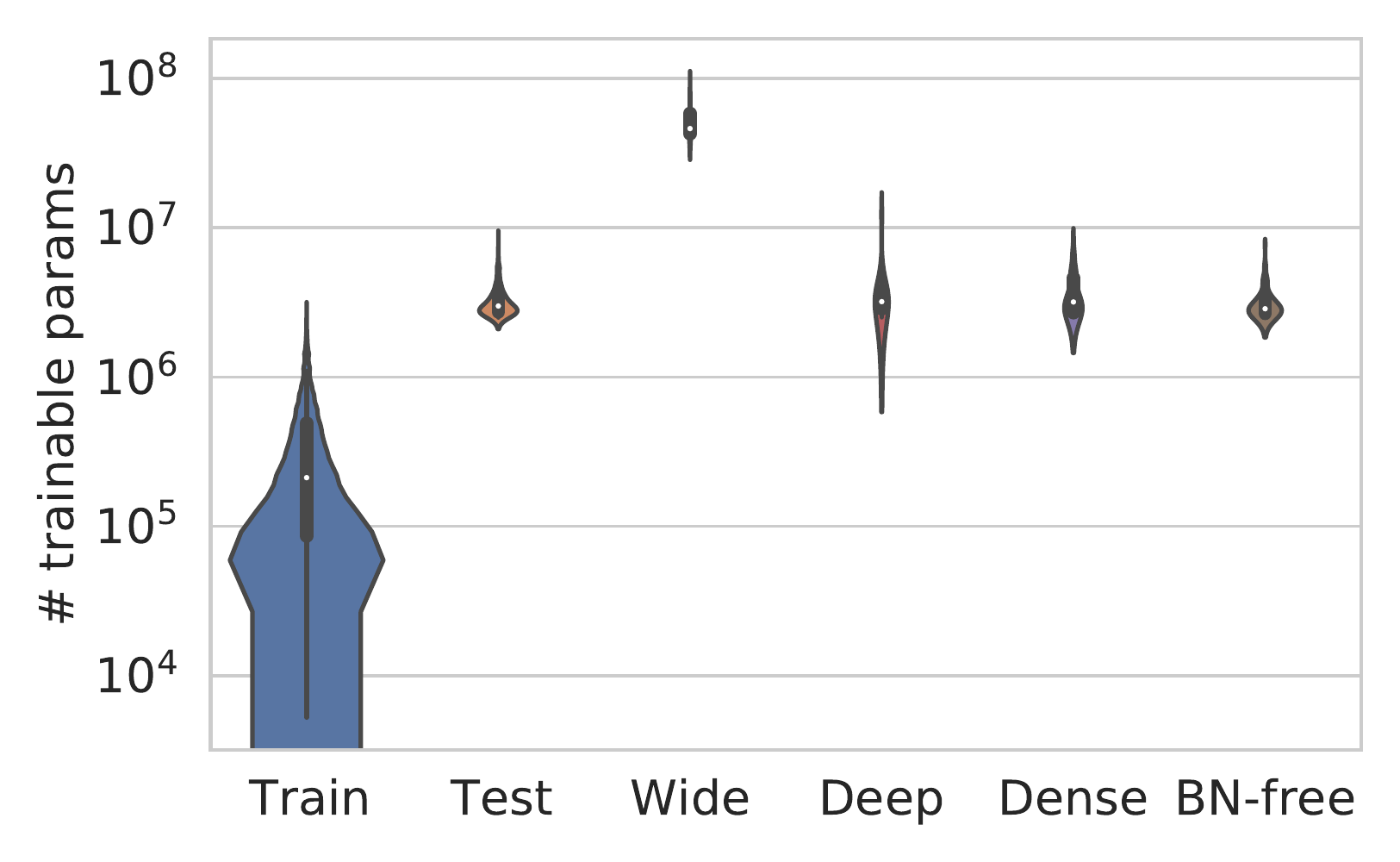}\\
		(a) & (b) & (c) & (d) \\
		\includegraphics[width=0.25\textwidth,align=c,trim={0 0 0 0},clip]{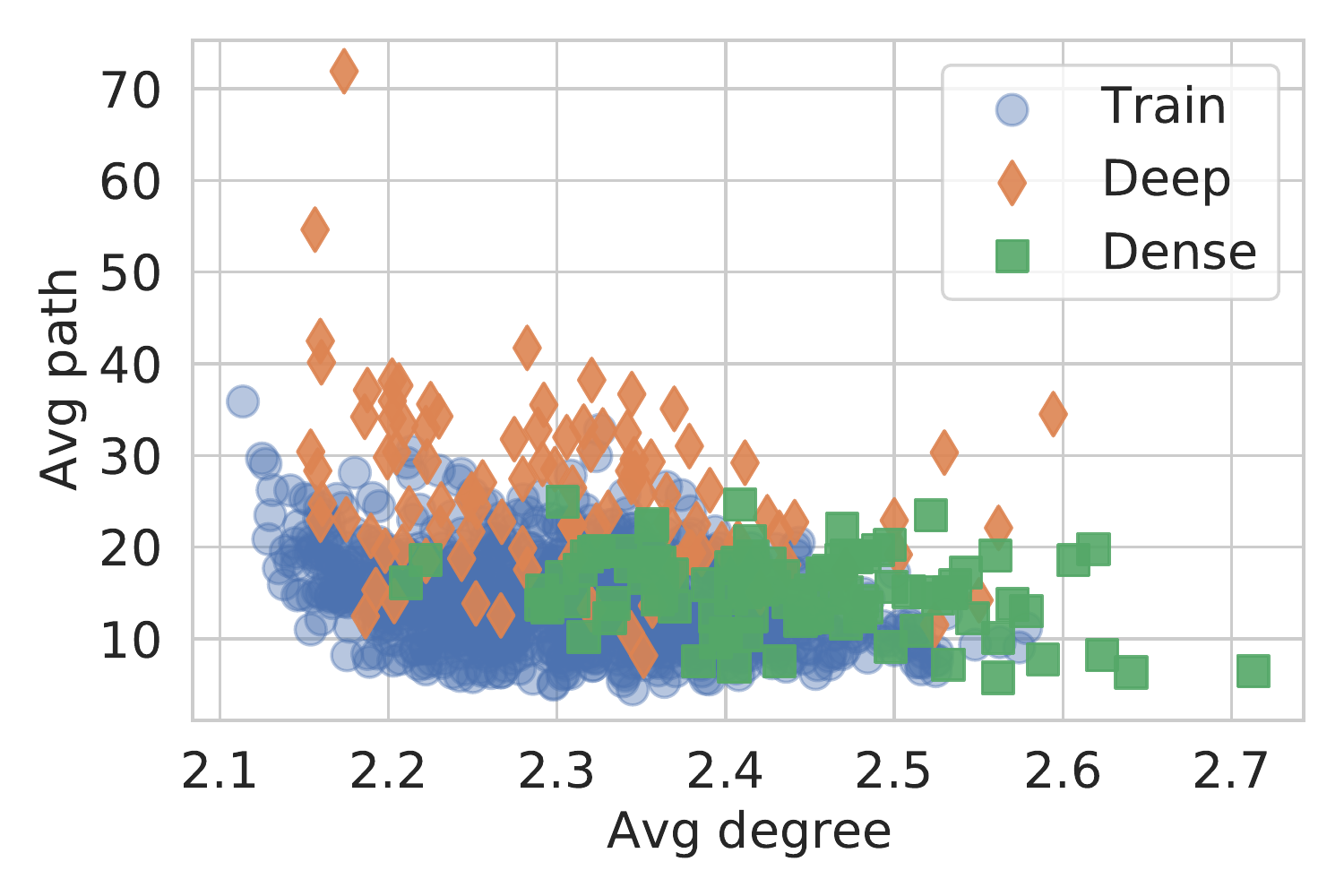} & 
		\includegraphics[width=0.25\textwidth,align=c,trim={0 0 0 0},clip]{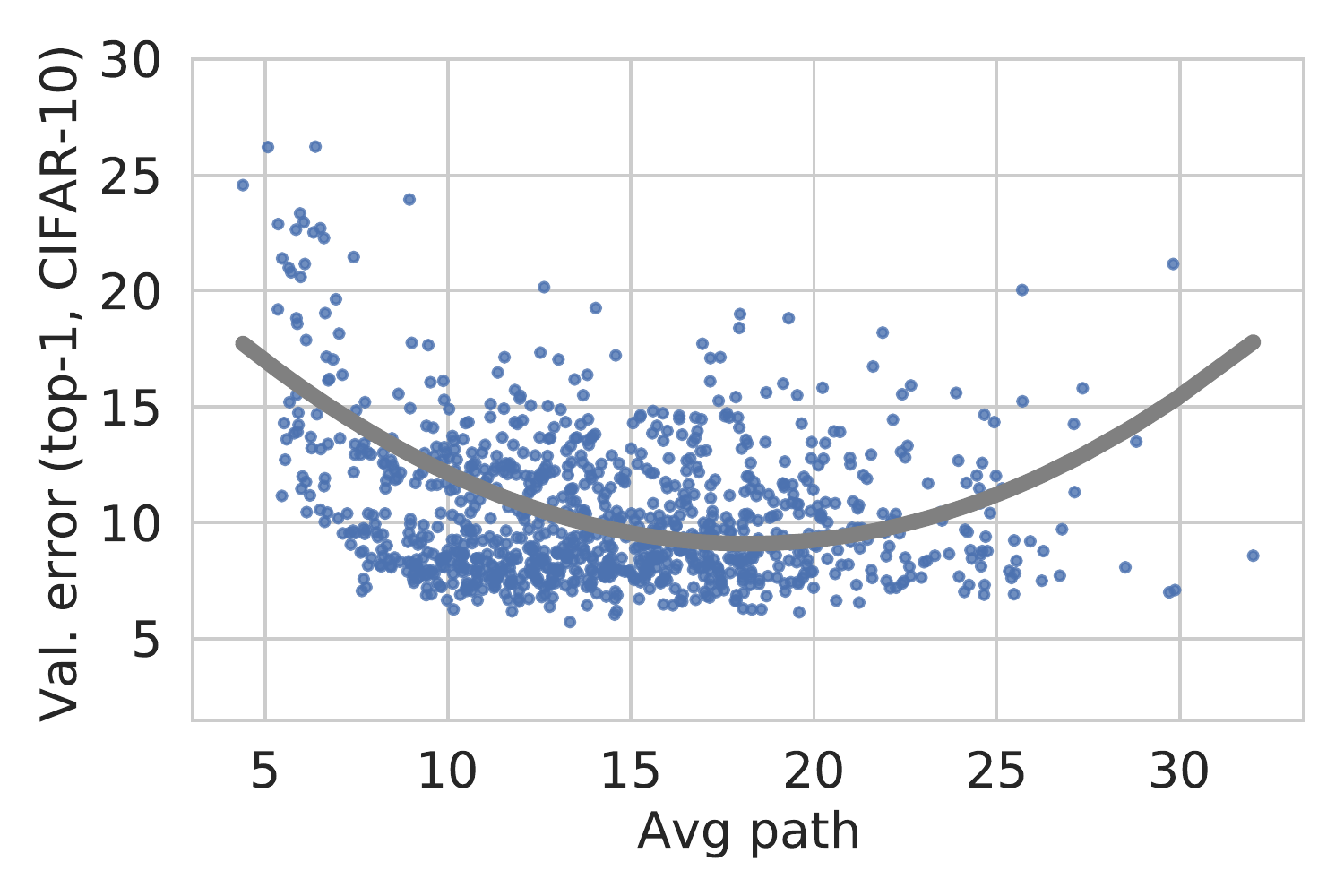} & 
		\includegraphics[width=0.25\textwidth,align=c,trim={0 0 0 0},clip]{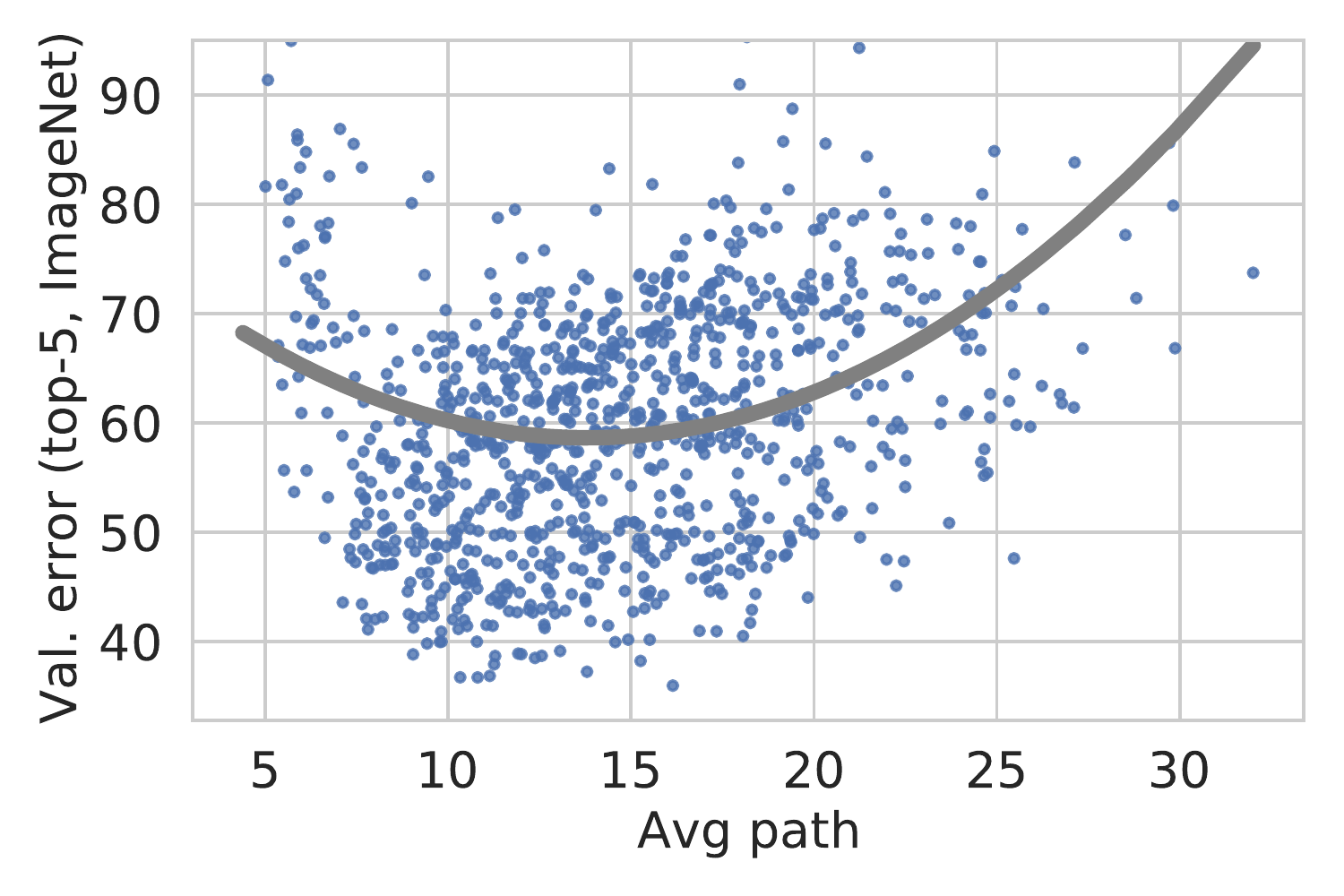} &
		\includegraphics[width=0.25\textwidth,align=c,trim={0 0 0 0},clip]{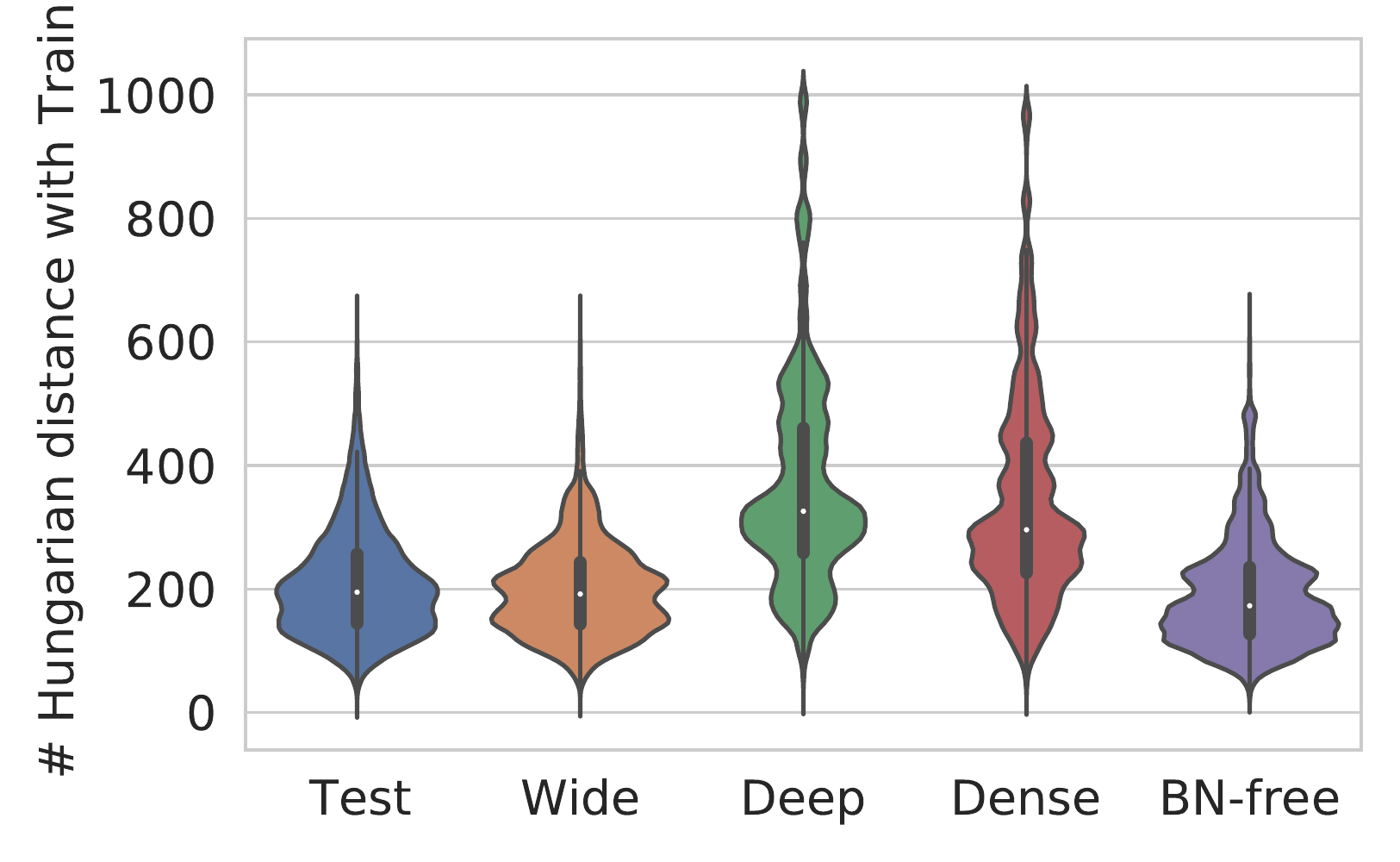}\\
		(e) & (f) & (g) & (h) \\
	\end{tabular}
	\vspace{-2pt}
	\caption{\small Visualized statistics of \dataset. (\textbf{a}) A violin plot of the number of nodes showing the distribution shift for the \deep and \dense subsets. (\textbf{b}) Node types (primitives) showing the distribution shift for the \bnfree subset. (\textbf{c}) Number of initial channels in networks (for \iidtrain, the number is for training GHNs on CIFAR-10). Here, the distribution shift is present for all test subsets (due to computational challenges of training GHNs on wide architectures), but the largest shift is for \wide. (\textbf{d}) Total numbers of trainable parameters in case of CIFAR-10, where the distribution shifts are similar to those for the number of channels. (\textbf{e}) Average shortest path length versus average node degree (other subsets that are not shown follow the distribution of \iidtrain), confirming that nodes of the \dense subset have generally more dense connections (larger degrees), while in \deep the networks are deeper (the shortest paths tend to be longer). (\textbf{f}) The validation error for 1000 \iidval+\iidtest architectures (trained with SGD for 50 epochs) versus their average shortest path lengths, indicating the ``sweet spot'' of architectures with strong performance (same axes as in~\cite{you2020graph}). (\textbf{g}) Same as ({f}), but with the y axis being the top-5 validation error on ImageNet of the same architectures trained for 1 epoch according to our experiments. (\textbf{h}) The distribution of the distances between the architectures of a given test subset and the ones from a subset of \iidtrain computed using the Hungarian algorithm~\cite{kuhn1955hungarian}, confirming that the evaluation architectures are different from the training ones.\looseness-1}
	\label{fig:vis_stats}
	\vspace{-2pt}
\end{figure}

\begin{figure}[thbp]
	\centering
	\newcommand{\width}{0.19\textwidth}
	\setlength{\tabcolsep}{0pt}
	\begin{tabular}{cp{0.2cm}cccc}
		\toprule
		\multicolumn{1}{c}{{ \textbf{\textsc{In-Distribution}}}} & &
		\multicolumn{4}{c}{{ \textbf{\textsc{Out-of-Distribution}}}}
		\Bstrut\\
		{\small \textbf{\iidtrain/\iidval/\iidtest}} & & {\small \textbf{\wide}} & {\small \textbf{\deep}} & {\small \textbf{\dense}} & {\small \textbf{\bnfree}} \\
		\cline{1-1}\cline{3-6} \\[-2ex]
		\multicolumn{1}{c}{{\includegraphics[width=\width,align=c,trim={2.3cm 3cm 2.3cm 3cm},clip]{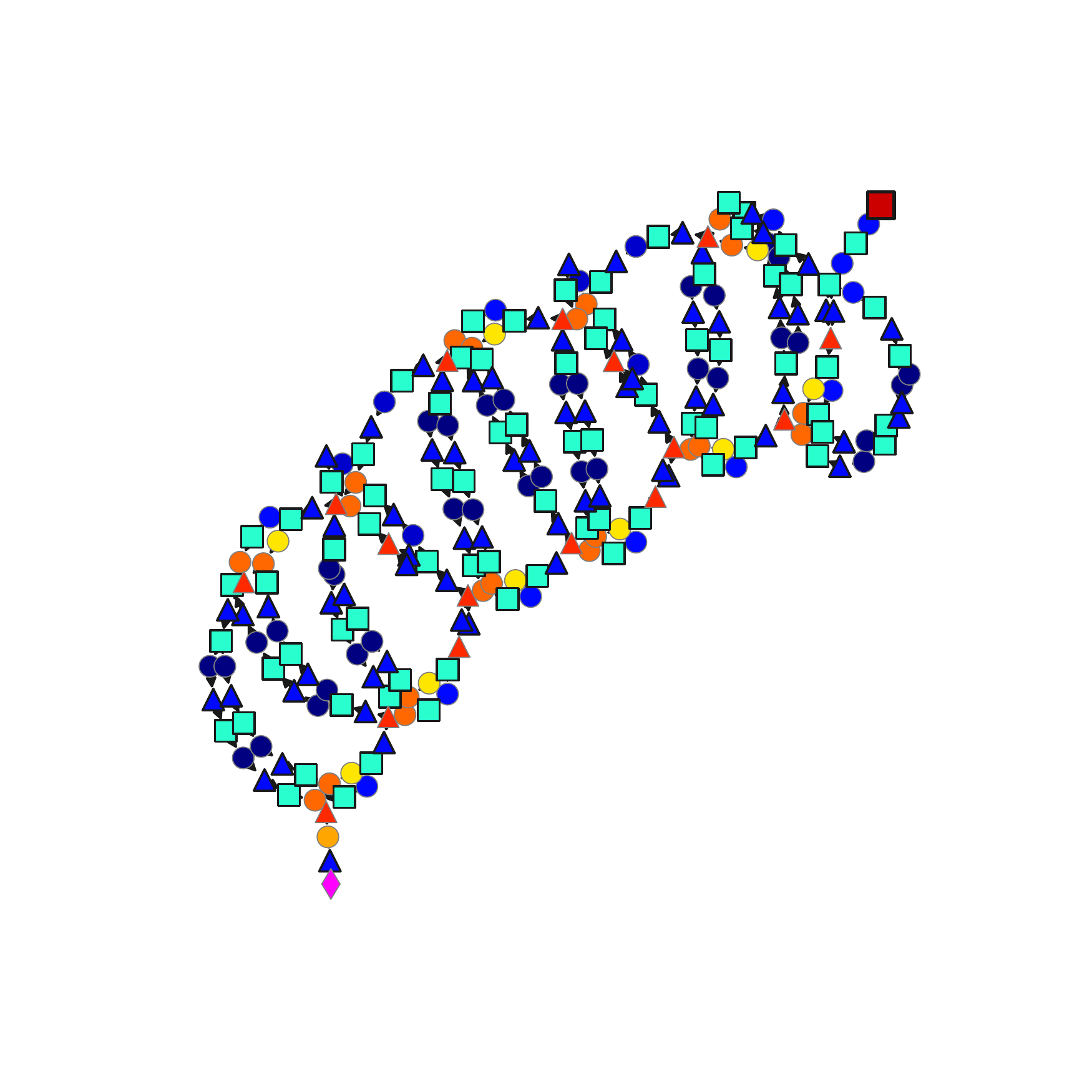}}} & & {\includegraphics[width=\width,align=c,trim={2.3cm 3cm 2.3cm 3cm},clip]{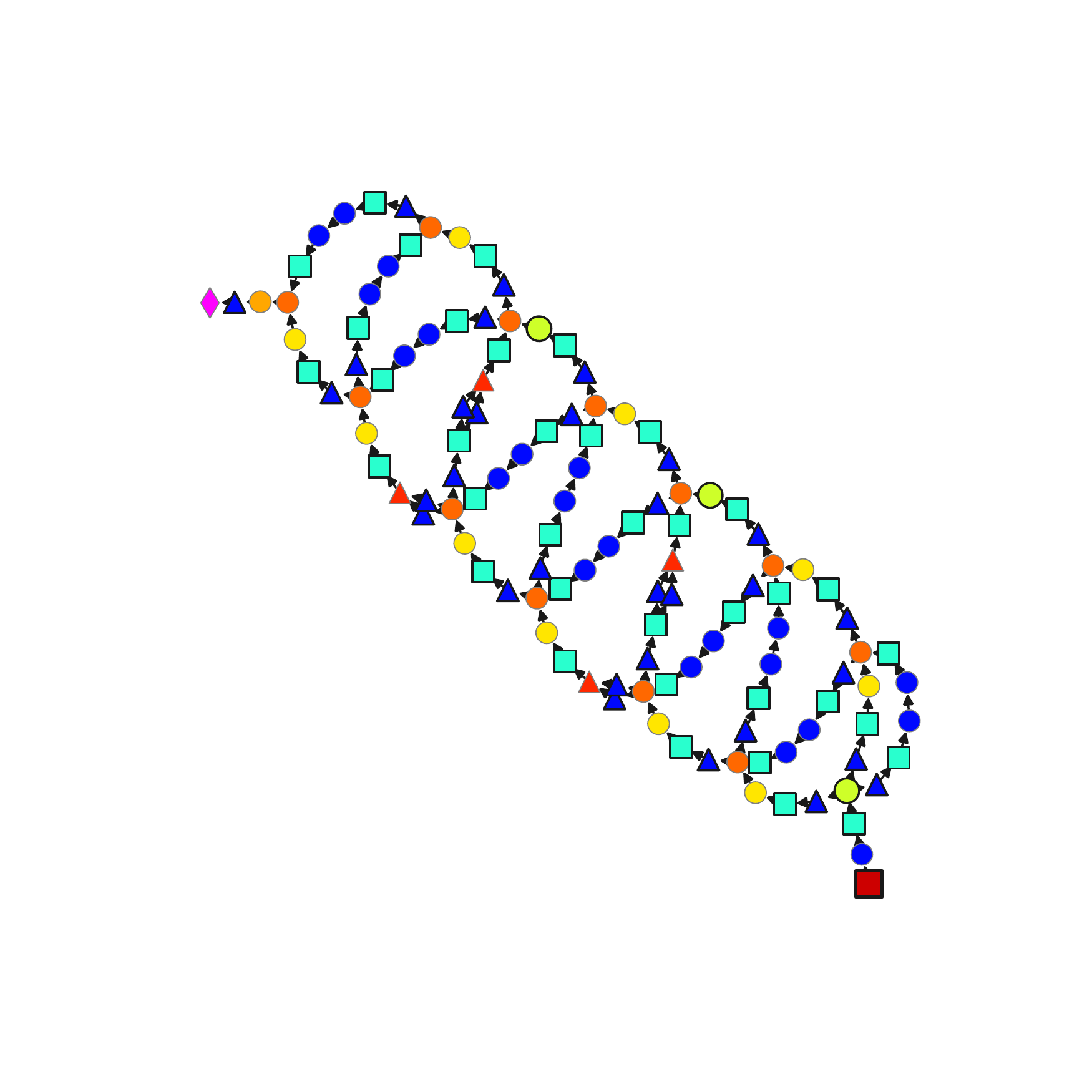}} & 
		{\includegraphics[width=\width,align=c,trim={2.3cm 3cm 2.3cm 3cm},clip]{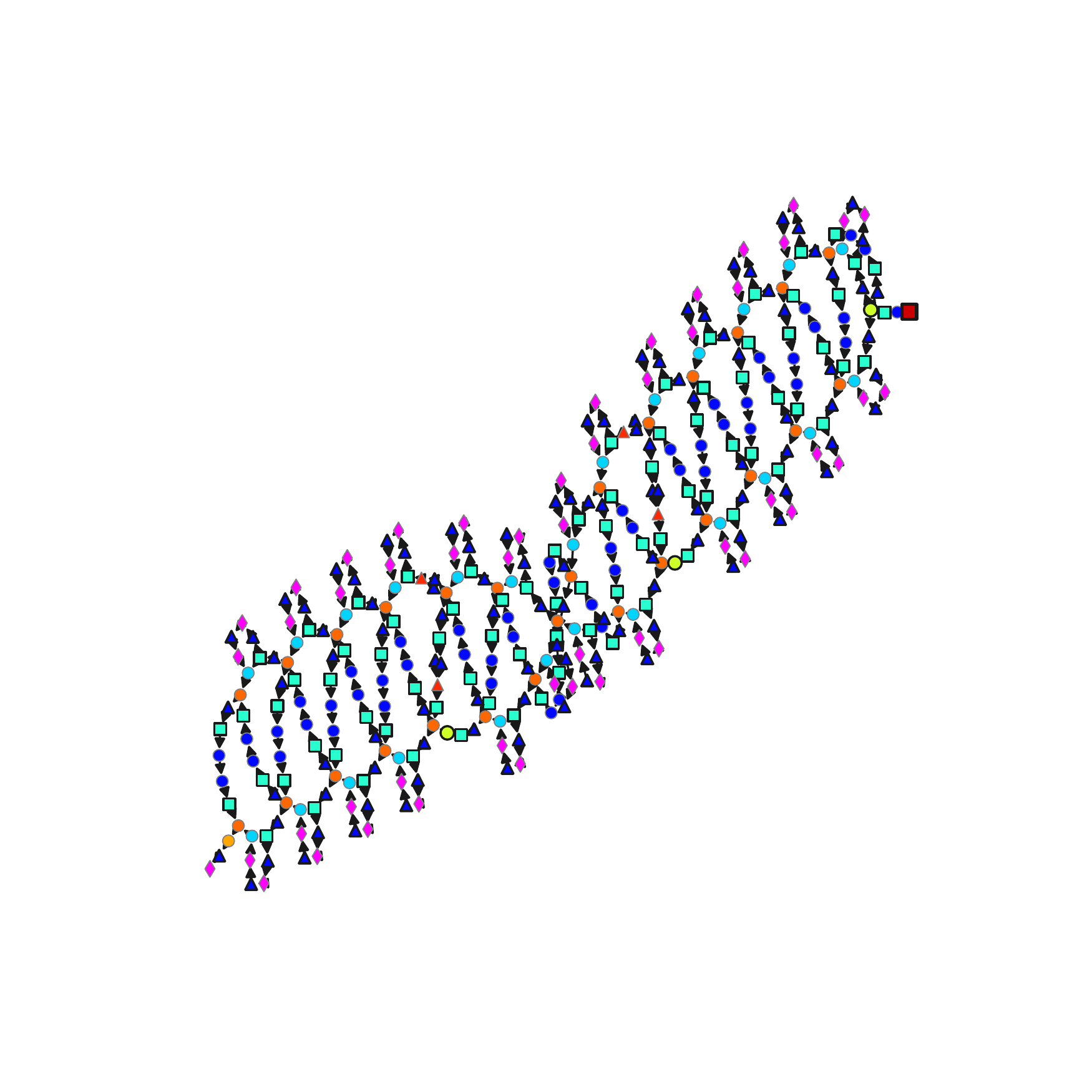}} & 
		\includegraphics[width=\width,align=c,trim={2.3cm 3cm 2.3cm 3cm},clip]{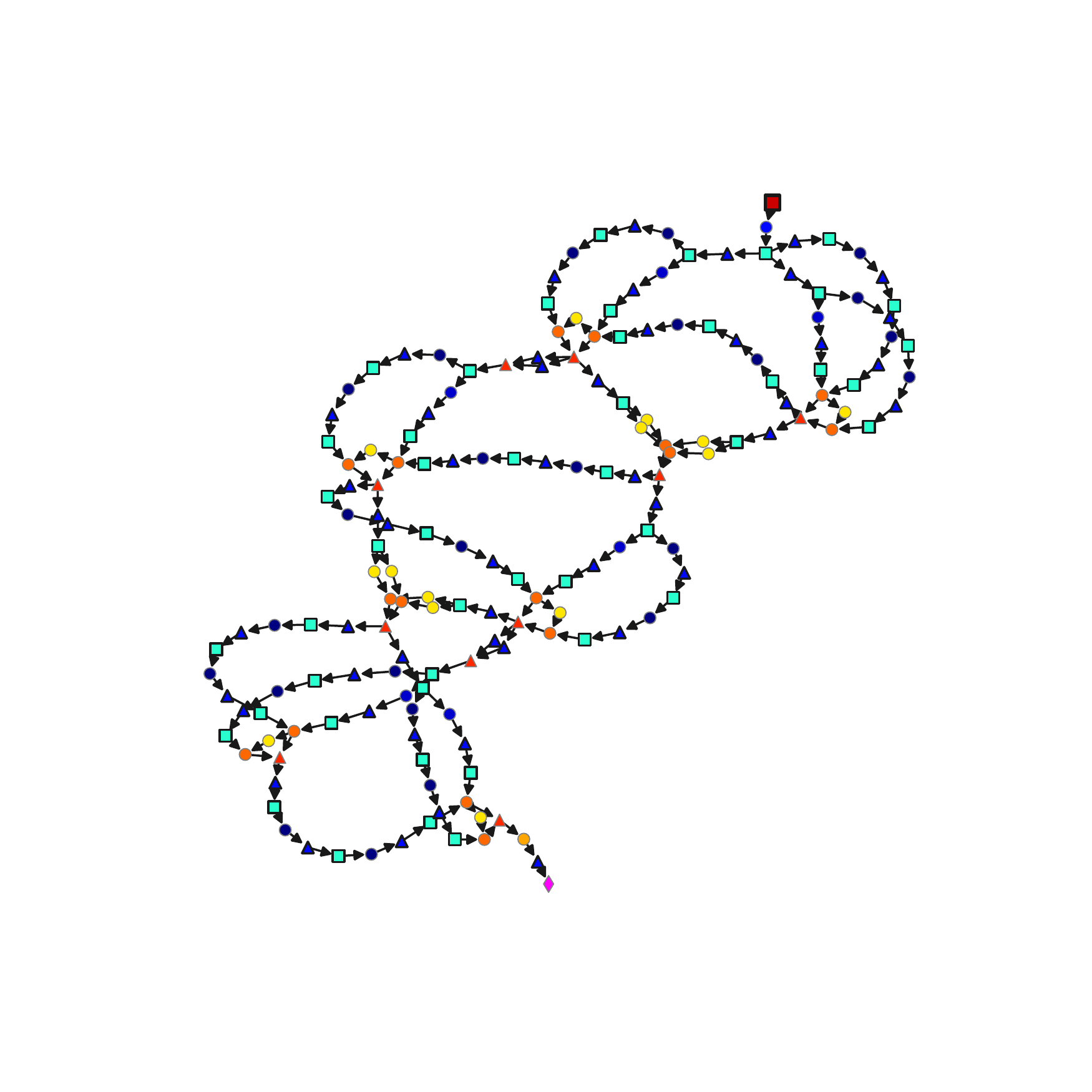} & 
		{\includegraphics[width=\width,align=c,trim={2.3cm 3cm 2.3cm 3cm},clip]{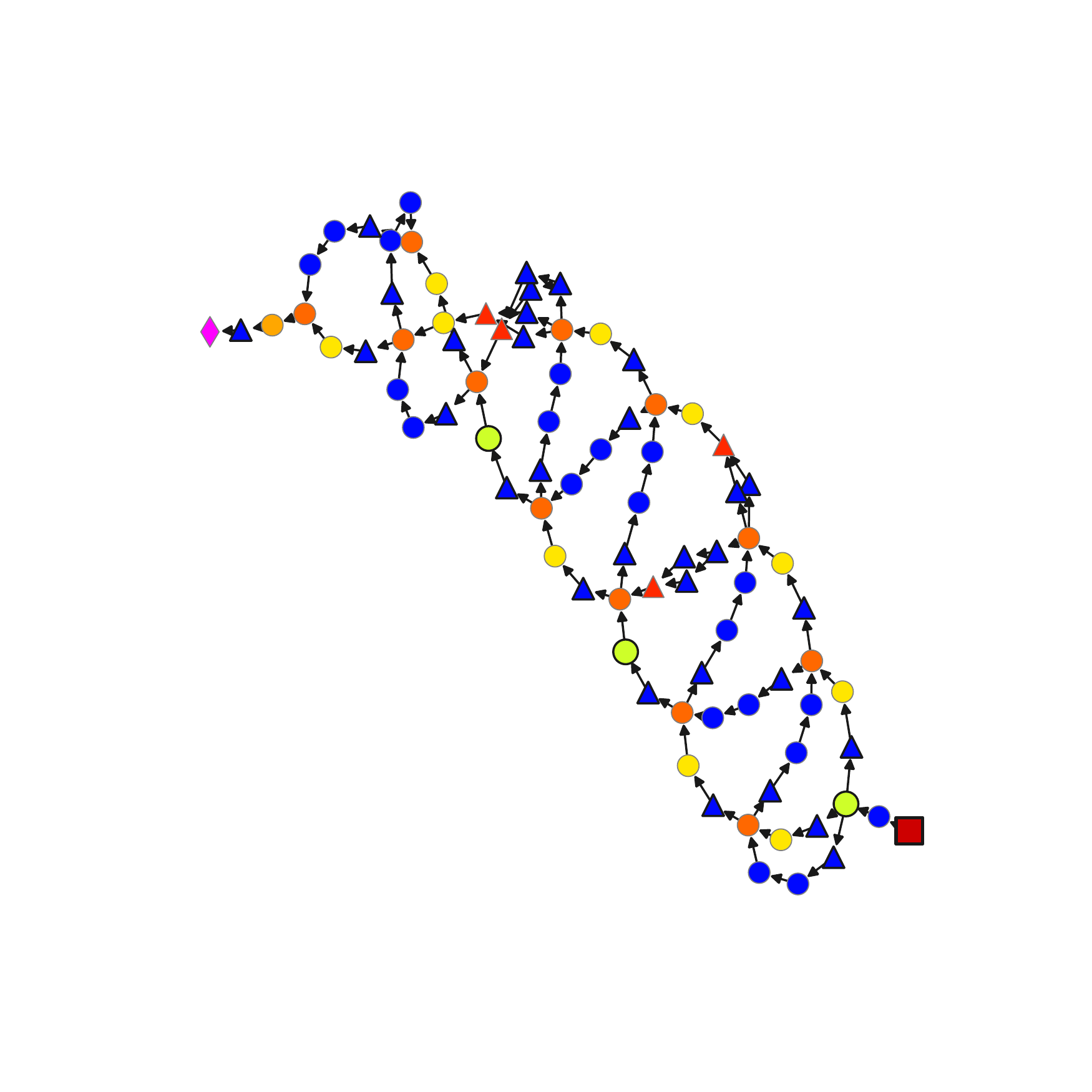}}\Bstrut\\
		
		\multicolumn{1}{c}{{\includegraphics[width=\width,align=c,trim={2.3cm 3cm 2.3cm 3cm},clip]{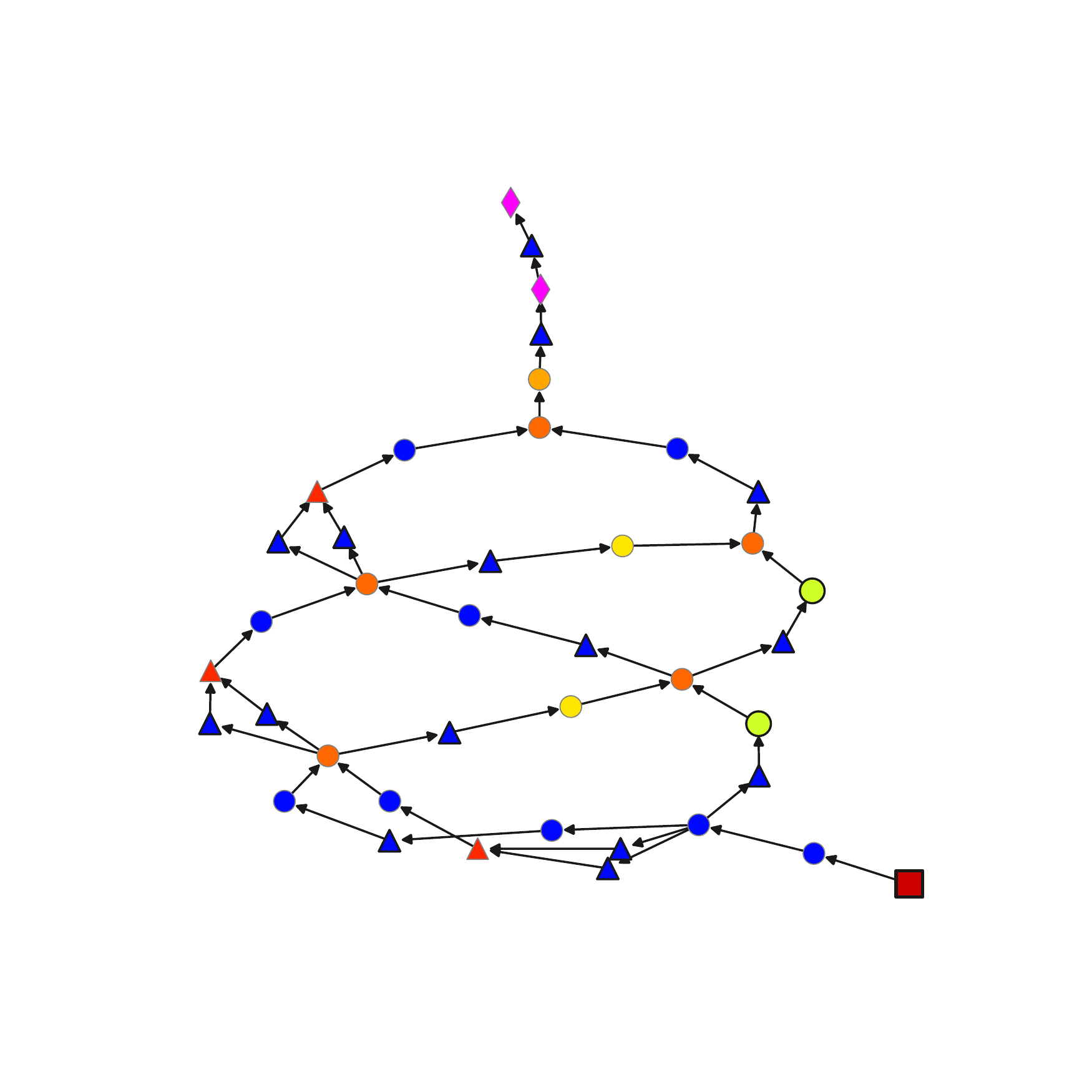}}} & & {\includegraphics[width=\width,align=c,trim={2.3cm 3cm 2.3cm 3cm},clip]{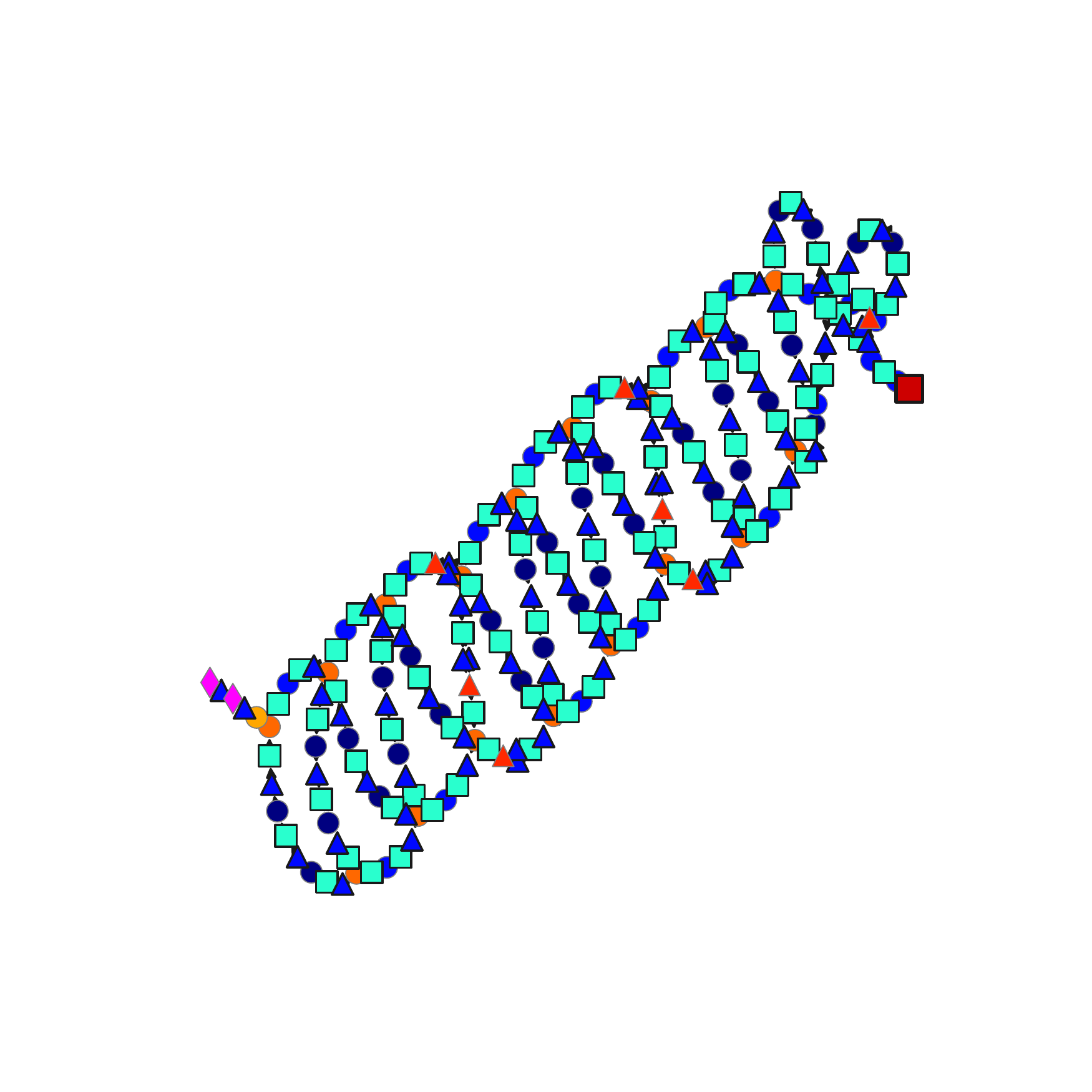}} & 
		{\includegraphics[width=\width,align=c,trim={2.3cm 3cm 2.3cm 3cm},clip]{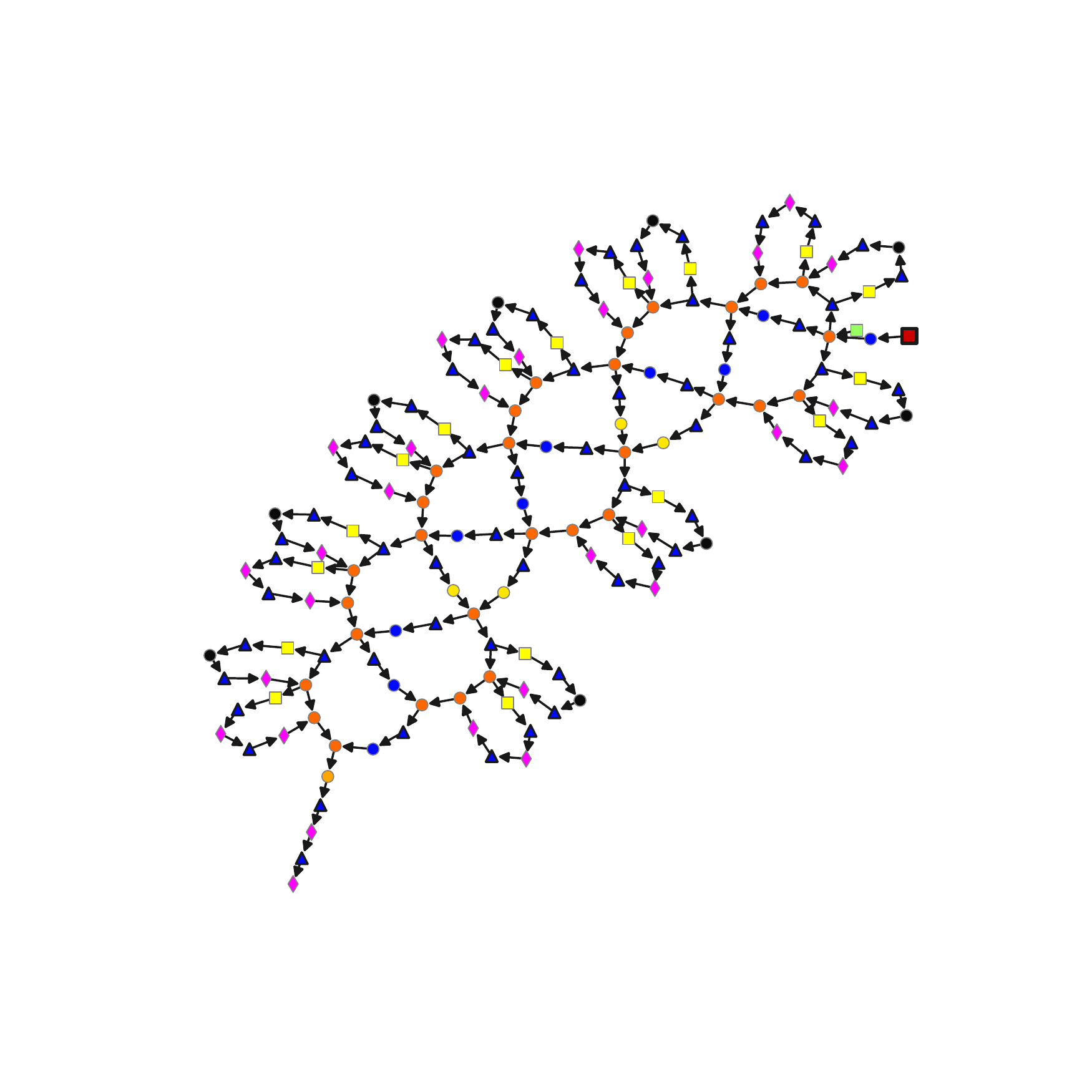}} & 
		\includegraphics[width=\width,align=c,trim={2.3cm 3cm 2.3cm 3cm},clip]{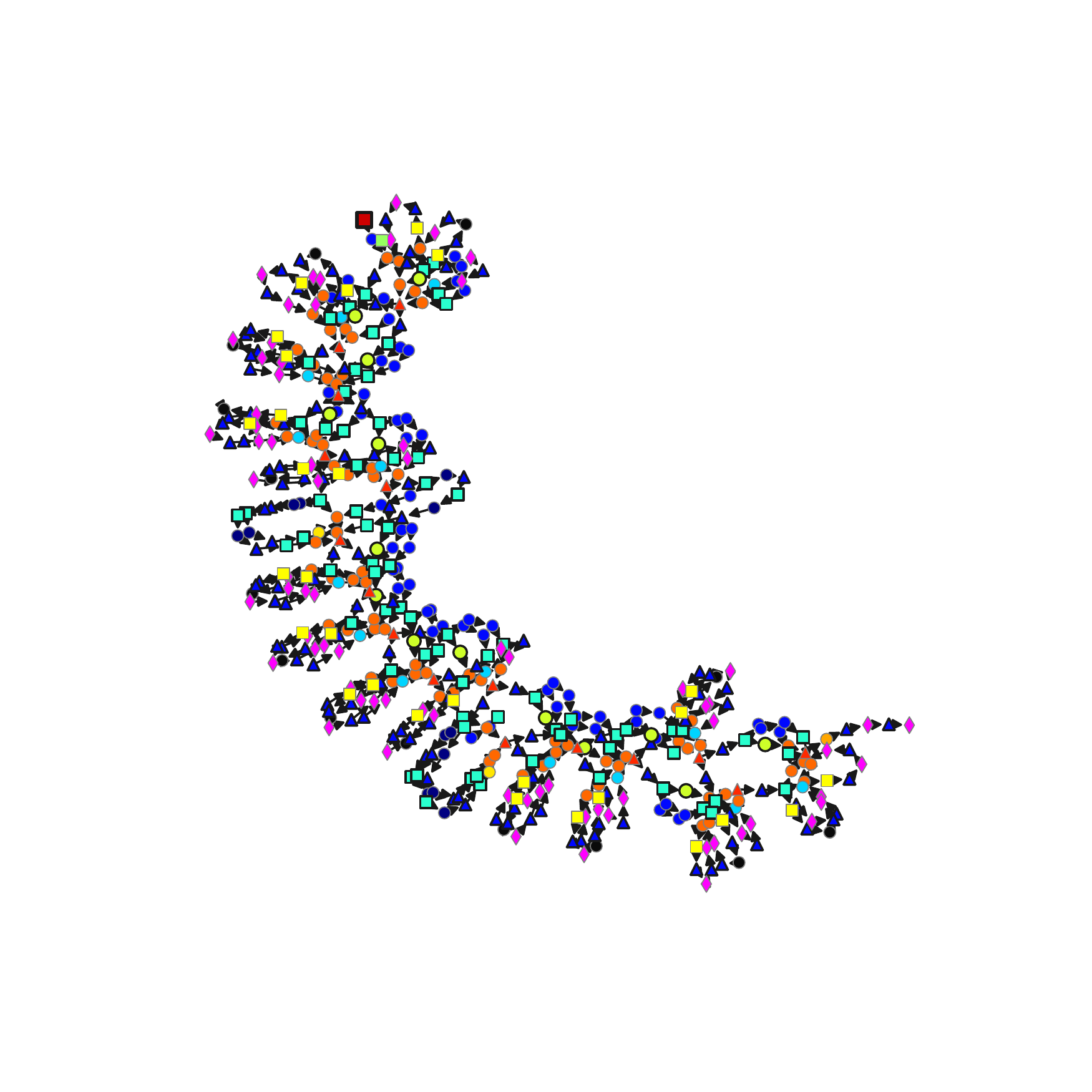} & 
		{\includegraphics[width=\width,align=c,trim={2.3cm 3cm 2.3cm 3cm},clip]{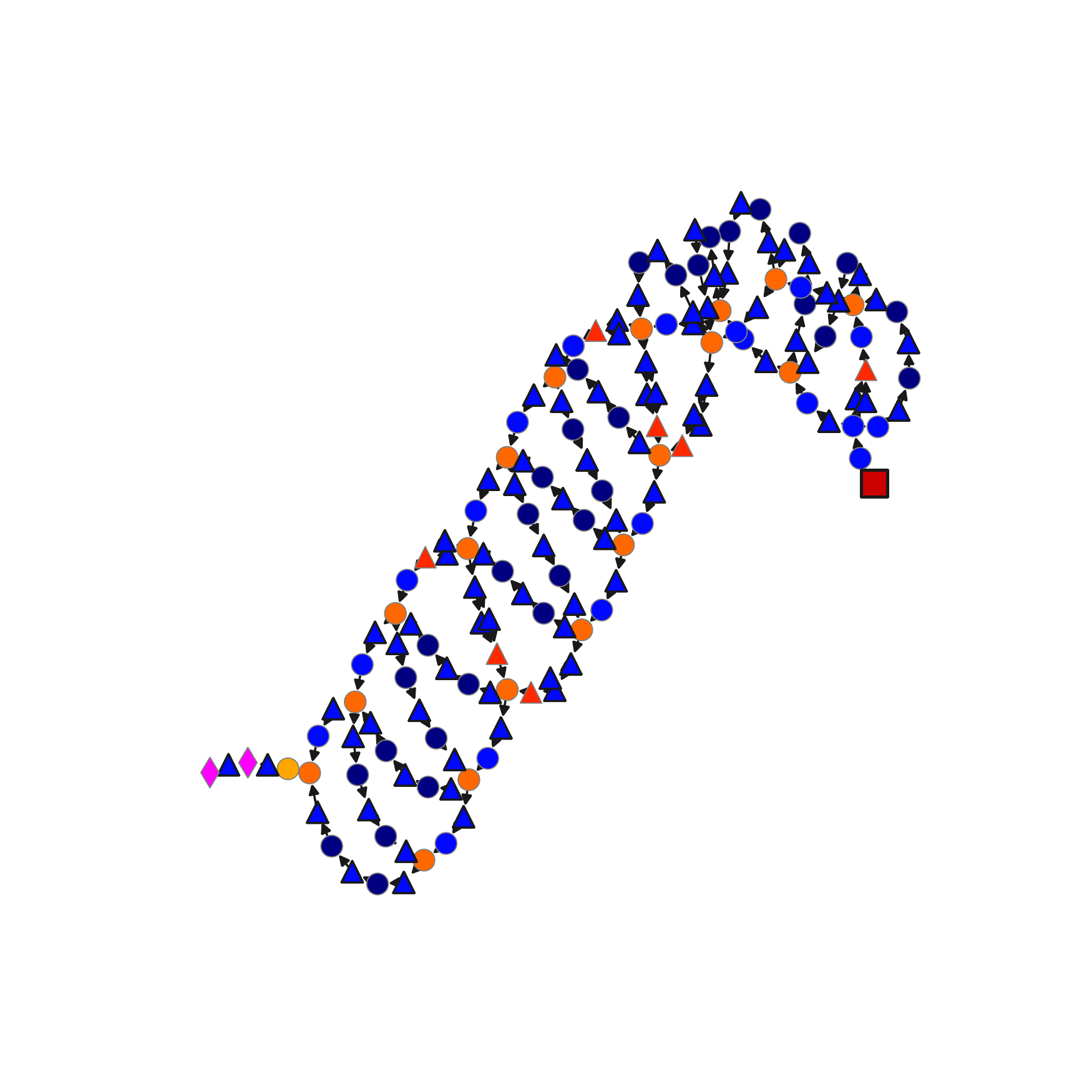}}\Bstrut\\
		
		\multicolumn{1}{c}{{\includegraphics[width=\width,align=c,trim={2.3cm 3cm 2.3cm 3cm},clip]{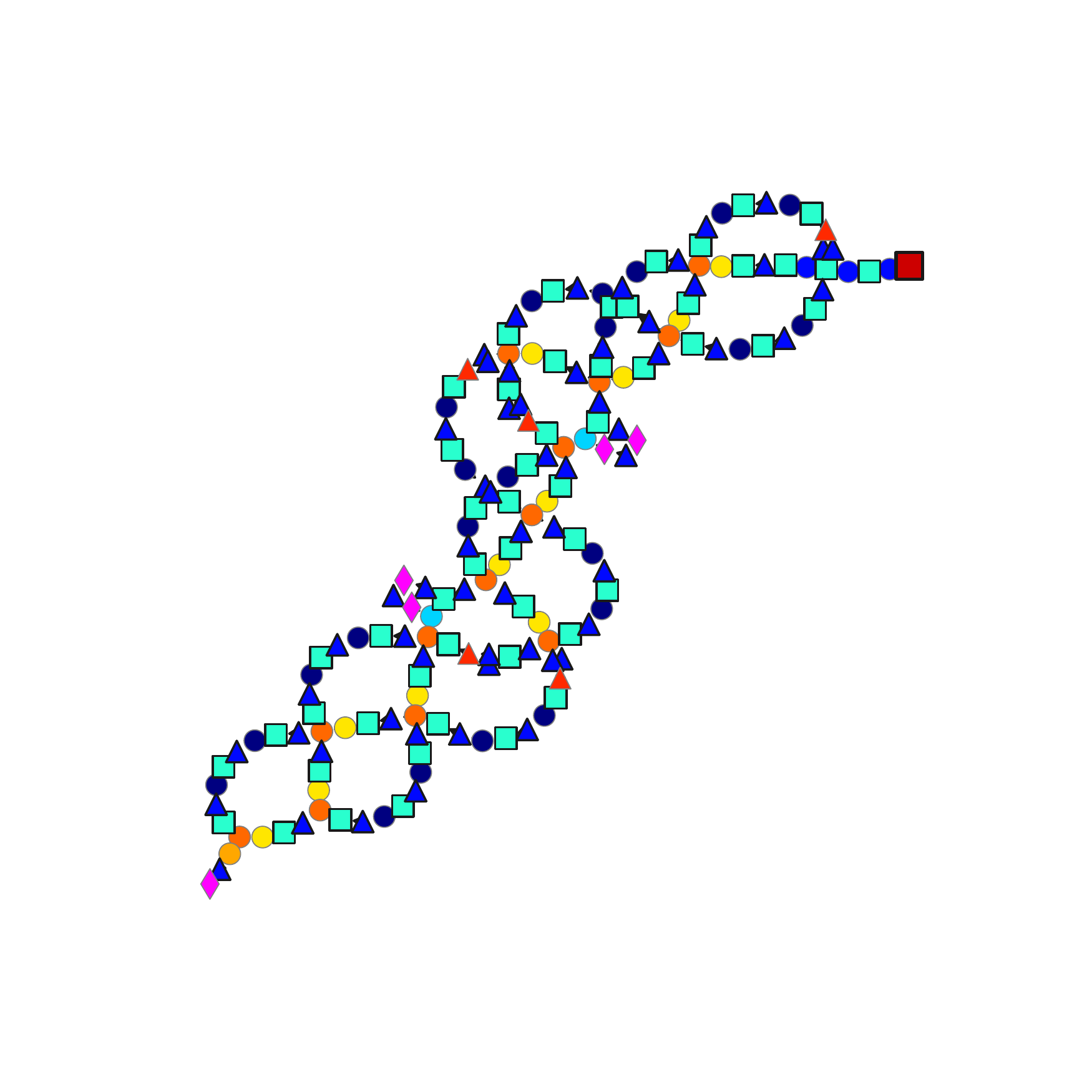}}} & & {\includegraphics[width=\width,align=c,trim={2.3cm 3cm 2.3cm 3cm},clip]{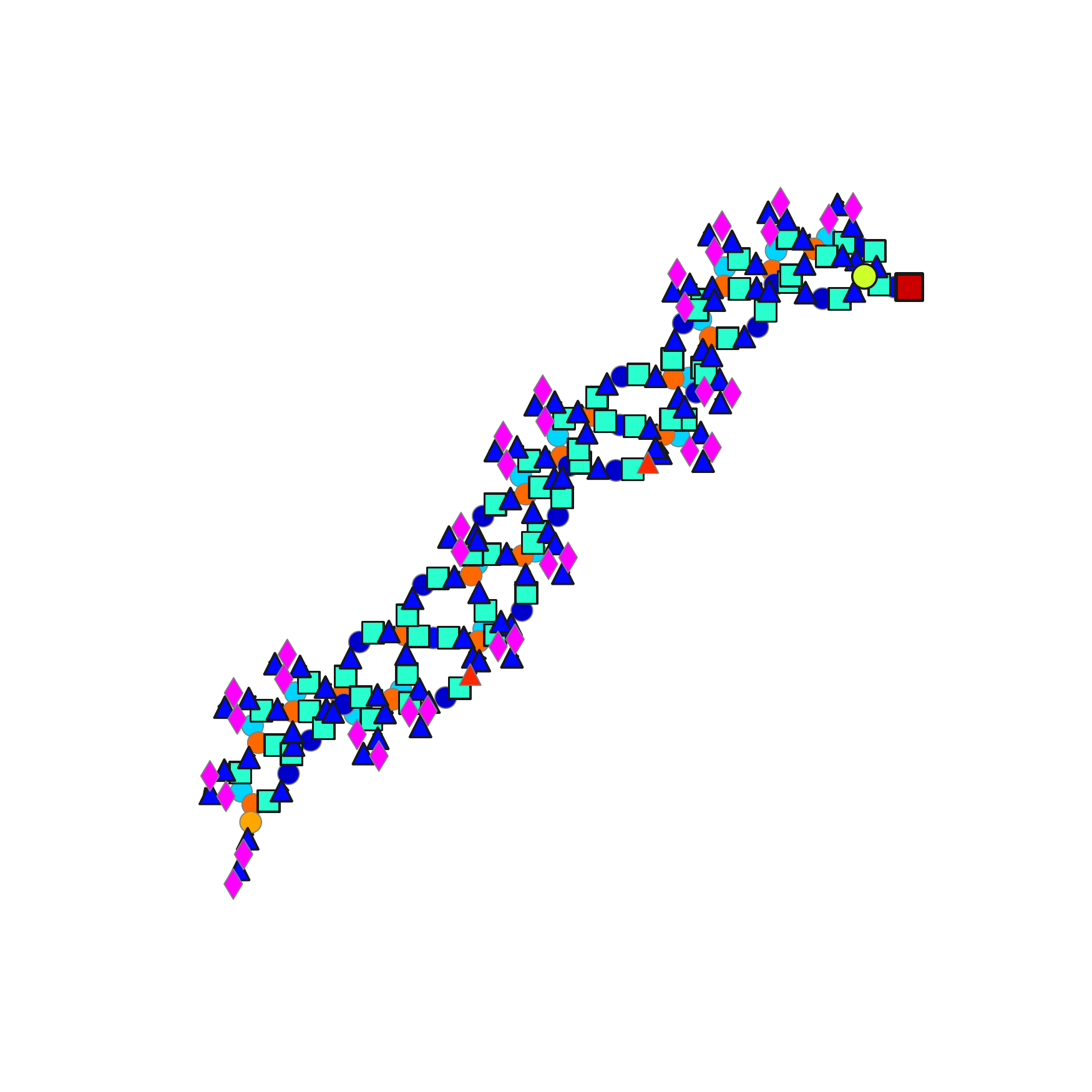}} & 
		{\includegraphics[width=\width,align=c,trim={2.3cm 3cm 2.3cm 3cm},clip]{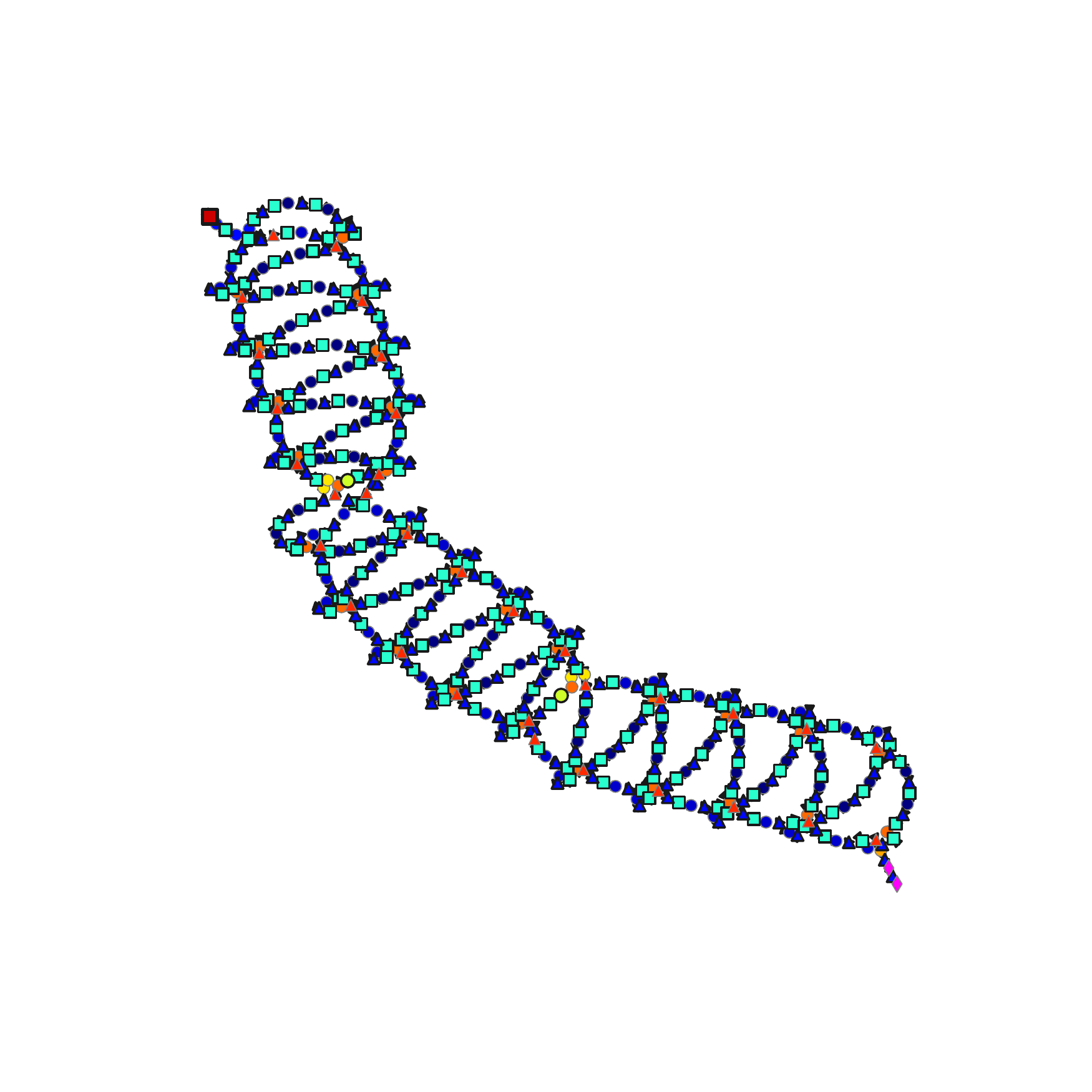}} & 
		\includegraphics[width=\width,align=c,trim={2.3cm 3cm 2.3cm 3cm},clip]{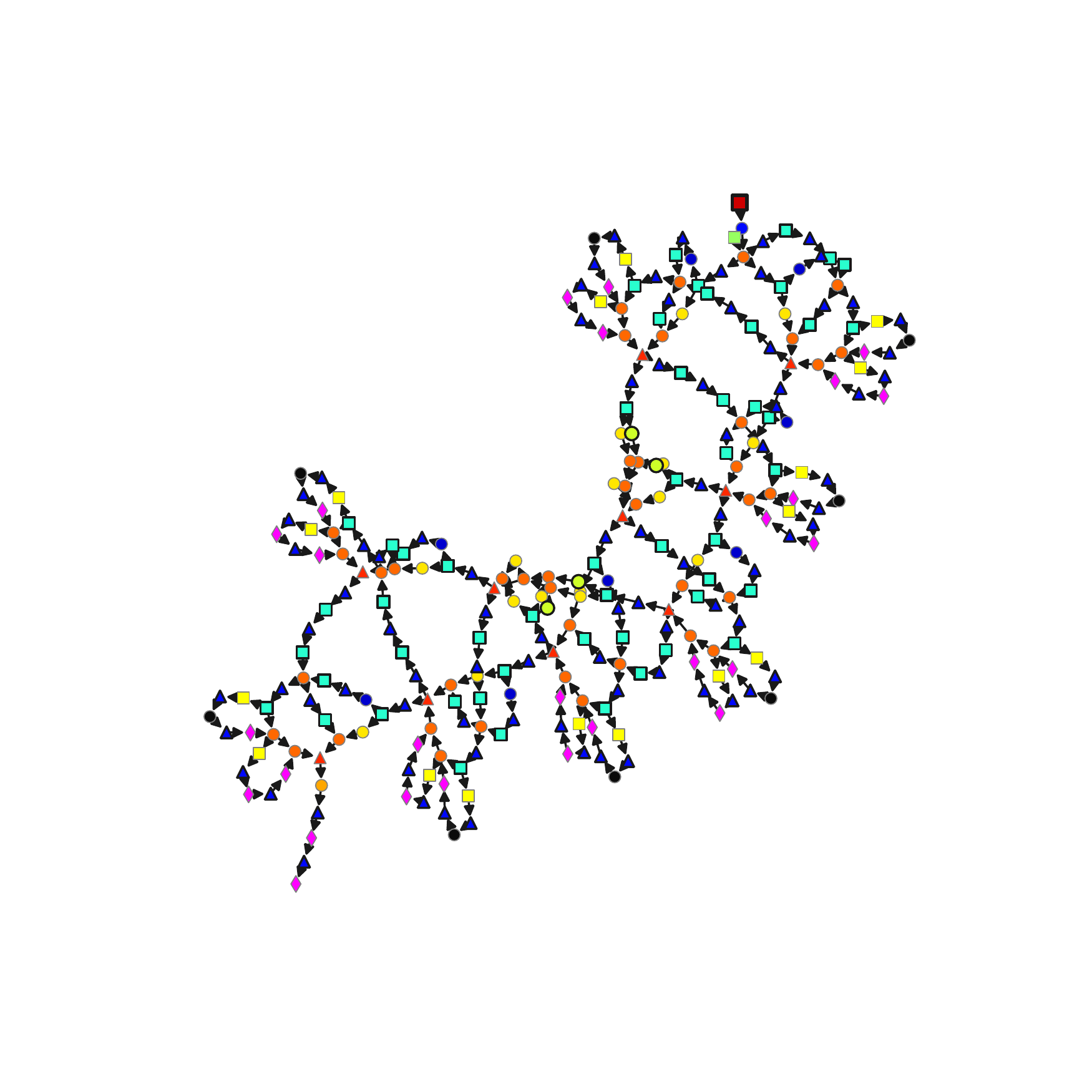} & 
		{\includegraphics[width=\width,align=c,trim={2.3cm 3cm 2.3cm 3cm},clip]{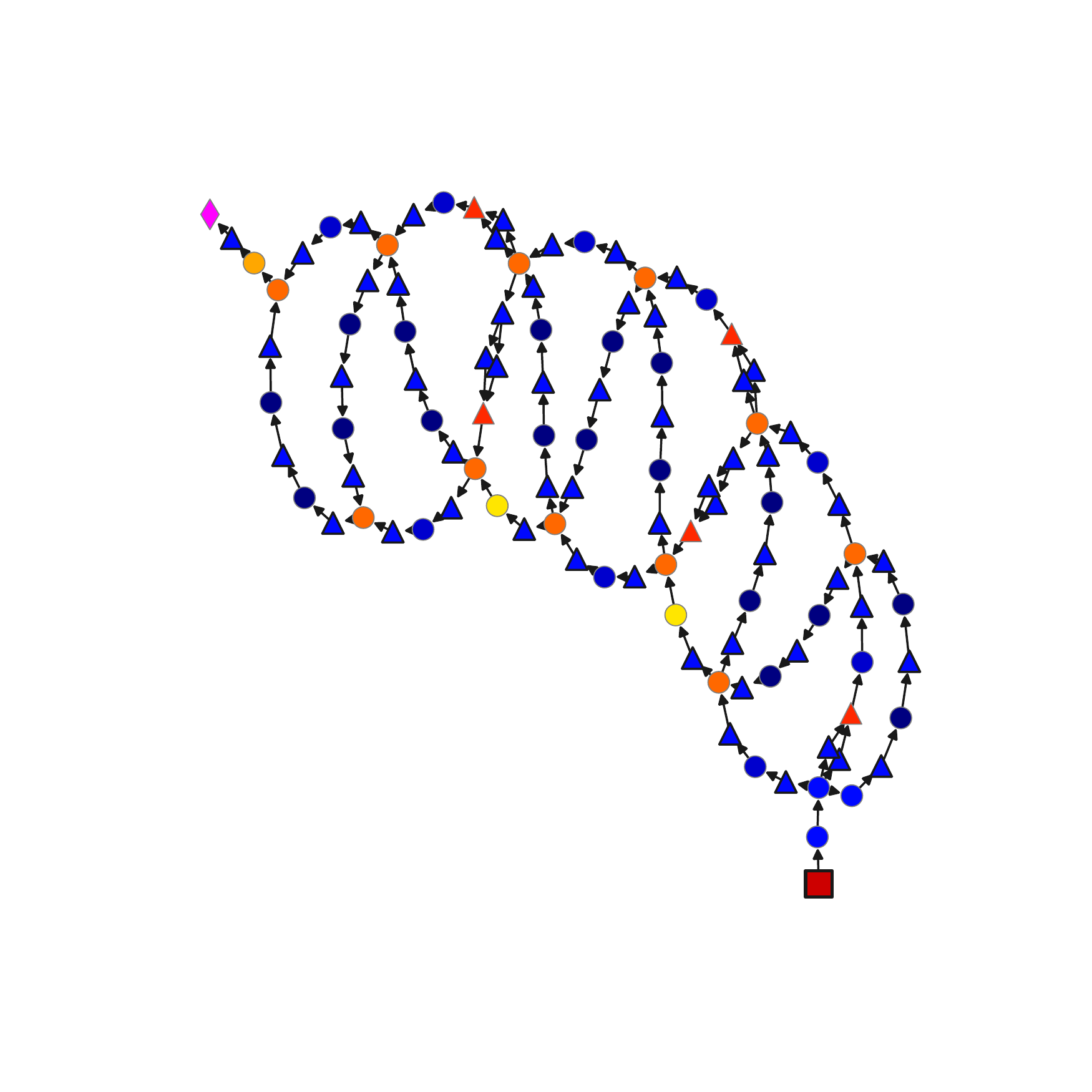}}\Bstrut\\
		\bottomrule
	\end{tabular}
	\caption{Examples of graphs in each subset of our \dataset visualized using NetworkX~\cite{hagberg2008exploring}.}
	\label{fig:more_examples}
	\vspace{-2pt}
\end{figure}

We show the statistics of the key properties of our \dataset in Fig.~\ref{fig:vis_stats} and more examples of computational graphs for different subsets in Fig.~\ref{fig:more_examples}.

\section{GHN Details\label{apdx:ghn_bg}}

\subsection{Baseline \textsc{Ghn}: \ghnbase\label{apdx:ghn_1}}

GHNs were designed for NAS, which typically make strong assumptions about the choice of operations and their possible dimensions to make search and learning feasible. For example, non-separable 2D convolutions (e.g. with weights like $512 \PLH 512 \PLH 3 \PLH 3$ in ResNet-50) are not supported. Our parameter prediction task is more general than NAS, and tackling it using the vanilla GHNs of~\cite{zhang2018graph} is not feasible (mainly, in terms of GPU memory and training efficiency) as we show in \S~\ref{apdx:ablations} (Table~\ref{tab:ablations_more}). So we first make the following modifications to GHNs and denote this baseline as \ghnbase.\looseness-1

\begin{enumerate}
	
	\item \textbf{Compact decoder:} We support the prediction of full 4D weights of shape $<$\textit{out channels $\PLH$ input channels $\PLH$ height $\PLH$ width}$>$	
	that is required for non-separable 2D convolutions.
	Using an MLP decoder of vanilla GHNs~\cite{zhang2018graph} would require it to have a prohibitive number of parameters (e.g.~$\sim$4 billion parameters, see Table~\ref{tab:ablations_more}).
	To prevent that, we use an MLP decoder only to predict a small 3D tensor and then apply a $1 \PLH 1$ convolutional layer across the channels to increase their number followed by reshaping it to a 4D tensor.
	
	\item  \textbf{Diverse channel dimensions:}
	To enable prediction of \params with channel dimensions larger than observed during training we implement a simple tiling strategy similar to~\cite{ha2016hypernetworks}. In particular, instead of predicting a tensor of the maximum shape that has to be known prior training (as done in~\cite{zhang2018graph}), we predict a tensor with fewer channels, but tile across channel dimensions as needed.
	Combining this with \#1 described above, our decoder first predicts a tensor of shape $128 \PLH {\cal S} \PLH {\cal S}$, where ${\cal S}=11$ for CIFAR-10 and ${\cal S}=16$ for ImageNet. Then, the $1 \PLH 1$ convolutional decoder with 256 hidden and 4096 output units transforms this tensor to the tensor of shape $64 \PLH 64 \PLH {\cal S} \PLH {\cal S}$.
	Modifications \#1 and \#2 can be viewed as strong regularizers that can hinder expressive power of a GHN and the \params it predicts, but on the other side permit a more generic and efficient model with just around 1.6M-2M parameters.
	
	\item  \textbf{Fewer decoders:} One alternative strategy to reduce the number of parameters in the decoder of GHNs is to design multiple specialized decoders. However, this strategy does not scale well with adding new operations. Our modifications \#1 and \#2 allow us to have only three decoders:
	for convolutional and fully connected weights, for 1D weights and biases, such as affine transformations in normalization layers, and for the classification layer. 
	
	\item \textbf{Shape encoding:} The vanilla GHNs does not leverage the information about channel dimensionalities  of parameters in operations. For example, the vanilla GHNs only differentiate between $3\PLH 3$ and $5\PLH 5$ convolutions, but not between $512 \PLH 512 \PLH 3\PLH 3$ and $512 \PLH 256 \PLH 3\PLH 3$.
	To alleviate that, we add two shape embedding layers: one for the spatial and one for the channel dimensions.
	For the spatial dimensions (height and width of convolutional weights) we predefine 11 possible values from 1 to $S$. 
	For the channel dimensions (input and output numbers of channels) we predefine 392 possible values from 1 to 8192. We describe 3D, 2D and 1D tensors as special cases of 4D tensors using the value of 1 for missing dimensionalities (e.g. $10\PLH1\PLH1\PLH1$ for CIFAR-10 biases in the classification layer). The shape embedding layers transform the input shape into four (two for spatial and two for channel dimensions) 8-dimensional learnable vectors that are concatenated to obtain a 32-dimensional vector. This vector is summed with a 32-dimensional vector encoding one of the 15 operation types (primitives). In the rare case of feeding the shape that is not one of the predefined values, we look up for the closest value. This can work well in some cases, but can also hurt the quality of predicted parameters, so some continuous encoding as in~\cite{vaswani2017attention} can be used in future work.
\end{enumerate}

\subsection{Our improved \textsc{Ghn}: \ghnours\label{apdx:ghn_2}}

\subsubsection{\ghnours Architecture}
Our improved \ghnours is obtained by adding to \ghnbase the differentiable normalization of predicted \params, virtual edges (and the associated `mlp\_ve' module, see the architecture below), meta-batching and layer normalization (and the associated `ln' module). A high-level PyTorch-based overview of the \ghnours model architecture used to predict the \params for ImageNet is shown below (see our code for implementation details).

\vspace{15pt}

\begin{small}
	\begin{Verbatim}[commandchars=\\\{\}]
	(ghn): GHN(  \textcolor{green}{# our hypernetwork H_D with total 2,319,752 parameters (theta)}
	(gcn): GCNGated(  \textcolor{green}{# Message Passing for Equations 3 and 4}
	(mlp): MLP(
	(fc): Sequential(
	(0): Linear(in_features=32, out_features=16, bias=True)
	(1): ReLU()
	(2): Linear(in_features=16, out_features=32, bias=True)
	(3): ReLU()
	)
	)
	(mlp_ve): MLP(  \textcolor{green}{# only in GHN-2 (see Eq. 4)}
	(fc): Sequential(
	(0): Linear(in_features=32, out_features=16, bias=True)
	(1): ReLU()
	(2): Linear(in_features=16, out_features=32, bias=True)
	(3): ReLU()
	)
	)
	(gru): GRUCell(32, 32)  \textcolor{green}{\scriptsize # all GHNs use d=32 for input, hidden and output node feature dimensionalities}
	)
	(ln): LayerNorm((32,), eps=1e-05, elementwise_affine=True)  \textcolor{green}{# only in GHN-2}
	(shape_embed_spatial): Embedding(11, 8)  \textcolor{green}{# encodes the spatial shape of predicted params}
	(shape_embed_channel): Embedding(392, 8)  \textcolor{green}{# encodes the channel shape of predicted params}
	(op_embed): Embedding(16, 32)  \textcolor{green}{# 15 primitives plus one extra embedding for dummy nodes}
	(decoder): Decoder(
	(fc): Sequential(
	(0): Linear(in_features=32, out_features=32768, bias=True)
	(1): ReLU()
	)  \textcolor{green}{# predicts a 128x16x16 tensor}
	(conv): Sequential(
	(0): Conv2d(128, 256, kernel_size=(1, 1), stride=(1, 1))
	(1): ReLU()
	(2): Conv2d(256, 4096, kernel_size=(1, 1), stride=(1, 1))
	)  \textcolor{green}{# predicts a 64x64x16x16 tensor}
	(class_layer_predictor): Sequential(
	(0): ReLU()
	(1): Conv2d(64, 1000, kernel_size=(1, 1), stride=(1, 1))
	)  \textcolor{green}{# predicts 64x1000 classification weights given the 64x64x16x16 tensor}
	)
	(norm_layers_predictor): Sequential(
	(0): Linear(in_features=32, out_features=64, bias=True)
	(1): ReLU()
	(2): Linear(in_features=64, out_features=128, bias=True)
	)  \textcolor{green}{# predicts 1x64 weights and 1x64 biases given the 1x32 node embeddings}
	(bias_predictor): Sequential(
	(0): ReLU()
	(1): Linear(in_features=64, out_features=1000, bias=True)
	)  \textcolor{green}{# predicts 1x1000 classification biases given the 64x64x16x16 tensor}
	)
	\end{Verbatim}
\end{small}

\subsubsection{Differentiable Normalization of Predicted Parameters\label{apdx:renorm}}

\begin{figure}[tbhp]
	\vspace{-1pt}
	\centering
	\setlength{\tabcolsep}{10pt}
	\begin{tabular}{cc}
		\includegraphics[width=0.46\textwidth,align=c,trim={0 0 0 0cm}, clip]{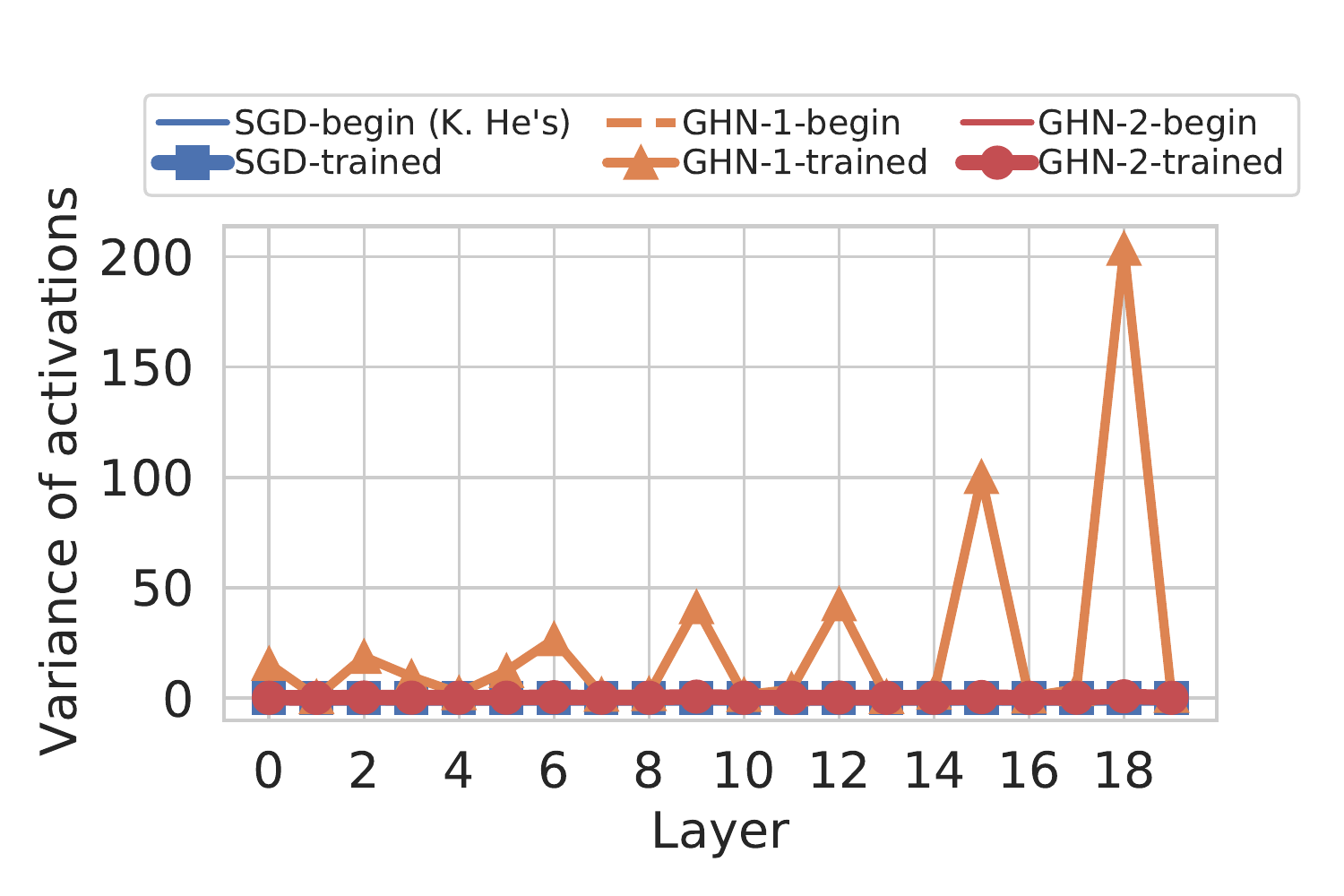} & \includegraphics[width=0.46\textwidth,align=c,trim={0 0 0 0cm}, clip]{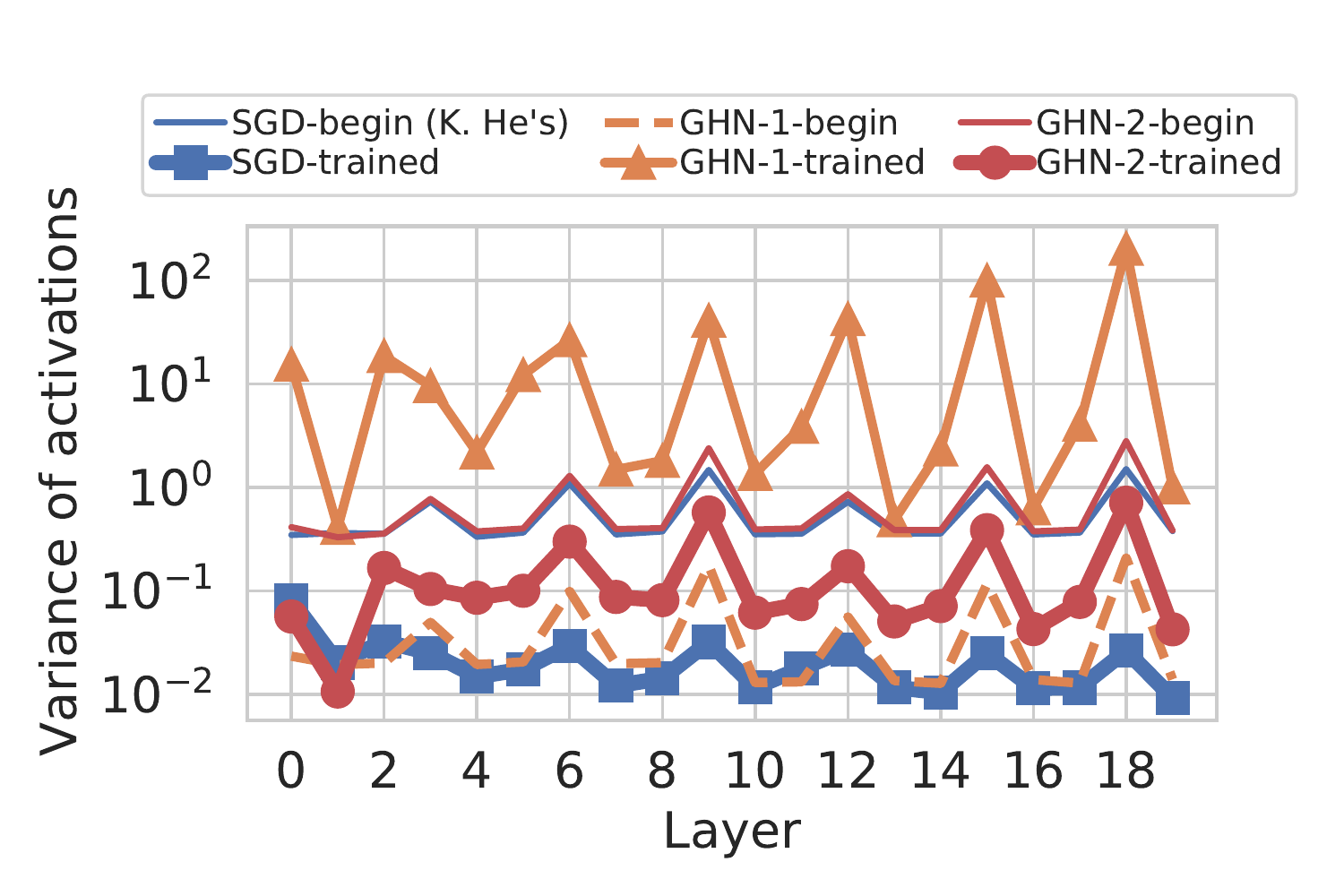}
	\end{tabular}
	\vspace{-5pt}
	\caption{The effect of normalizing predicted parameters on the variance of activations in the first several layers of ResNet-50: a linear (\textbf{left}) and log (\textbf{right}) scale on the y axis. }
	\label{fig:activations}
	\vspace{-1pt}
\end{figure}

We analyze in more detail the effect of normalizing predicted parameters (Fig.~\ref{fig:activations}).

\paragraph{Setup.}
As we discussed in \S~\ref{sec:renorm}, \citet{chang2019principled} proposed a method to initialize a hypernetwork to stabilize the activations in the network for which the parameters are predicted. However, this technique requires knowing upfront the shapes of the predicted \params, and therefore is not applicable out-of-the-box in our setting, where we predict the \params of diverse architectures with arbitrary shapes of weights. So, instead we apply {operation-dependent normalizations}. We analyze the effect of this normalization by taking a GHN in the beginning and end of training on CIFAR-10 and predicting parameters of ResNet-50. To compute the variance of activations in the ResNet-50, we forward pass a batch of test images through the predicted parameters. 

\paragraph{Observations.} The activations obtained using \ghnbase explode after training, which aligns with the analysis of \cite{chang2019principled} (this is more obvious on the left of Fig.~\ref{fig:activations}, where a linear scale is used on the y axis). For \ghnours, in the beginning of training the activations match the ones of ResNet-50 initialized randomly using Kaiming He's method~\cite{he2015delving}, which validates the choice of our normalization equations in \S~\ref{sec:renorm}. By the end of training, the activations of models for the random-based and the \ghnours-based cases decrease (perhaps, due to the weight decay), however, the ones of \ghnours reduce less, indicating that the predicted \params do not reach the state of those trained with SGD from scratch. In contrast, the activations corresponding to \ghnbase have small values in the beginning, but explode by the end of training. Matching the activations of the models trained with SGD can be useful to improve training of GHNs and, for example, to make fine-tuning of predicated parameters easier as we show in \S~\ref{sec:finetune}, where the parameters predicted by \ghnbase are shown difficult to be fine-tuned with SGD.\looseness-1

\subsubsection{Meta-batching\label{apdx:meta}}
\begin{figure}[tbhp]
	\vspace{-1pt}
	\centering
	\setlength{\tabcolsep}{10pt}
	\begin{tabular}{cc}
		\includegraphics[width=0.46\textwidth,align=b,trim={0 0 0 0cm}, clip]{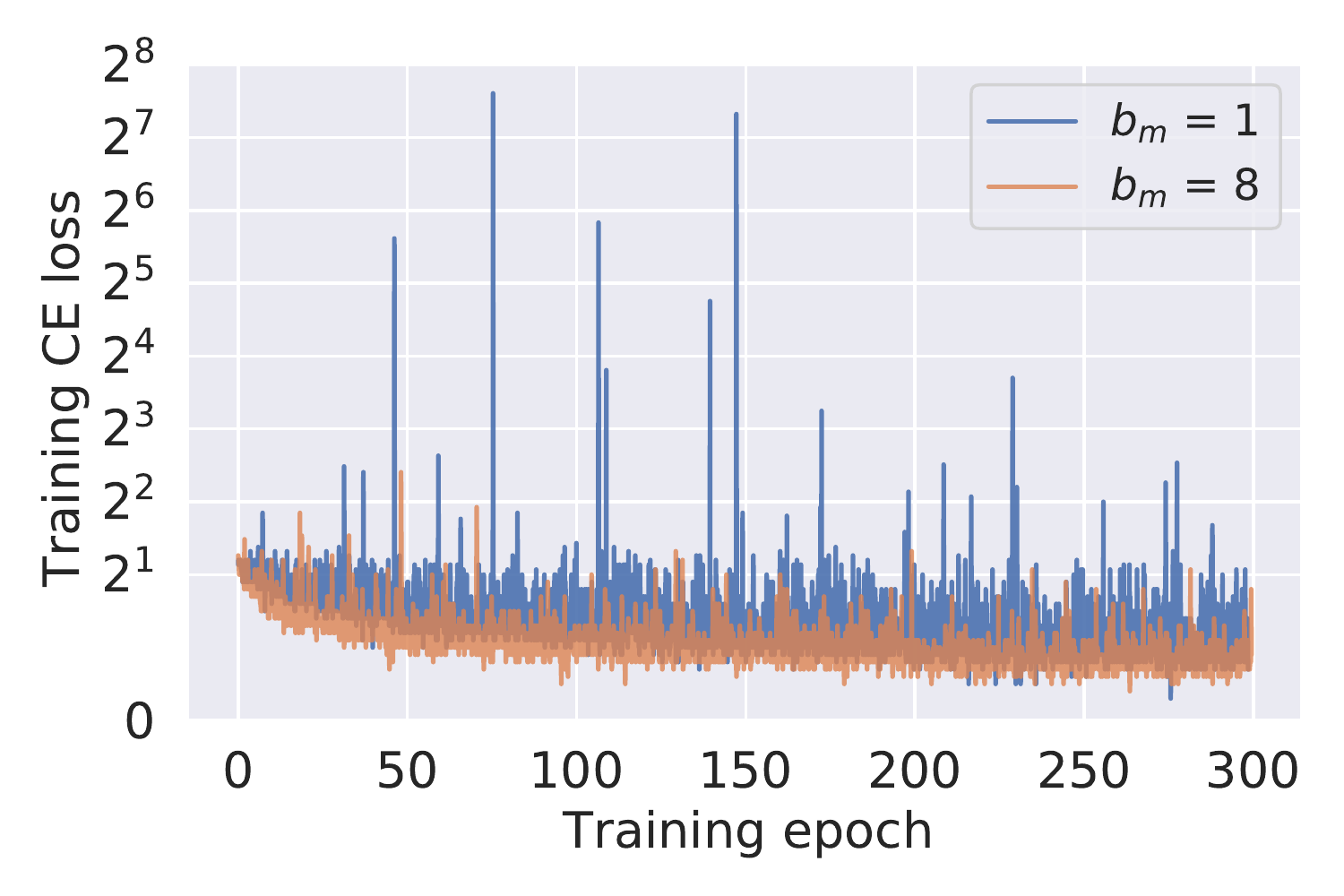} & \includegraphics[width=0.46\textwidth,align=b,trim={0 0 0 0cm}, clip]{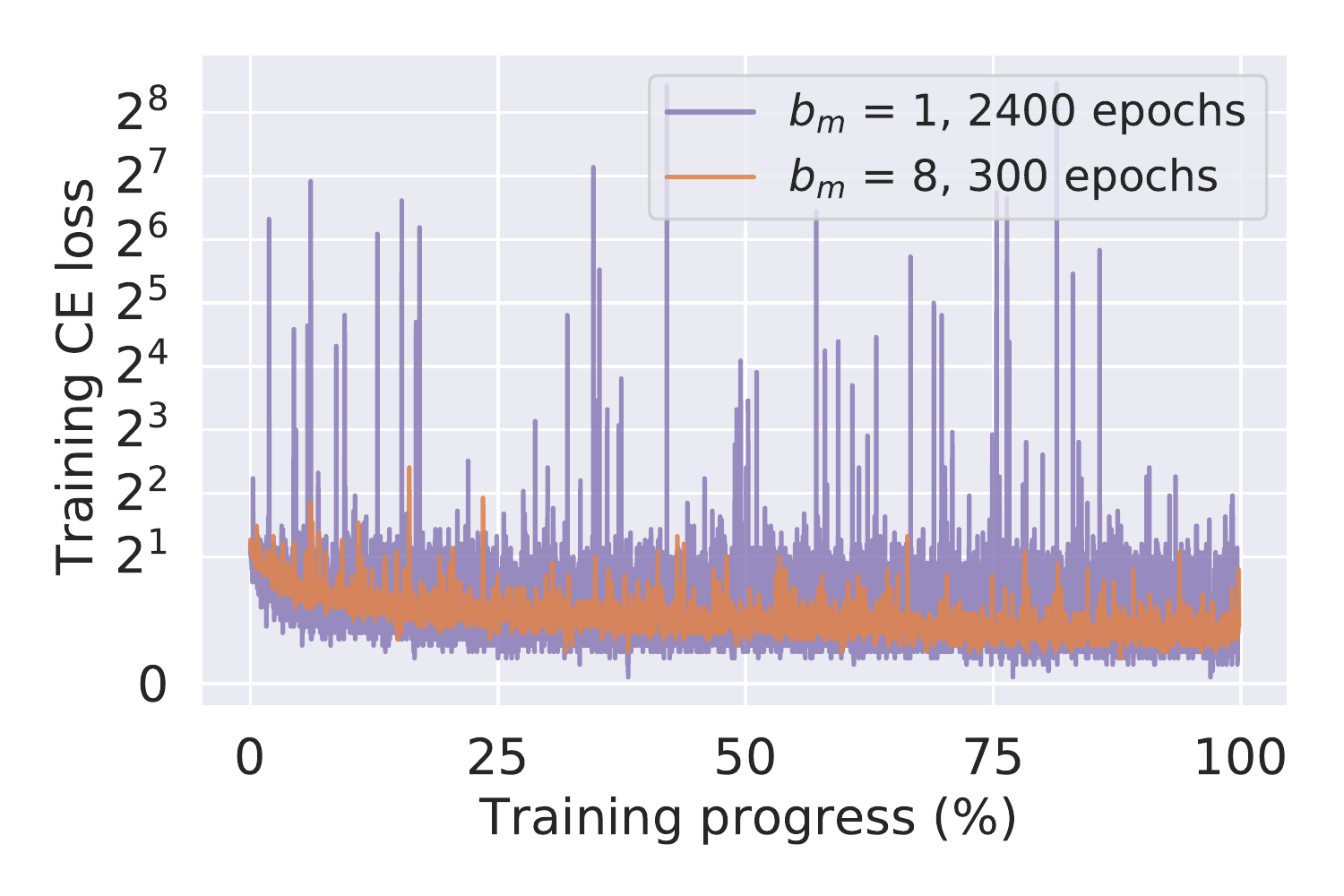}\\
	\end{tabular}
	\vspace{-5pt}
	\caption{Effect of using more architectures per batch of images on the training loss (\textbf{left}) and comparison to training longer (\textbf{right}). In the figure on the right, we shrink the x axis of the $b_m=1$ case, so that both plots can be compared w.r.t. the total number of epochs.   }
	\label{fig:meta_batch}
	\vspace{-1pt}
\end{figure}

We analyze how meta-batching affects the training loss when training GHNs on CIFAR-10 (Fig.~\ref{fig:meta_batch}). The loss of the \ghnours with $b_m=8$ is less noisy and is lower throughout the training compared to using $b_m=1$. In fact, the loss of $b_m=1$ is unstable to the extent that oftentimes the training fails due to the numerical overflow, in which case we ignore the current architecture and resample a new one.
For example, the standard deviation of gradient norms with $b_m=8$ is significantly lower than with $b_m=1$: 18 vs 145 with means 2.7 and 7.4 respectively.
Training the model with $b_m=1$ eight times longer (Fig.~\ref{fig:meta_batch}, right) boosts the performance of predicted parameters (Table~\ref{tab:ablations_more}), but still does not reach the level of $b_m=8$; it also still suffers from the aforementioned numerical issues and does not leverage the parallelism of $b_m=8$ (all architectures in a meta-batch can be processed in parallel). 
Further increasing the meta-batch size is an interesting avenue for future research.

\section{Additional Experiments and Details\label{apdx:exper}}

\subsection{Experimental Details\label{apdx:details}}

\paragraph{Training GHNs.} 
We train all GHNs for 300 epochs using Adam with an initial learning rate 0.001, batch size of 64 images for CIFAR-10 (256 for ImageNet), weight decay of 1e-5. The learning rate is decayed after 200 and 250 epochs. On ImageNet \ghnours's validation accuracy plateaued faster (in terms of training epochs), so we stopped training after 150 epochs decaying it after 100 and 140 epochs.\looseness-1

\paragraph{Evaluation of networks with BN using GHNs.} While our GHNs predict all \textit{trainable} parameters, batch norm networks have running statistics that \textit{are not learned by gradient descent}~\cite{ioffe2015batch}, and so are not predicted by GHNs. To obtain these statistics, we evaluate the networks with BN by computing per batch statistics on the test set as in \cite{zhang2018graph} using a batch size 64. Another alternative we have considered is updating the running statistics by forward passing several batches of the training or testing images through each evaluation network (no labels, gradients or parameter updates are required for this stage). For example on CIFAR-10 if we forward pass 200 batches of the training images (takes less than 10 seconds on a GPU), we obtain a higher accuracy of 71.7\% on \iid-\iidtest compared to 66.9\% when the former strategy is used. On ImageNet, the difference between the two strategies is less noticeable.
For both strategies, the results can slightly change depending on the order of images for which the statistics is estimated.

\subsection{Additional Results\label{apdx:results}}

\subsubsection{Additional Ablations\label{apdx:ablations}}

We present results on CIFAR-10 for different variants of GHNs (Table~\ref{tab:ablations_more}) in addition to those presented in Tables~\ref{tab:bench_c10} and~\ref{tab:ablations}.
Overall, different GHN variants show worse or comparable results on all evaluation architectures compared to our \ghnours, while in some cases having too many trainable parameters making training infeasible in terms of GPU memory or being less efficient to train (e.g. with $T=5$ propagation steps).
For ViT, \ghnours is worse compared to other GHN variants, which requires further investigation.

\begin{table}[tbhp]
	\caption{\small CIFAR-10 results of predicted parameters for the evaluation architectures of \dataset. Mean (\sem{}standard error of the mean) accuracies are reported. Different ablated \ghnours models are evaluated. \textsuperscript{*}GPU seconds per a batch of 64 images and $b_m$ architectures. \textsuperscript{**}In~\cite{zhang2018graph} the GHNs have fewer parameters due to a more constrained network design space (as discussed in~\ref{apdx:ghn_1}) and applying specialized decoders for different operations. The best result in each column is bolded, the best results with $b_m = 1$ (excluding training 8 $\PLH$ longer) are underlined.
	}
	\label{tab:ablations_more}
	\centering
	\tiny
	\vspace{3pt}
	\setlength{\tabcolsep}{2pt}
	\newcommand{\better}[1]{\underline{#1}}
	\rowcolors{4}{white}{gray!10}
	\begin{tabular}{lcccp{0.1cm}ccp{0.1cm}llp{0.1cm}lllll}
		\toprule
		\textbf{\textsc{\small Model}} & \textbf{Norm} & \textbf{Virt.} & \textbf{M. batch} & & \textbf{\#GHN} & \textbf{Train.} & & \multicolumn{2}{c}{\textbf{\textsc{\small \iid-test}}} &
		& 
		\multicolumn{5}{c}{\small \textbf{\textsc{OOD-test}}} \\
		& $\hat{\w}_p$ & \textbf{edges} & $b_m=8$ & & \textbf{params} & \textbf{speed\textsuperscript{*}} & & \multicolumn{1}{c}{avg} & max & & \wide & \deep & \dense & \bnfree & \tiny \textsc{ResNet/ViT}\Bstrut\\ 
		\cline{1-4}\cline{6-7}\cline{9-10}\cline{12-16} \\
		\ghnours & \cmark & \cmark & \cmark & & 1.6M & 3.6 & & \textbf{66.9}\sem{0.3} & {77.1} & & \textbf{64.0}\sem{1.1} & \textbf{60.5}\sem{1.2} & \textbf{65.8}\sem{0.7} & {36.8}\sem{1.5} & \textbf{58.6}/11.4\Bstrut\\
		\midrule

		1000 archs & \cmark & \cmark & \cmark & & 1.6M & 3.6 & & 65.1\sem{0.5} & \textbf{78.4} & & 61.5\sem{1.6} & 56.0\sem{1.5} & 65.0\sem{0.9} & 27.6\sem{1.1} & 58.2/10.5\Tstrut \\
		100 archs & \cmark & \cmark & \cmark & & 1.6M & 3.6 & & 47.1\sem{0.8} & 77.1 & & 38.8\sem{1.9} & 28.3\sem{1.6} & 41.9\sem{1.5} & 11.0\sem{0.2} & 38.7/10.3 \\
		No normalization & \xmark & \cmark & \cmark & & 1.6M & 3.6 & & 62.6\sem{0.6} & 75.9 & & 52.3\sem{2.1} & 59.5\sem{1.1} & 62.3\sem{1.2} & 14.4\sem{0.4} & 58.3/17.0 \\
		No virtual edges & \cmark & \xmark & \cmark & & 1.6M & 3.6 & & 61.5\sem{0.4} & 73.2 & & 58.2\sem{1.0} & 55.0\sem{0.9} & 61.5\sem{0.6} & \textbf{40.8}\sem{0.8} & 41.9/12.1\\
		No LayerNorm & \cmark & \cmark & \cmark & & 1.6M & 3.6 & & 64.5\sem{0.4} & 75.9 & & 62.4\sem{1.1} & 59.0\sem{1.1} & 64.6\sem{0.6} & 39.6\sem{1.2} & 55.1/8.9\\
		No GatedGNN (MLP) & \cmark & \cmark & \cmark & & 1.6M & 1.5 & & 42.2\sem{0.6} & 60.2 & & 22.3\sem{0.9} & 37.9\sem{1.2} & 44.8\sem{1.1} & 23.9\sem{0.7} & 17.7/10.0\\
		Train 8$\PLH$ longer & \cmark & \cmark & \xmark & & 1.6M & 0.7 & & 62.4\sem{0.5} & 75.8 & & 63.0\sem{1.3} & 58.0\sem{1.3} & 62.1\sem{0.9} & 24.5\sem{0.7} & 57.0/14.6\Bstrut\\
		\midrule
		
		No Meta-batch ($b_m=1$) & \cmark & \cmark & \xmark & & 1.6M & 0.7 & & 54.3\sem{0.3} & 63.0 & & \better{53.1}\sem{0.8} & 51.9\sem{0.6} & 53.4\sem{0.5} & \better{31.7}\sem{0.8} & \better{50.6}/\better{17.8}\Tstrut\\
		
		No Shape Encoding & \cmark & \cmark & \xmark & & 1.6M & 0.7 & & 53.1\sem{0.4} & 61.7 & & 52.4\sem{0.9} & 51.4\sem{0.7} & 53.5\sem{0.6} & 24.7\sem{0.8} & 31.5/14.0\\
		
		No virtual edges (VE) & \cmark & \xmark & \xmark & & 1.6M & 0.7 & & 51.7\sem{0.4} & 62.0 & & 49.7\sem{0.8} & 47.4\sem{0.8} & 52.0\sem{0.8} & 24.5\sem{0.5} & 34.2/14.7 \\
		
		+Stacked GHN, no VE & \cmark & \xmark & \xmark & & 1.6M & 0.7 & & 52.2\sem{0.4} & 63.6 & & 51.3\sem{0.9} & 46.9\sem{0.9} & 52.3\sem{0.7} & 20.2\sem{0.6} & 44.5/15.4\\
		
		+Stacked GHN & \cmark & \cmark & \xmark & & 1.6M & 0.9 & & 53.1\sem{0.4} & 61.3 & & 51.5\sem{1.1} & 50.9\sem{0.7} & 53.5\sem{0.7} & 23.1\sem{0.7} & 42.7/15.1 \\

		$\pi=$ only fw (Eq.~\ref{eq:ghn_prop}) & \cmark & \cmark & \xmark & & 1.6M & 0.4 & & 53.9\sem{0.3} & 62.5 & & 51.2\sem{1.0} & 51.7\sem{0.7} & \better{54.3}\sem{0.5} & 31.0\sem{0.7} & 49.9/11.5 \\
		$T=5$ (Eq.~\ref{eq:ghn_prop}) & \cmark & \cmark & \xmark & & 1.6M & 2.6 & & \better{54.4}\sem{0.4} & 63.3 & & 52.8\sem{1.0} & 50.4\sem{0.9} & 53.4\sem{0.8} & 22.6\sem{1.0} & 50.1/10.1\\ 
		Fan-out & \cmark & \cmark & \xmark & & 1.6M & 0.7 & & 53.8\sem{0.4} & 63.6 & & 52.6\sem{1.0} & 51.2\sem{0.8} & \better{54.3}\sem{0.8} & 19.8\sem{0.6} & 48.5/11.1\\
		MLP decoder & \cmark & \cmark & \xmark & & 32M & 0.7 & & 53.1\sem{0.4} & \better{64.0} & & 52.9\sem{1.0} & \better{52.5}\sem{0.7} & 54.0\sem{0.8} & 22.1\sem{0.5} & 44.1/16.3\\
		No tiling & \cmark & \cmark & \xmark & & 135M & & & \multicolumn{8}{c}{out of GPU memory} \\
		\parbox{2.3cm}{No tiling, MLP decoder \\ (as in a vanilla GHN~\cite{zhang2018graph})} & \cmark & \cmark & \xmark  & & 4.1B\textsuperscript{**} & & & \multicolumn{8}{c}{out of GPU memory} \Bstrut\\
		\midrule
		\ghnbase & \xmark & \xmark & \xmark & & 1.6M & 0.6 & & 51.4\sem{0.4} & 59.9 & & 43.1\sem{1.7} & 48.3\sem{0.8} & 51.8\sem{0.9} & 13.7\sem{0.3} & 19.2/\textbf{18.2}\Tstrut \\
		\ghnbase + LayerNorm & \xmark & \xmark & \xmark & & 1.6M & 0.6 & & 50.1\sem{0.5} & 58.9 & & 43.8\sem{1.4} & 47.5\sem{0.8} & 50.8\sem{1.0} & 11.4\sem{0.2} & 49.2/16.3\\ 
		\bottomrule
	\end{tabular}
\end{table}

\subsubsection{Generalization Properties\label{apdx:gener}}
\begin{figure}[tbhp]
	\centering
	\setlength{\tabcolsep}{1pt}
	\begin{tabular}{cccc}
		{\includegraphics[width=0.24\textwidth,trim={0.5cm 0.5cm 0.5cm 0.5cm},clip,align=c]{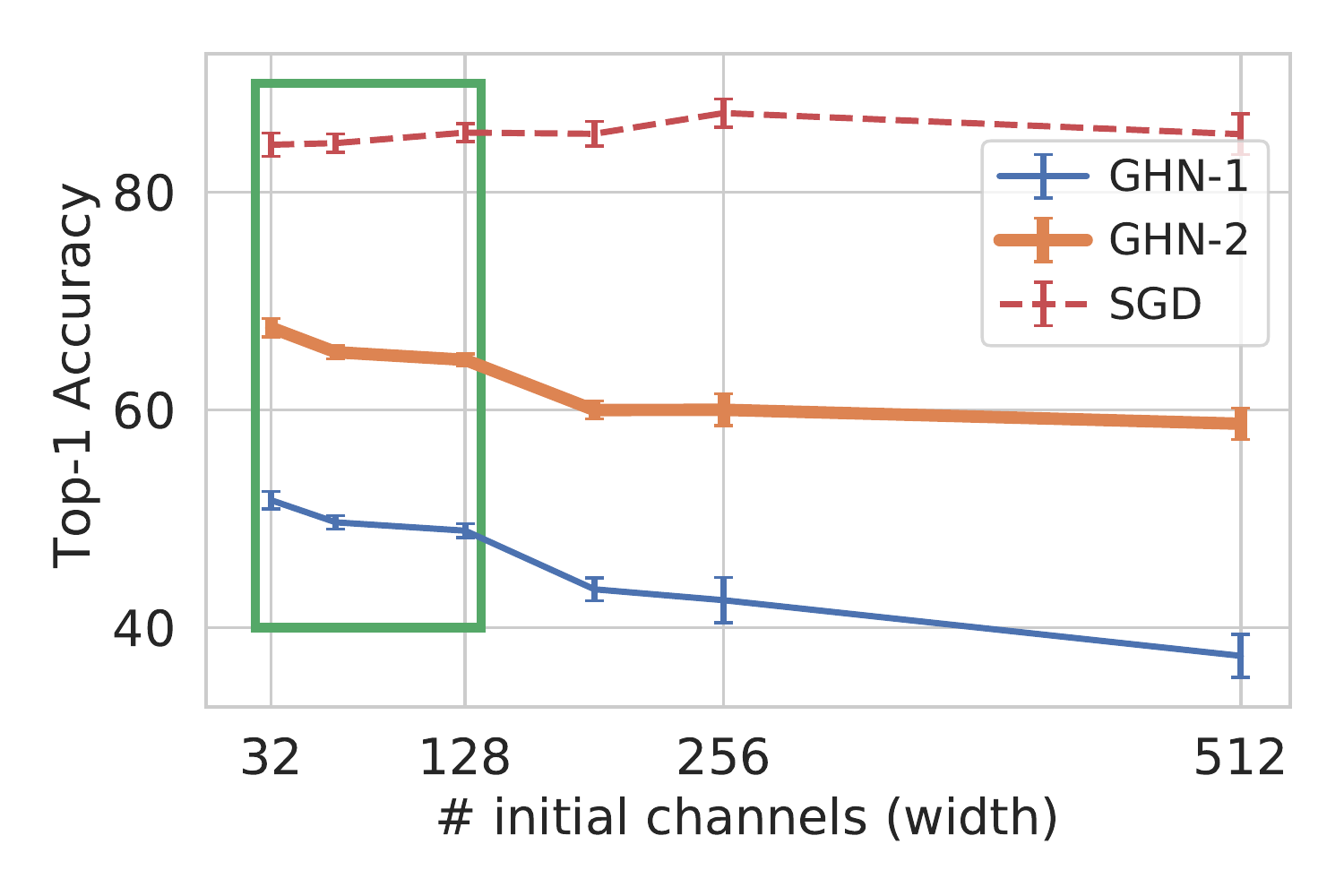}} & \includegraphics[width=0.24\textwidth,trim={0.5cm 0.5cm 0.5cm 0.5cm},clip,align=c]{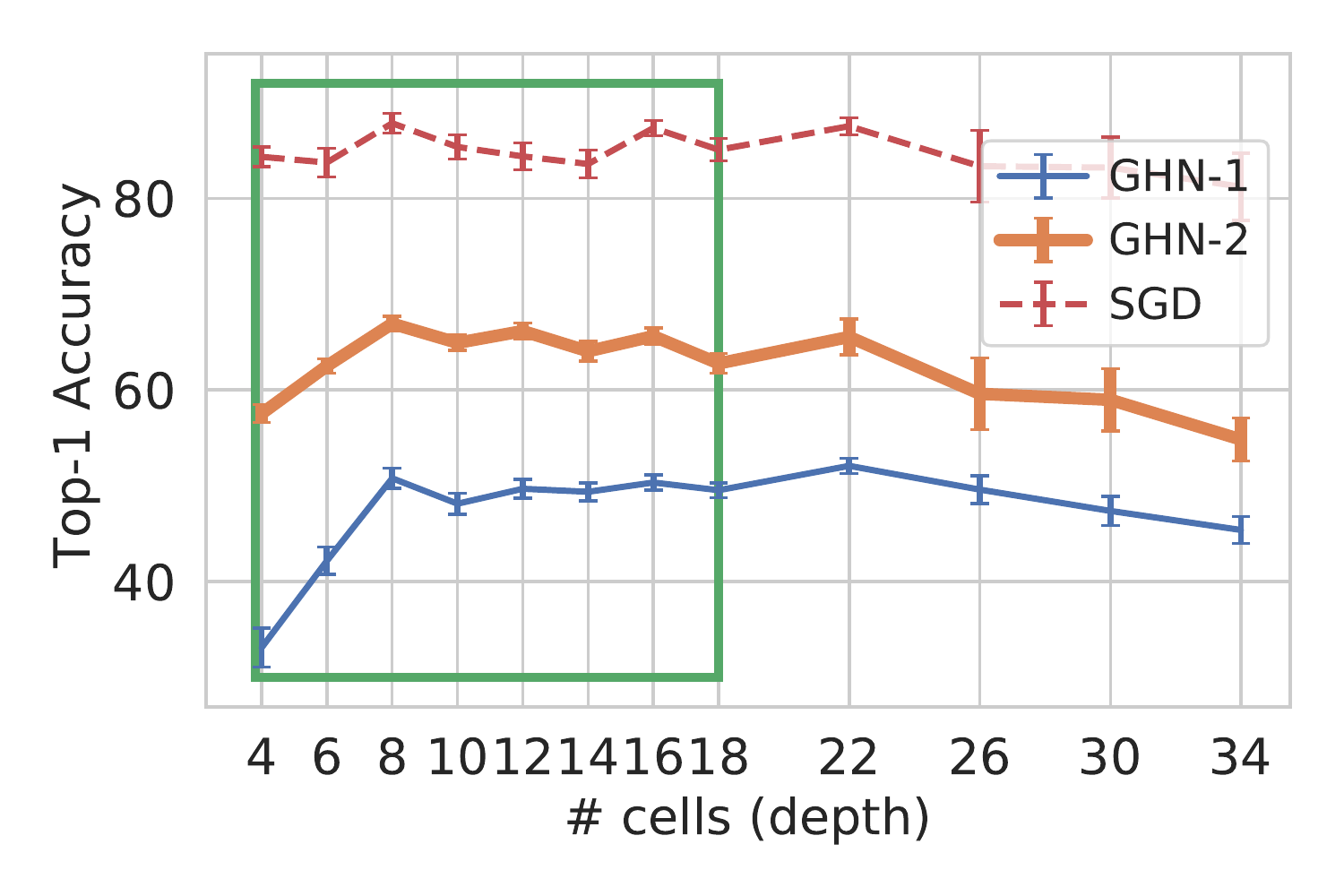} &
		\includegraphics[width=0.24\textwidth,trim={0.5cm 0.5cm 0.5cm 0.5cm},clip,align=c]{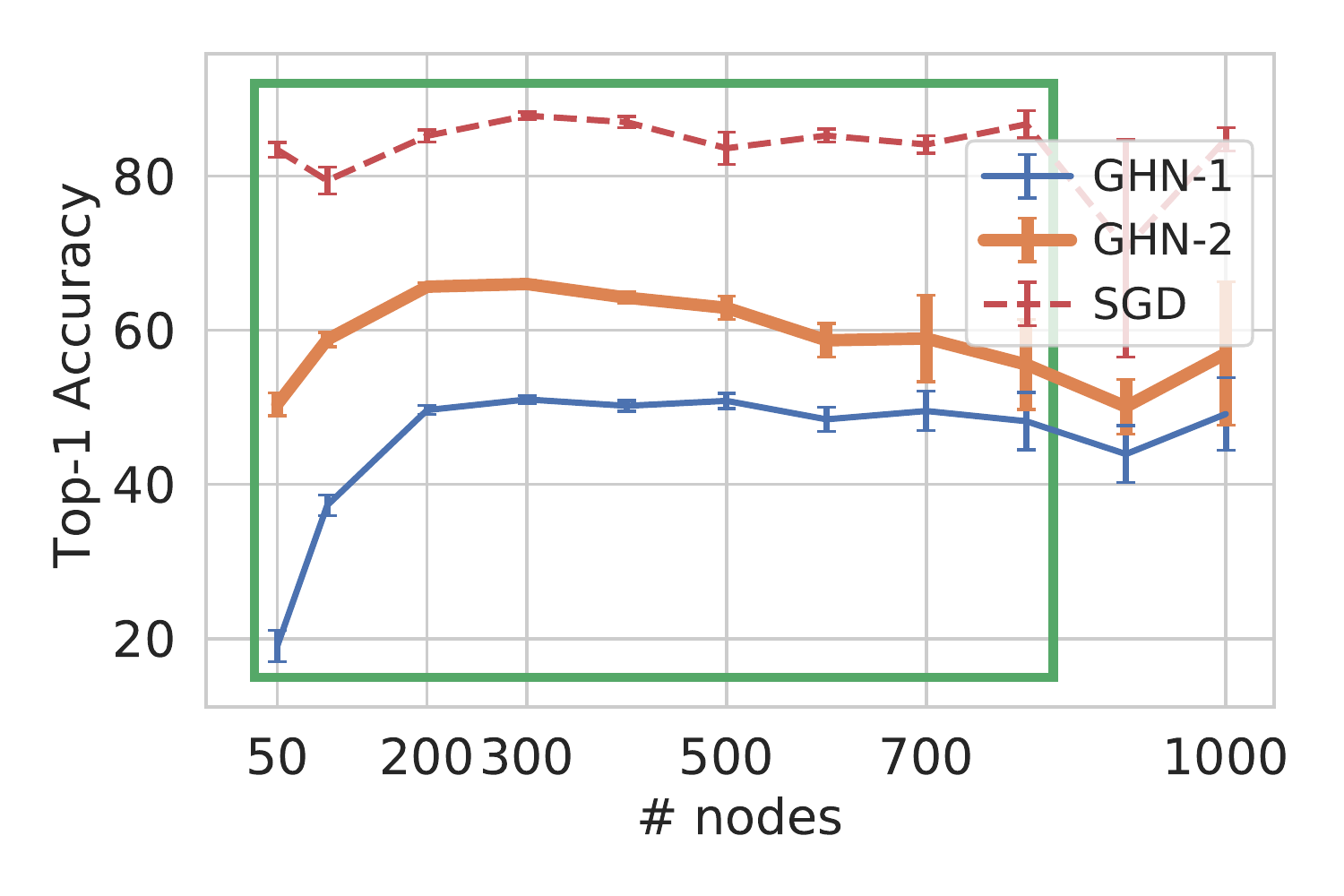} &
		\includegraphics[width=0.24\textwidth,trim={0.5cm 0.5cm 0.5cm 0.5cm},clip,align=c]{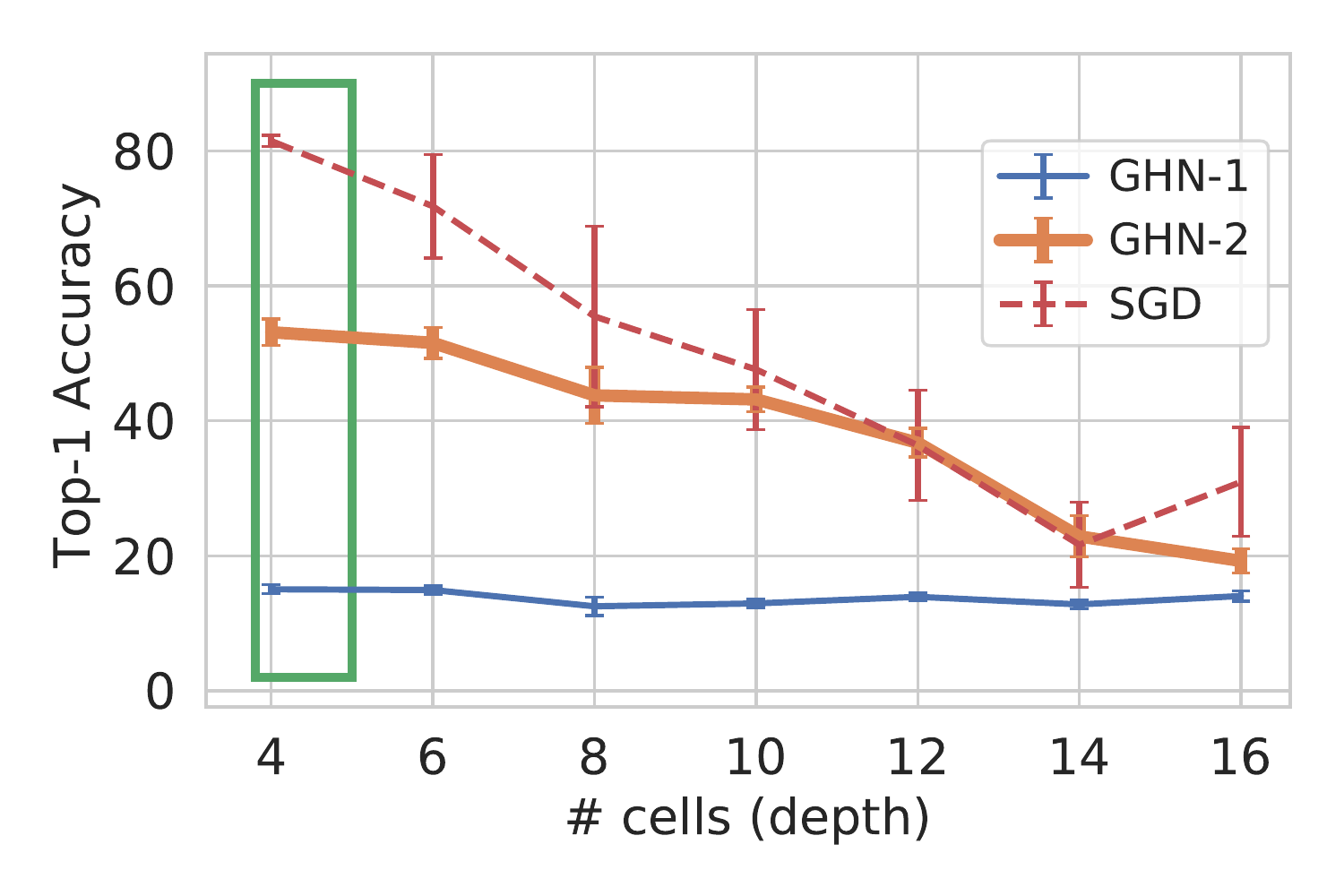} \\
		\includegraphics[width=0.24\textwidth,trim={0.5cm 0.5cm 0.5cm 0.5cm},clip,align=c]{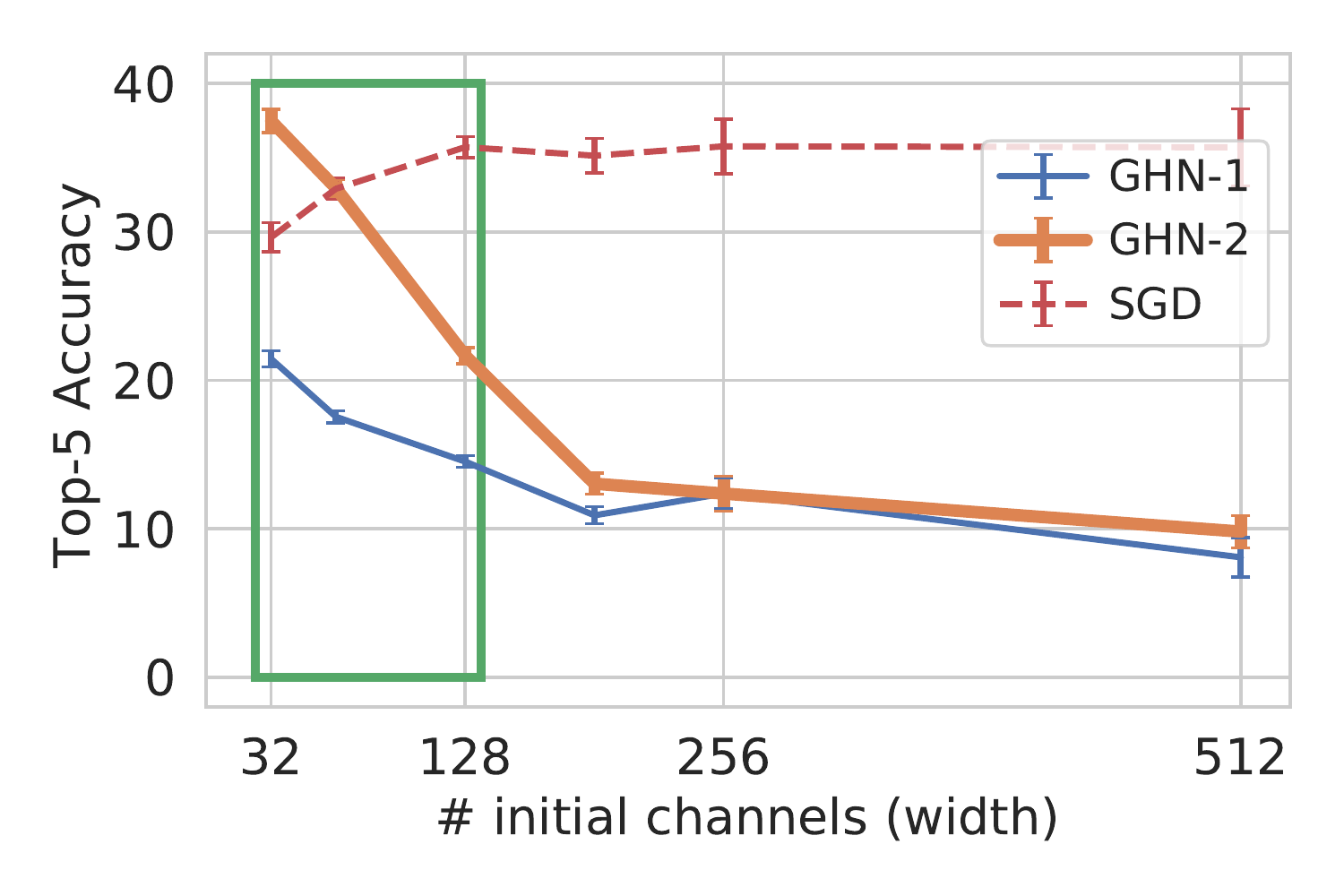} & \includegraphics[width=0.24\textwidth,trim={0.5cm 0.5cm 0.5cm 0.5cm},clip,align=c]{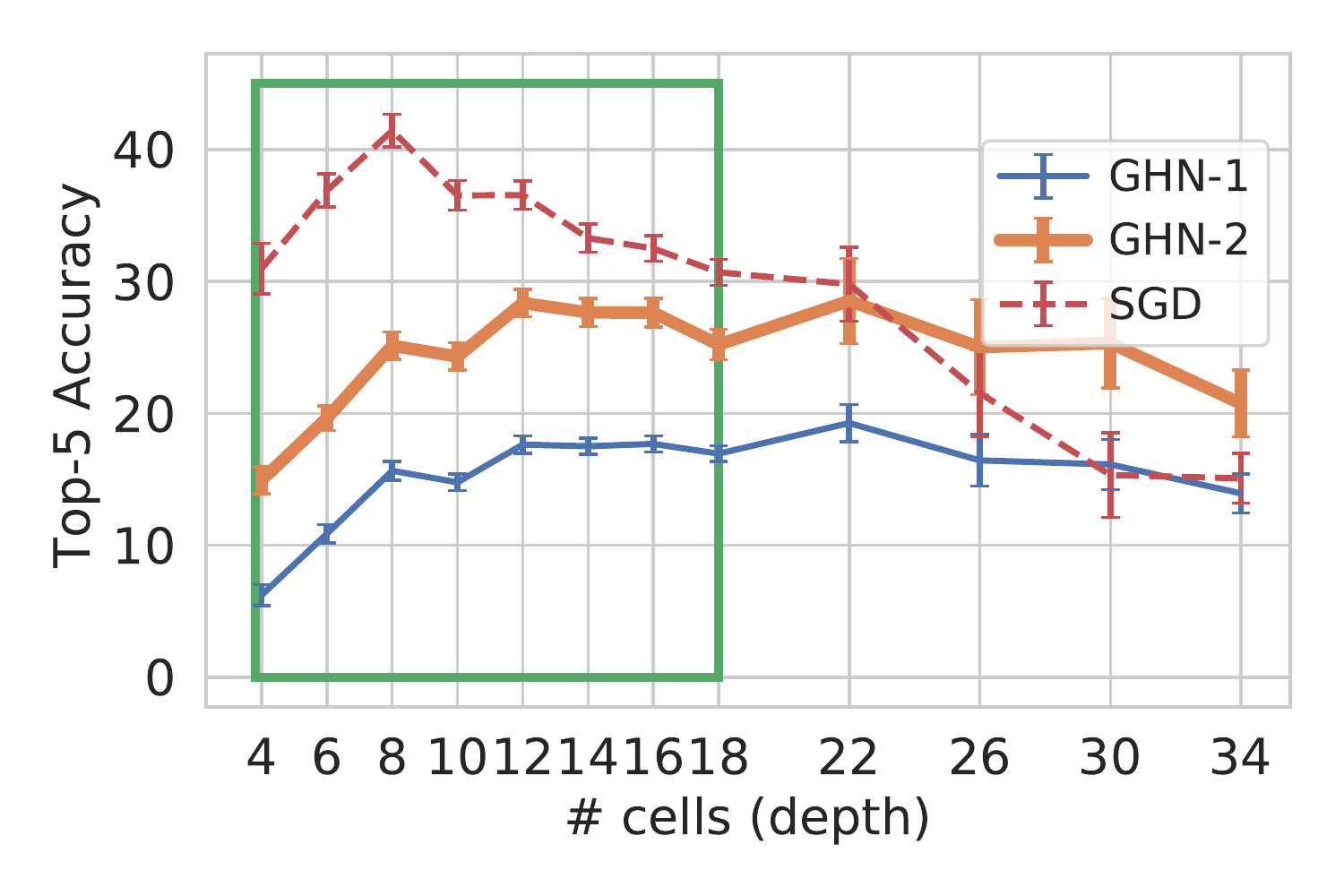} &
		\includegraphics[width=0.24\textwidth,trim={0.5cm 0.5cm 0.5cm 0.5cm},clip,align=c]{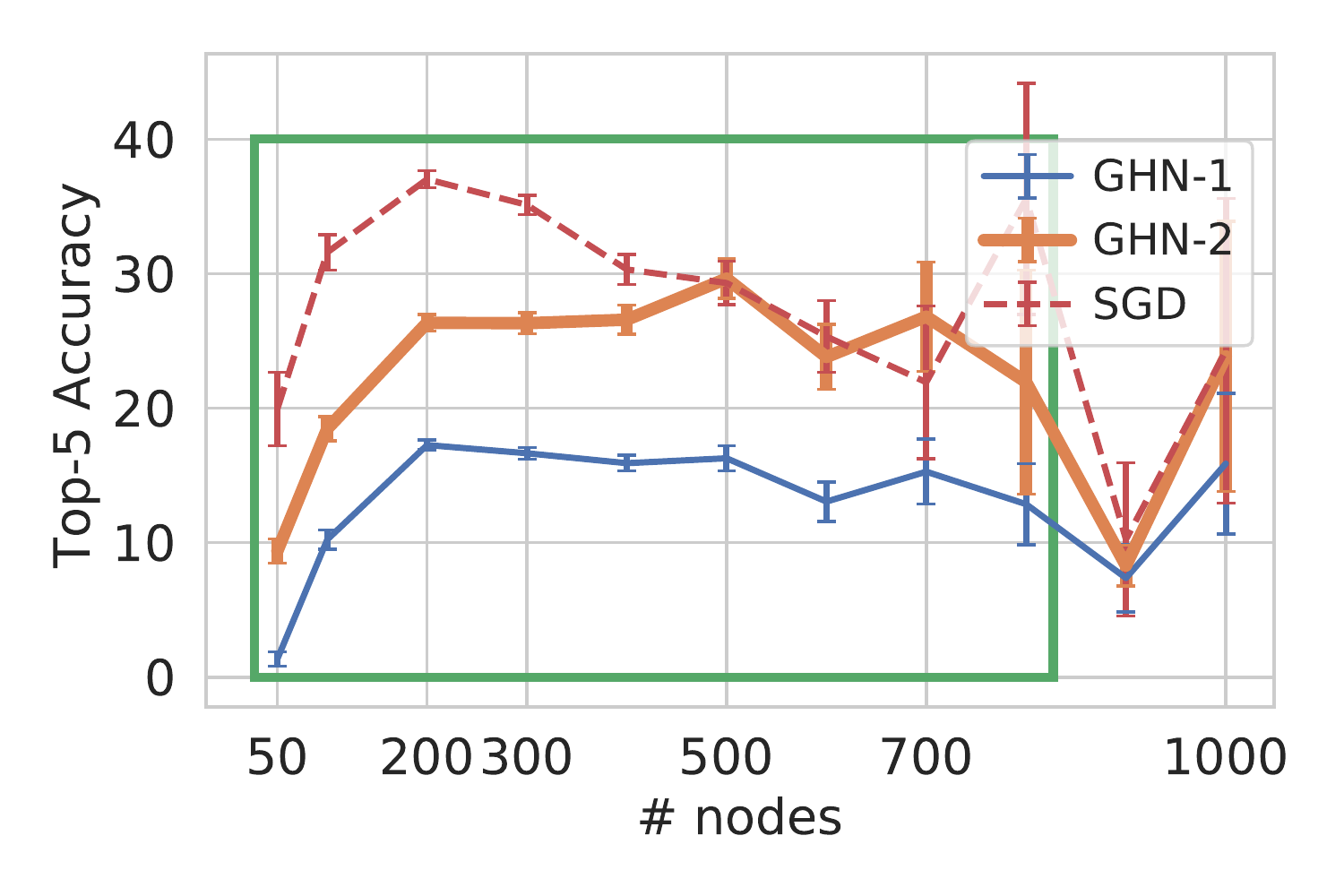} &
		\includegraphics[width=0.24\textwidth,trim={0.5cm 0.5cm 0.5cm 0.5cm},clip,align=c]{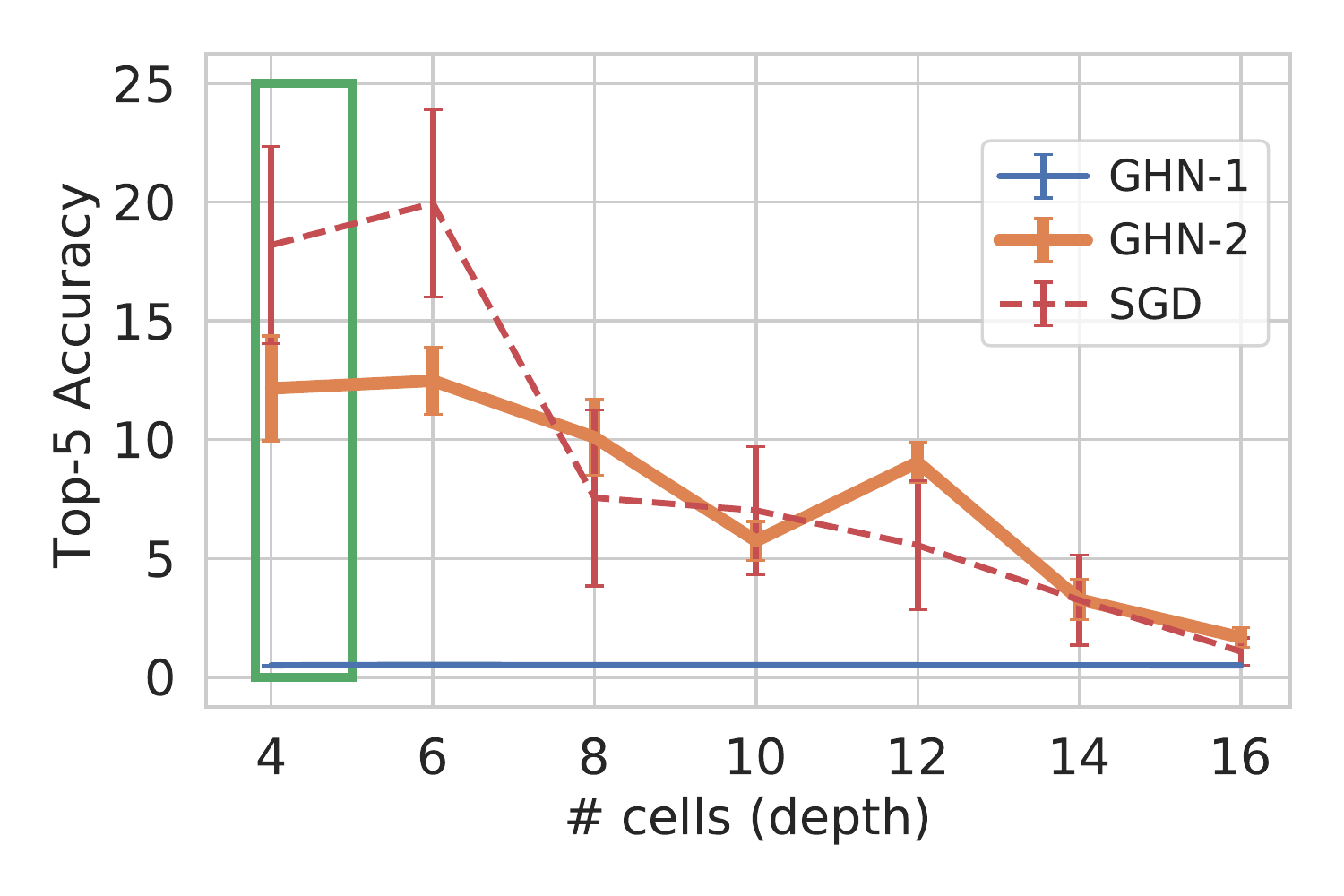} \\
	\end{tabular}
	\caption{Generalization performance w.r.t. (from left to right): width, depth, number of nodes and depths for architectures without batch norm. A green rectangle denotes the training regime; bars are standard errors of the mean. Top row: CIFAR-10, bottom row: ImageNet.}
	\label{fig:generalize_all}
\end{figure}

On the \ood subsets, GHN results are lower than on \iid-\iidtest as expected, so we inspect in more detail \textit{how} performance changes with an increased distribution shift (Fig.~\ref{fig:generalize_all}).
For example, training wider nets with SGD leads to similar or better performance, perhaps, due to increased capacity. 
However, GHNs degrade with larger width, since wide architectures are underrepresented during training for computational reasons (Table~\ref{tab:graphs}, Fig.~\ref{fig:vis_stats}). As for the depth and number of nodes, there is a certain range of values (``sweet spot'') with higher performance. For architectures without batch norm (Fig.~\ref{fig:generalize_all}, the very right column), the results of \ghnours are strong starting from a certain depth ($\geq$ 8-10 cells) matching the ones of training with SGD from scratch (for 50 epochs on CIFAR-10 and 1 epoch on ImageNet). This can be explained by the difficulty of training models without BN from scratch with SGD, while parameter prediction with \ghnours is less affected by that.
Generalization appears to be worse on ImageNet, perhaps due to its more challenging nature.

\subsubsection{Property Prediction\label{apdx:prop}}

In \S~\ref{sec:prop_pred}, we experiment with four properties of neural architectures that can be estimated given an architecture and image dataset:
\begin{enumerate}
	\item ``Clean'' classification accuracy measured on the validation sets of CIFAR-10 and ImageNet.
	\item Classification accuracy on the corrupted images, which is created by adding the Gaussian noise of medium intensity to the validation images following~\cite{hendrycks2019benchmarking}\footnote{https://github.com/hendrycks/robustness}: with zero mean and the standard deviation of 0.08 on CIFAR-10 and 0.18 on ImageNet.
	\item The inference speed is measured as GPU seconds required to process the entire validation set.\looseness-1
	\item For the convergence speed, we measure the number of SGD iterations to achieve a training top-1 accuracy of 95\% on CIFAR-10 and top-5 accuracy of 10\% on ImageNet.
\end{enumerate}

\begin{figure}[tbhp]
	\centering
	\vspace{-10pt}
	\small
	\setlength{\tabcolsep}{0pt}
	\begin{tabular}{ccc}
		{\includegraphics[width=0.33\textwidth,align=c,trim={0 0cm 0 0cm},clip]{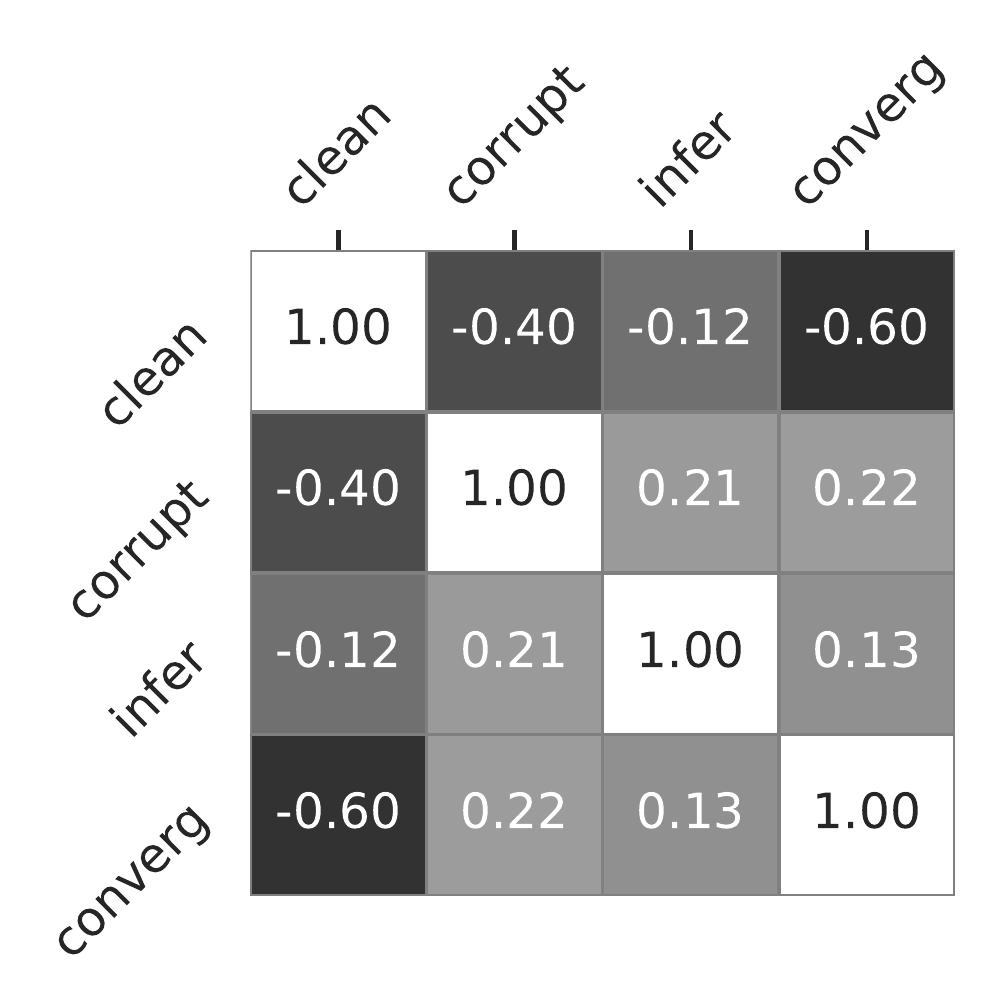}} & {\includegraphics[width=0.33\textwidth,align=c,trim={0 0cm 0 0cm},clip]{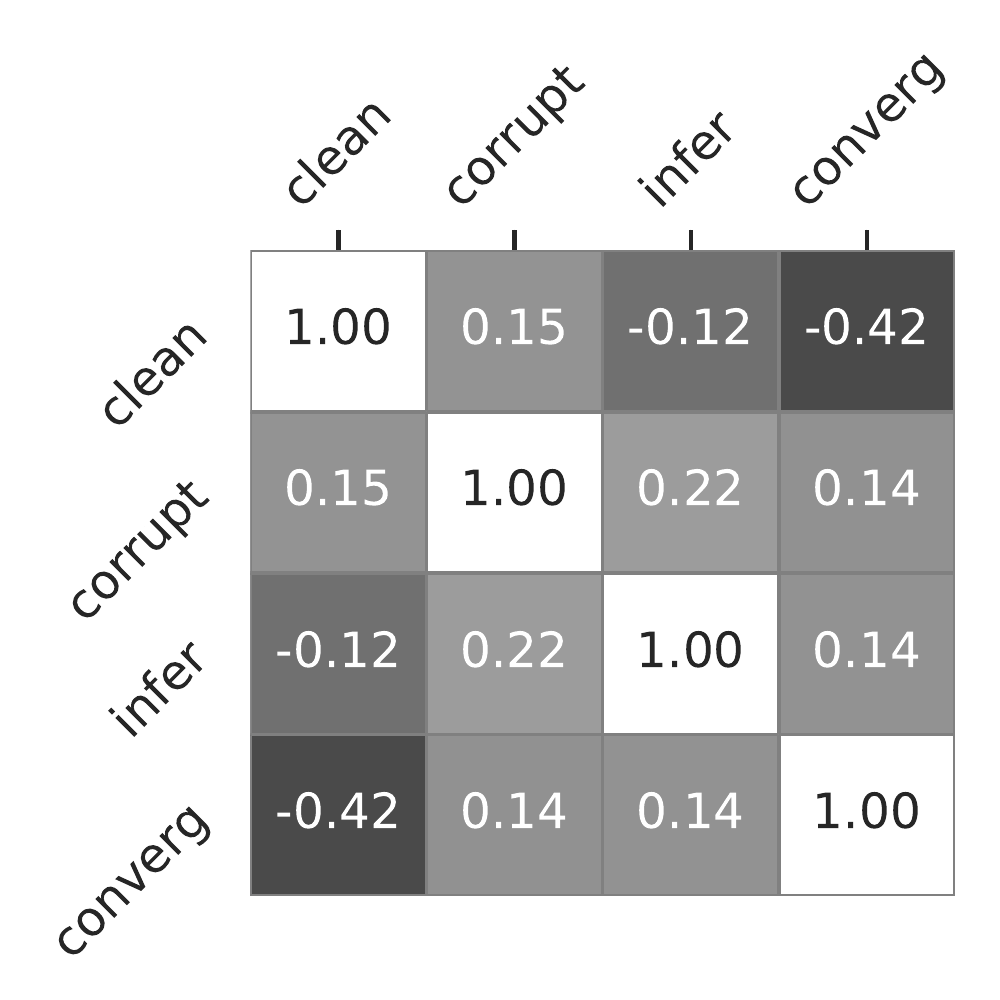}} &
		{\includegraphics[width=0.33\textwidth,align=c,trim={0 0cm 0 0cm},clip]{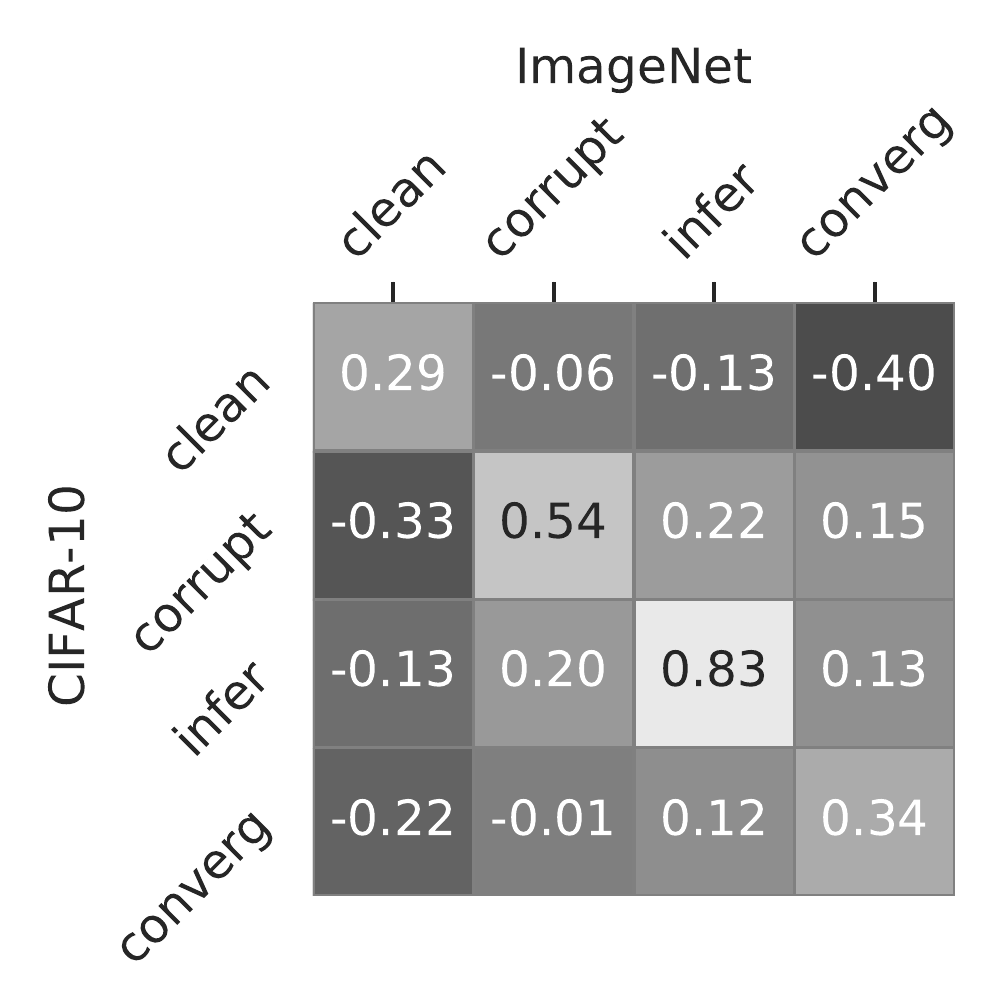}}\\
		\textbf{CIFAR-10} & \textbf{ImageNet} & \textbf{CIFAR-10 vs ImageNet} \vspace{0pt}\\
	\end{tabular}
	\vspace{-5pt}
	\caption{Cross correlation (Kendall's Tau) between ground truth values of properties.}
	\label{fig:properties_cross}
	\vspace{-5pt}
\end{figure}

\paragraph{Ground truth property values.} We first obtain ground truth values for each property for each of the 500 \iid-\iidval and 500 \iid-\iidtest architectures of \dataset trained from scratch with SGD for 50 epochs (on CIFAR-10) and 1 epoch (on ImageNet) as described in \S~\ref{sec:exper}. The ground truth values between these properties and between CIFAR-10 and ImageNet correlate poorly (Fig.~\ref{fig:properties_cross}). Interestingly, on CIFAR-10 the architectures that rank higher on the clean images generally rank lower on the corrupted set and vice verse (correlation is -0.40). On ImageNet the correlation between these two properties is positive, but low (0.15). Also, the transferability of architectures between CIFAR-10 and ImageNet is poor contrary to a common belief, e.g. with the correlation of 0.29 on the clean and -0.06 on the corrupted sets. However, this might be due to training on ImageNet only for 1 epoch. The networks that converge faster have generally worse performance on the clean set, but tend to perform better on the corrupted set. The networks that classify images faster (have higher inference speed), also tend to perform better on the corrupted set. Investigating the underlying reasons for these relationships as well as extending the set of properties would be interesting in future work.

\paragraph{Estimating properties by predicting parameters.}  A straightforward way to estimate this kind of properties using GHNs is by predicting the architecture parameters and forward passing images as was done in~\cite{zhang2018graph} for accuracy. 
However, this strategy has two issues: (a) the performance of parameters predicted by GHNs is strongly affected by the training distribution of architectures (Fig.~\ref{fig:generalize_all});
(b) estimating properties of thousands of networks for large datasets such as ImageNet can be time consuming.
Regarding (a), for example the rank correlation on CIFAR-10 between the accuracy of the parameters predicted by \ghnours and those trained with SGD is only 0.4 (down from 0.8 obtained with the regression model, Fig.~\ref{fig:properties}). 

\paragraph{Training regression models.} To report the results in Fig.~\ref{fig:properties}, we treat the 500 \iid-\iidval architectures of \dataset as the training ones in this experiment. We train a simple regression model for each property using graph embeddings (obtained using MLP, \ghnbase or \ghnours) and ground truth property values of these architectures. We use Support Vector Regression\footnote{https://scikit-learn.org/stable/modules/generated/sklearn.svm.SVR.html} with the RBF kernel and tune hyperparameters (C, gamma, epsilon) on these 500 architectures using 5-fold cross-validation. We then estimate the properties given a trained regression model on the 500 \iid-\iidtest architectures of \dataset and measure Kendall's Tau rank correlation with the ground truth test values. We repeat the experiment 5 times with different random seeds, i.e. different cross-validation folds. In Fig.~\ref{fig:properties}, we show the average and standard deviation of correlation results across 5 runs.
To train the graph convolutional network of Neural Predictor ~\cite{wen2020neural}, we use the open source implementation\footnote{\url{https://github.com/ultmaster/neuralpredictor.pytorch}} and train it on the 500 validation architectures from scratch for each property.

\paragraph{Downstream results.} Next, we verify if higher correlations translate to downstream gains. We consider the clean accuracy on CIFAR-10 in this experiment as an example. We retrain the regression model on the graph embeddings of the combined \iid-\iidval and \iid-\iidtest sets and generate 100k new architectures (similarly to how \iid-\iidtest are generated) to pick the most accurate one according to the trained regression model. We train the chosen architecture from scratch following~\cite{liu2018darts,zhang2018graph,chen2019progressive}, i.e. with SGD for 600 epochs using cutout augmentation (C), an auxiliary classifier (A) and the drop path (D) regularization. In addition, we train the chosen architecture for just 50 epochs using the same hyperparameters we used to train our \iid-\iidval and \iid-\iidtest architectures as in \S~\ref{sec:our_task}. We train the chosen architecture 5 times using 5 random seeds and report an average and standard deviation of the final classification accuracy on the test set of CIFAR-10 (Table~\ref{tab:nas_results}). We perform this experiment for \ghnbase and \ghnours in the same way.
Among the methods we compare in Table~\ref{tab:nas_results}, our \ghnours-based search finds the most performant network if training is done for 50 epochs (without and with C, A and D) and finds a competitive network if training is done for 600 epochs with C, A and D.

\begin{table}[tbhp]
	\centering
	\caption{CIFAR-10 best architectures and their performance on the test set. C --- cutout augmentation, A --- auxiliary classifier, D --- drop path regularization. The best result in each row is bolded.}
	\label{tab:nas_results}
	\vspace{3pt}
	\small
	\setlength{\tabcolsep}{0pt}
	\begin{tabular}{lcccc}
		\toprule
		& \textbf{\ghnbase} & \textbf{\ghnours} & \textbf{DARTS}~\cite{liu2018darts} & \textbf{PDARTS}~\cite{chen2019progressive} \Bstrut\\
		\midrule
		& {\includegraphics[width=0.18\textwidth,align=c,trim={2.3cm 3cm 2.3cm 3cm},clip]{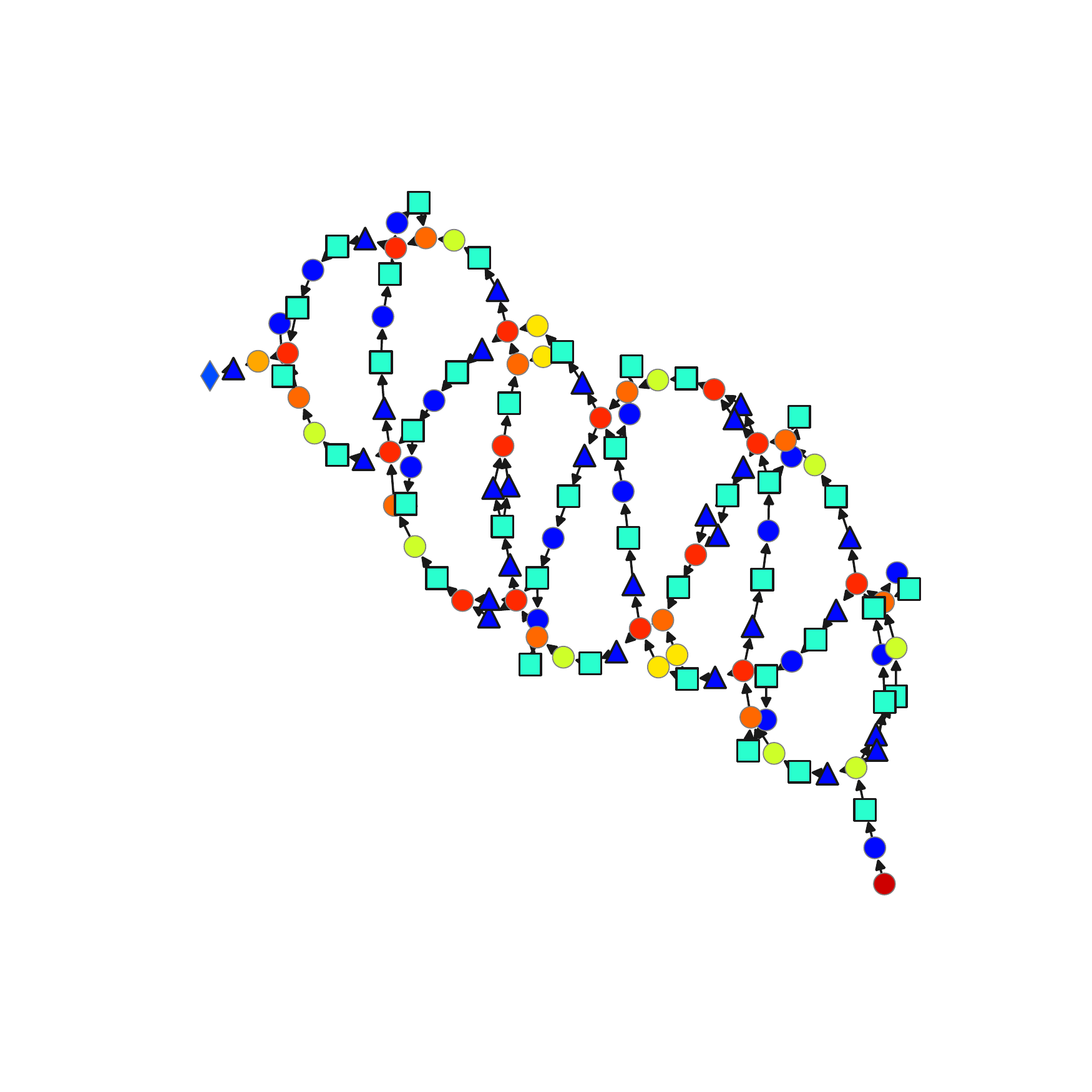}} & {\includegraphics[width=0.18\textwidth,align=c,trim={2.3cm 3cm 2.3cm 3cm},clip]{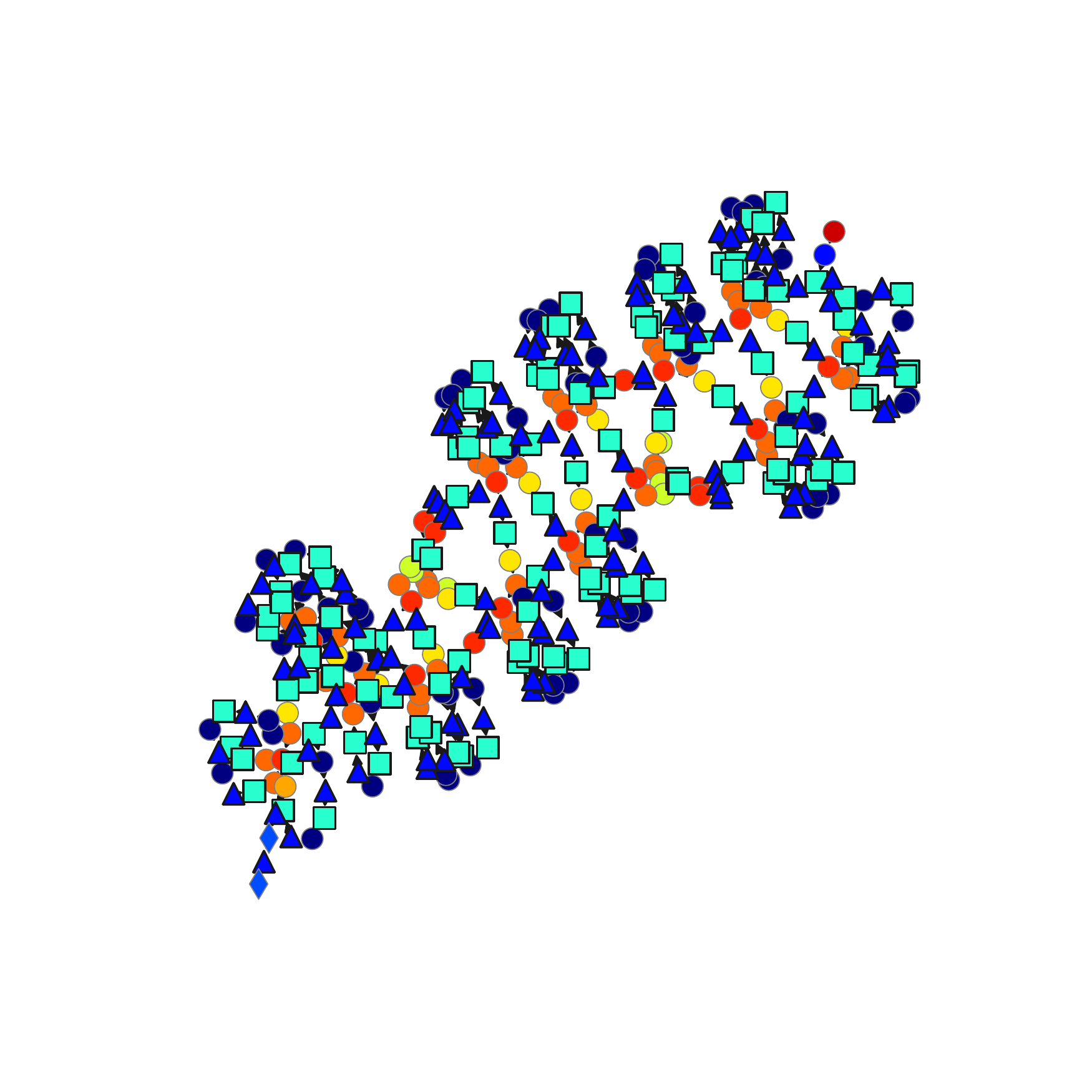}} & {\includegraphics[width=0.18\textwidth,align=c,trim={2.3cm 3cm 2.3cm 3cm},clip]{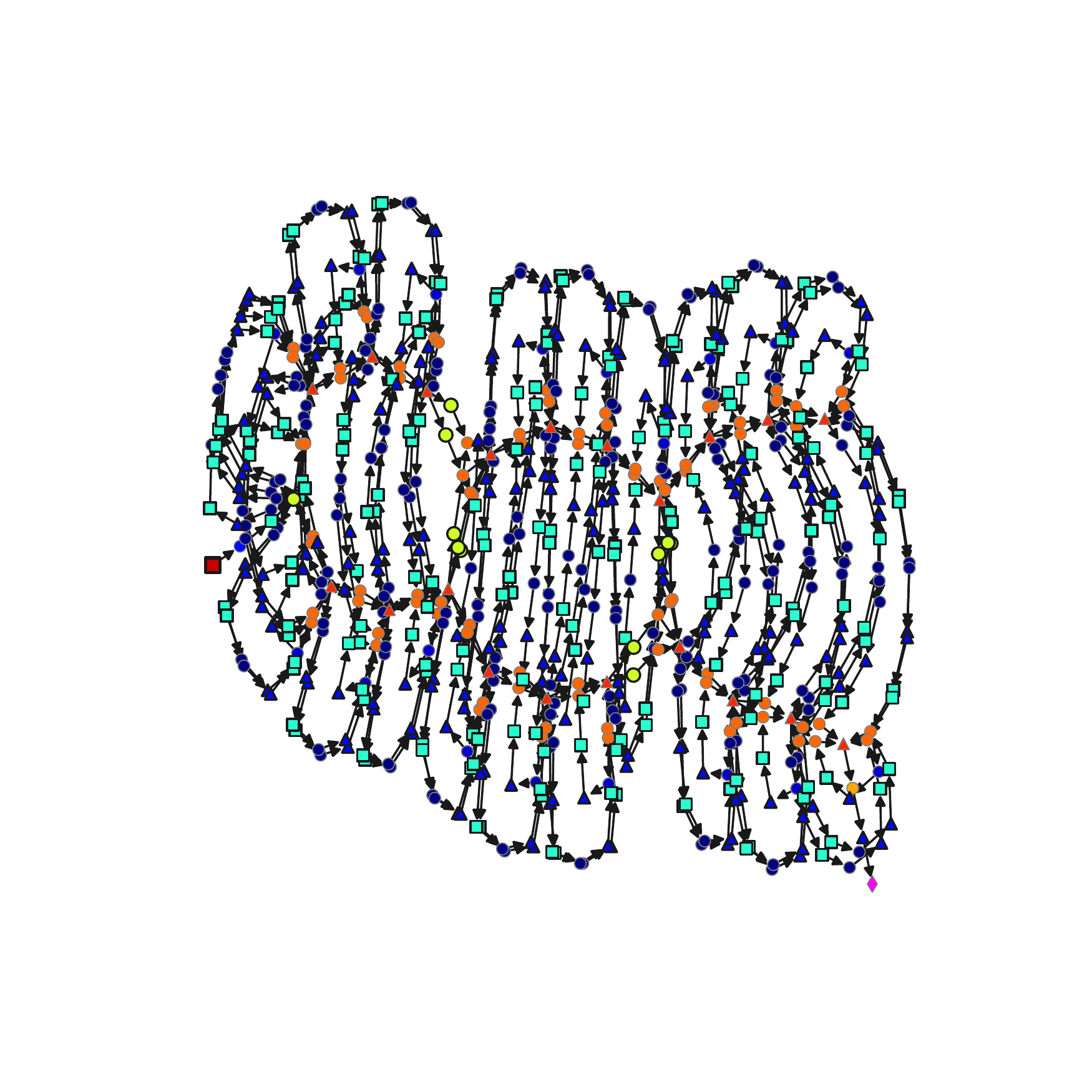}} & {\includegraphics[width=0.18\textwidth,align=c,trim={2.3cm 3cm 2.3cm 3cm},clip]{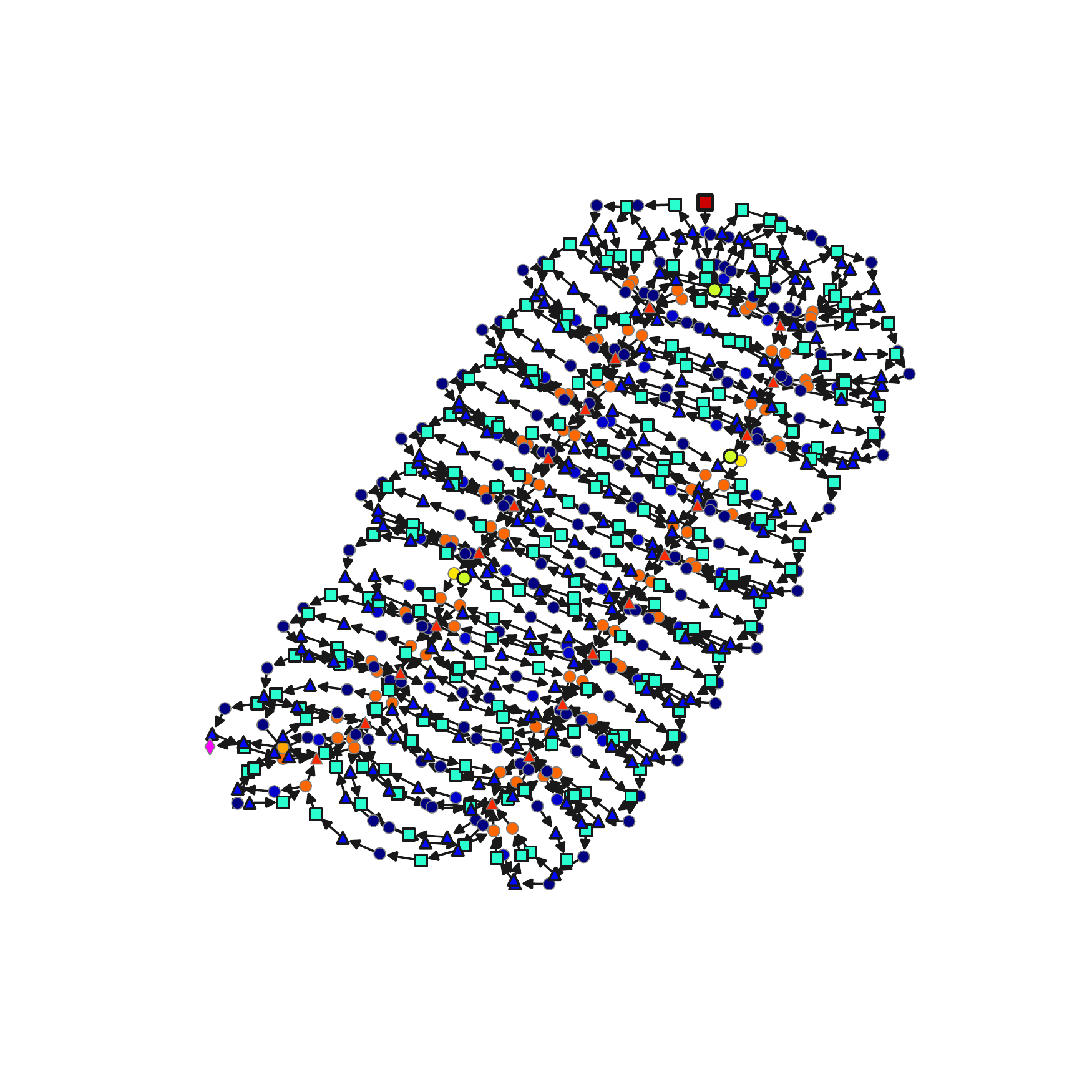}}\Tstrut\\
		\midrule
		\# params (M) &  3.1 & 3.1 & 3.3 & 3.4 \Tstrut\Bstrut\\
		50 epochs & 92.61\std{0.16} & \textbf{93.94}\std{0.11} & 93.11\std{0.09} & 92.95\std{0.14}\\
		50 epochs + C,A,D~\cite{liu2018darts} & 91.80\std{0.14} & \textbf{95.24}\std{0.14} & 94.50\std{0.08} & 94.22\std{0.06} \\
		600 epochs + C,A,D~\cite{liu2018darts} & 95.90\std{0.08} & 97.26\std{0.09} & 97.17\std{0.06} & \textbf{97.48}\std{0.06} \\
		\bottomrule
	\end{tabular}
\end{table}

\subsubsection{Comparing Neural Architectures\label{apdx:graph_compare}}

In addition to our experiments in \S~\ref{sec:prop_pred}, we further evaluate representation power of GHNs by computing the distance between neural architectures in the graph embedding space.

\textbf{Experimental setup.}
We compute the pairwise distance between the computational graphs $a_1$ and $a_2$ as the $\ell_2$ norm between their graph embedding $||\h_{a_1} - \h_{a_2}||$. 
We chose ResNet-50 as a reference network and compare its graph structure to the other three predefined architectures: ResNet-34, ResNet-50 without skip connections, and ViT. 
Among these, ViT is intuitively expected to have the largest $\ell_2$ distance, because it is mainly composed of non-convolutional operations. ResNet-50-No-Skip should be closer (in the $\ell_2$ sense) to ResNet-50 than ViT, because it is based on convolutions, but further away than ResNet-34, since it does not have skip connections. 

\textbf{Additional baseline.}
As a reference graph distance, we employ the Hungarian algorithm~\cite{kuhn1955hungarian}.
This algorithm computes the total assignment ``cost'' between two sets of nodes. It is used as an efficient approximation of the exact graph distance algorithms~\cite{ma2021deep,bai2019simgnn}, which are infeasible for graphs of the size we consider here.\looseness-1

\begin{table}[tbhp]
	\caption{Comparing ResNet-50 to the three predefined architectures using the GHNs trained on CIFAR-10 in terms of the $\ell_2$ distance between graph embeddings.}
	\label{fig:qualit}
	\vspace{5pt}
	\definecolor{bad}{rgb}{0.958, 0.788, 0.878}
	\centering
	\setlength{\tabcolsep}{0pt}
	\newcommand{\width}{0.2\textwidth}
	\scriptsize
	\renewcommand{\arraystretch}{0.8}
	\begin{tabular}{l|ccc}
		\multicolumn{1}{c|}{\scriptsize \scriptsize \textsc{ResNet-50}} & \scriptsize \textsc{ResNet-34} & \tiny \textsc{ResNet-50-No-Skip} & \scriptsize \textsc{ViT} \Bstrut\\
		\toprule
		{\includegraphics[width=\width,align=c,trim={2.3cm 3cm 2.3cm 3cm},clip]{figs/dag_resnet_50.pdf}} & 
		{\includegraphics[width=\width,align=c,trim={2.3cm 3cm 2.3cm 3cm},clip]{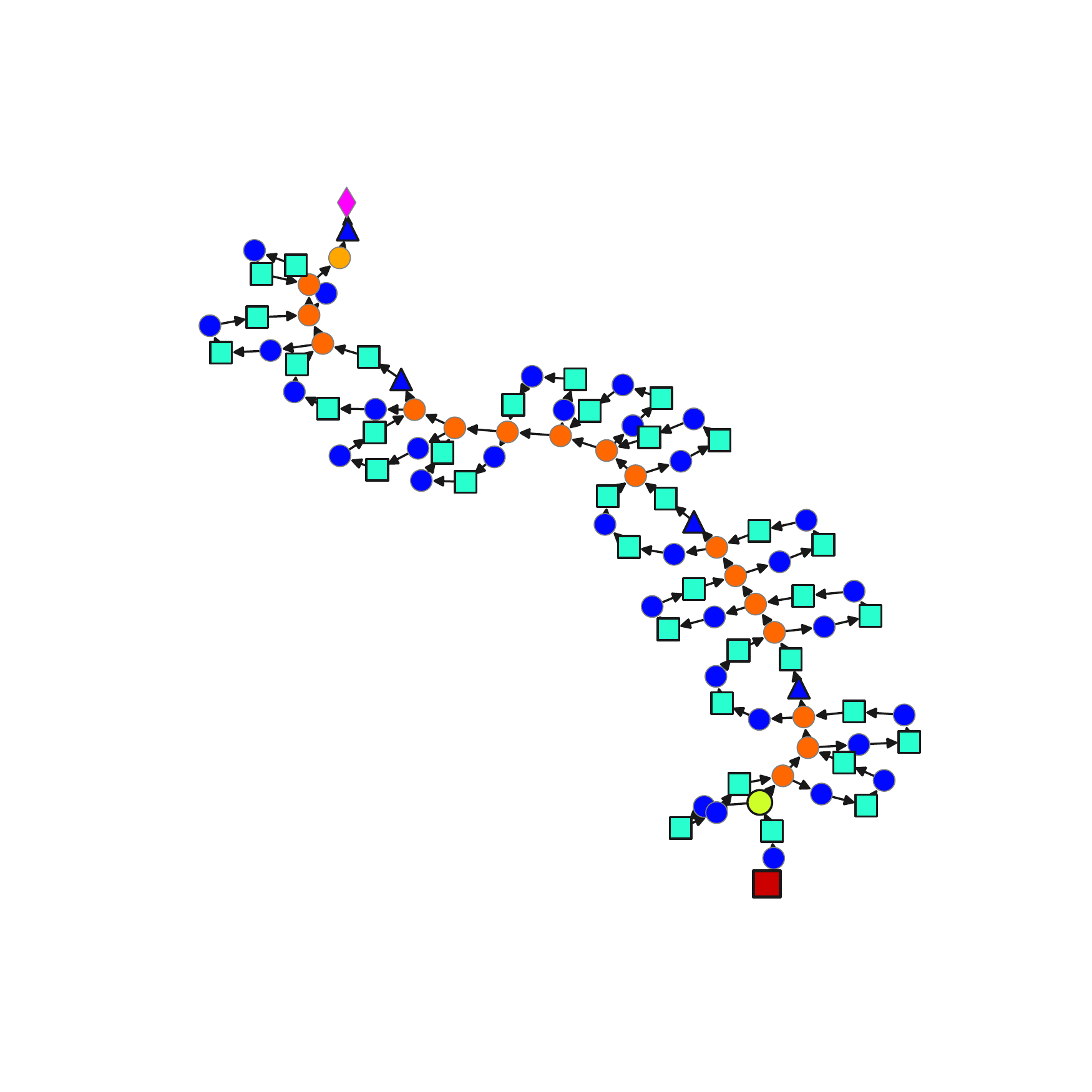}}
		& {\includegraphics[width=\width,align=c,trim={2.3cm 3cm 2.3cm 3cm},clip]{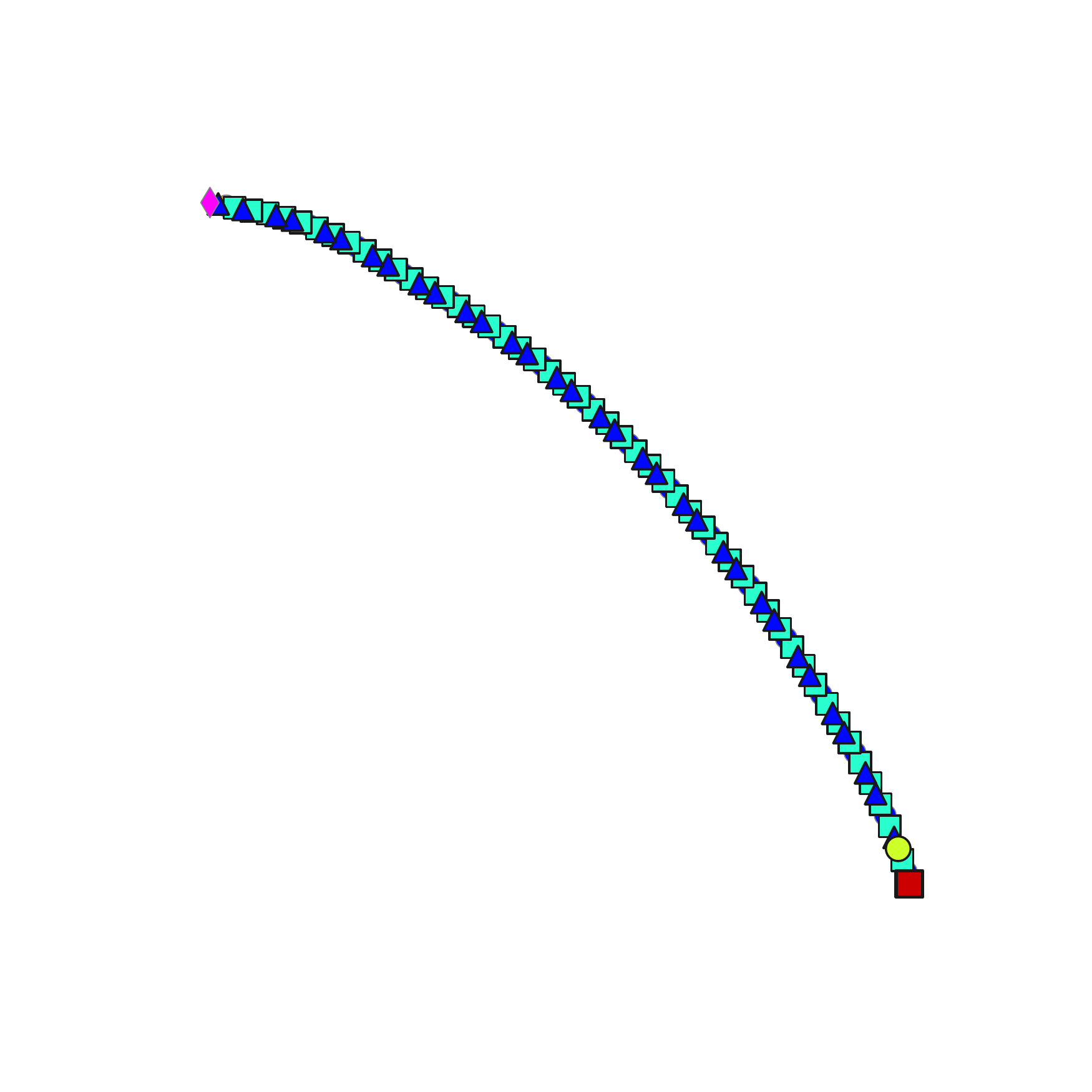}} & {\includegraphics[width=\width,align=c,trim={2.3cm 3cm 2.3cm 3cm},clip]{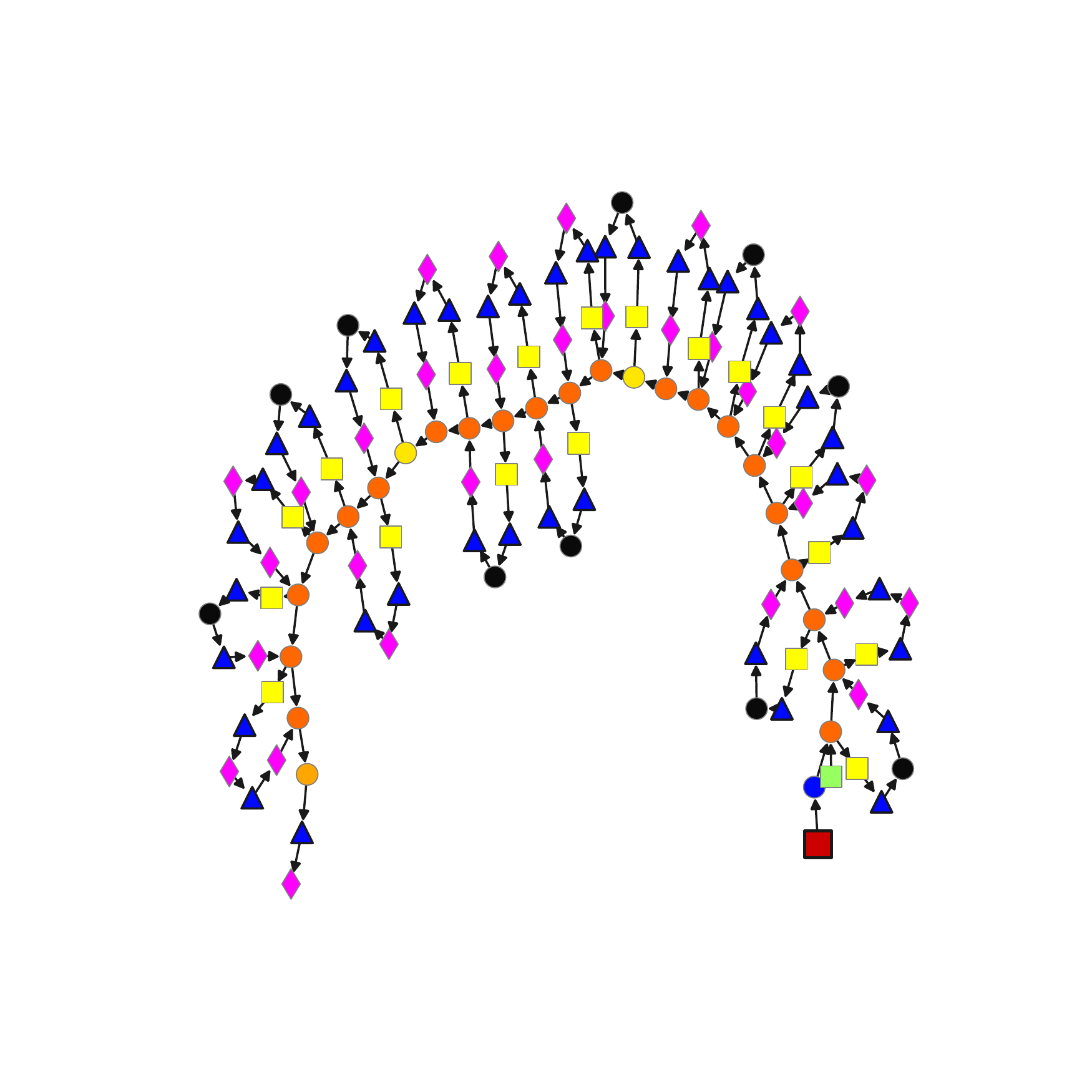}}\Tstrut \\
		\midrule		
		Hungarian & 118.0 & 134.0 & 207.5 \\
		MLP & 0.4 & \cellcolor{bad}0.2 & 1.0\\
		\ghnbase & 0.6 & \cellcolor{bad}0.5 & 1.7 \\
		\ghnours & 1.0 & 1.3 & 3.1 \\
		\bottomrule
	\end{tabular}
	
\end{table}

\textbf{Qualitative results.}
The $\ell_2$ distances computed based on \ghnours align well with our initial hypothesis of the relative distances between the architectures as well as with the Hungarian distances (Table~\ref{fig:qualit}). In contrast, the baselines inappropriately embed ResNet-50 closer to ResNet-50-No-Skip  than to ResNet-34. Thus, based on this simple evaluation, \ghnours captures graph structures better than the baselines.

\begin{figure}[htpb]
	\centering
	\scriptsize
	\vspace{0pt}
	\setlength{\tabcolsep}{1pt}
	\begin{tabular}{cc|cc}
		{\includegraphics[width=0.22\textwidth,align=c,trim={2.3cm 3cm 2.3cm 3cm},clip]{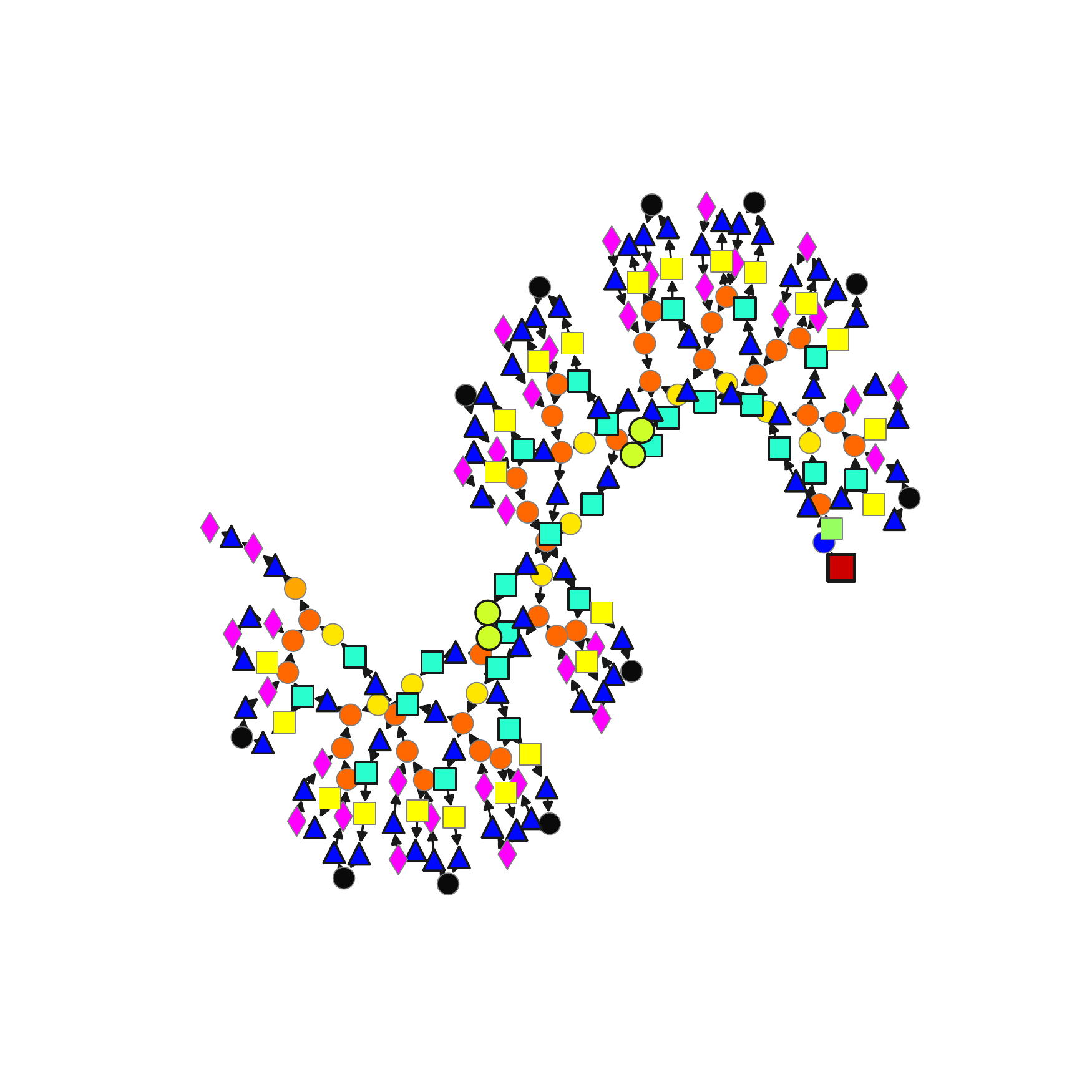}} & {\includegraphics[width=0.22\textwidth,align=c,trim={2.3cm 3cm 2.3cm 3cm},clip]{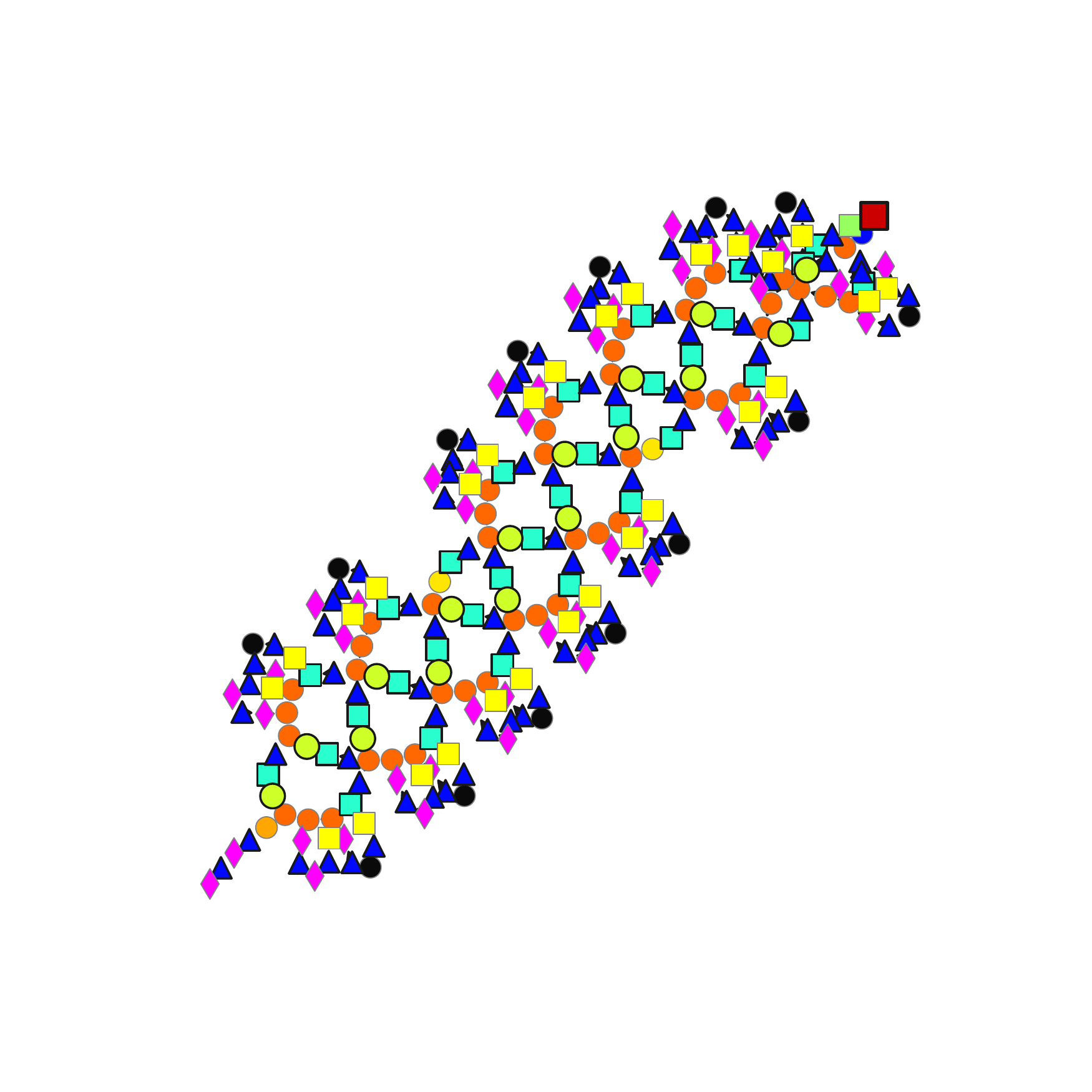}} & 
		{\includegraphics[width=0.22\textwidth,align=c,trim={2.3cm 3cm 2.3cm 3cm},clip]{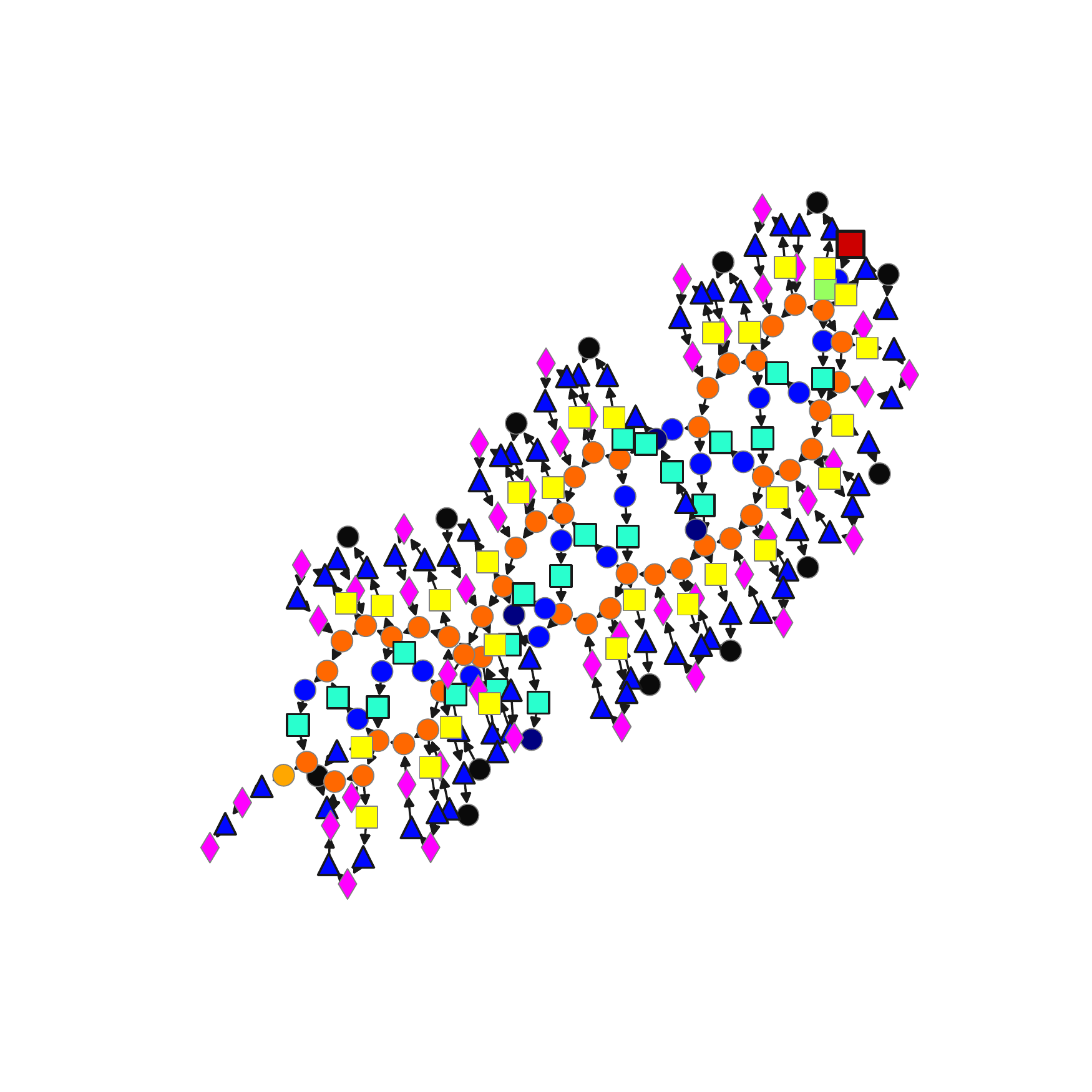}} & {\includegraphics[width=0.22\textwidth,align=c,trim={2.3cm 3cm 2.3cm 3cm},clip]{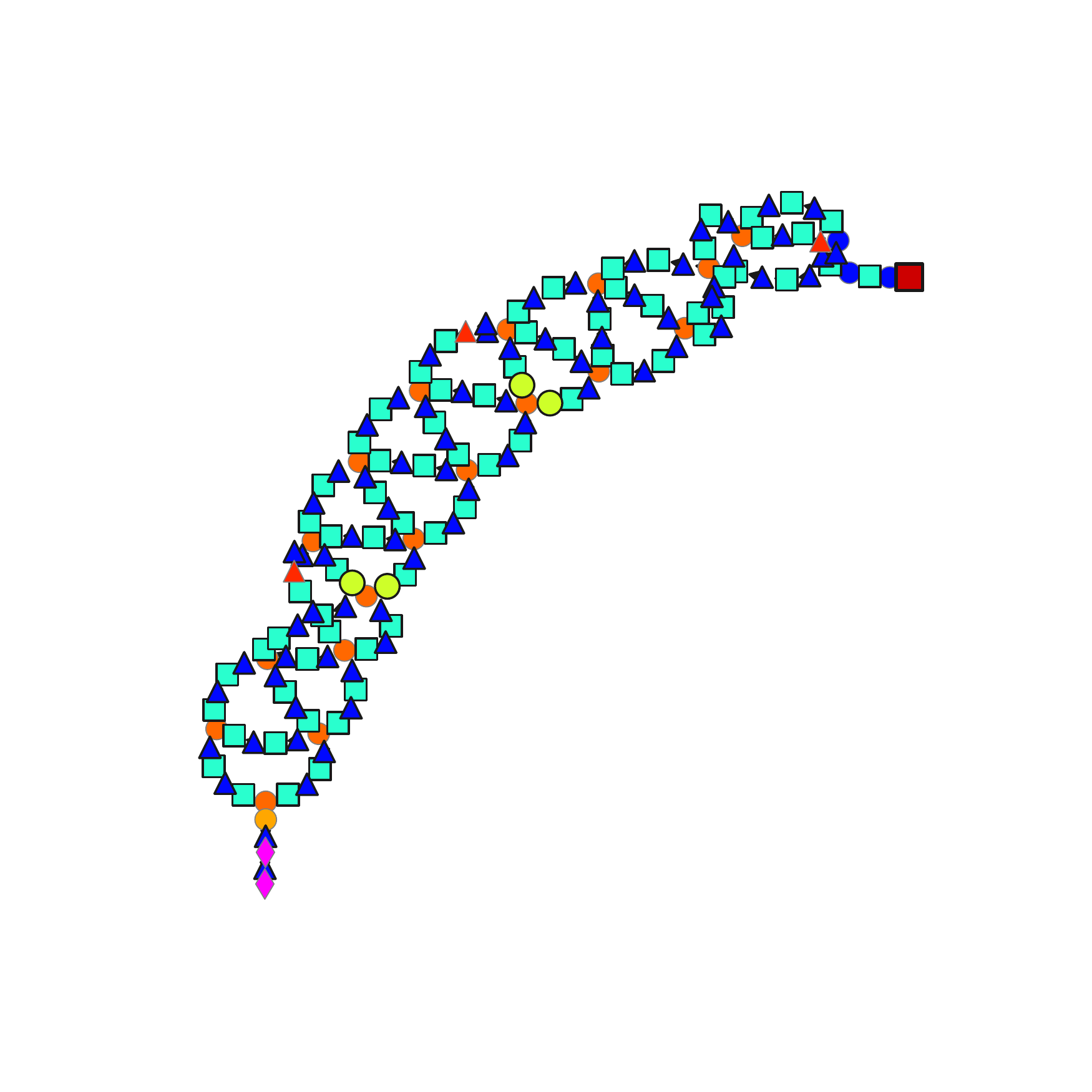}} \\
		\multicolumn{2}{c}{$\ell_2=0.02$} & \multicolumn{2}{c}{$\ell_2=2.4$} \Bstrut \\
		
		{\includegraphics[width=0.22\textwidth,align=c,trim={2.3cm 3cm 2.3cm 3cm},clip]{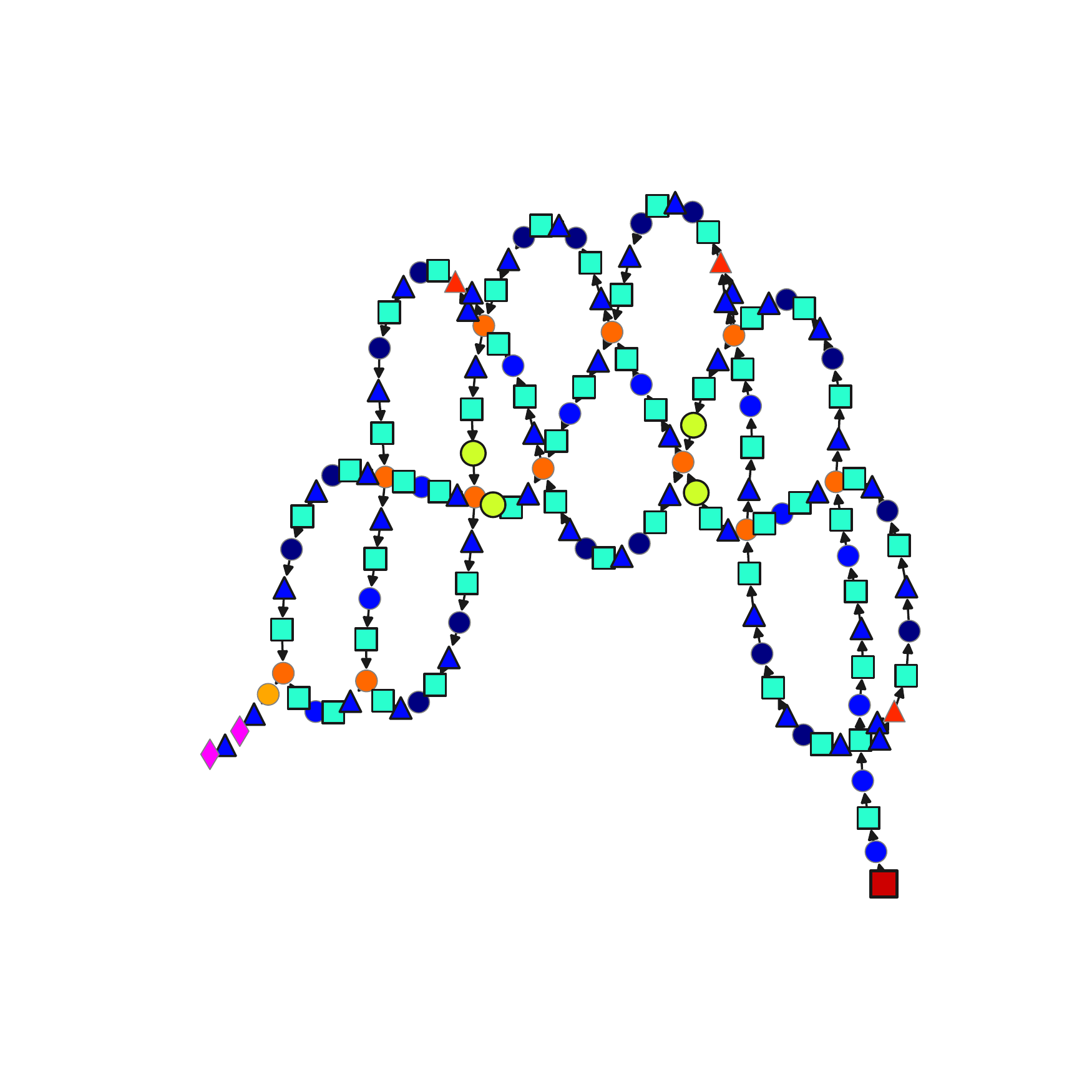}} & {\includegraphics[width=0.22\textwidth,align=c,trim={2.3cm 3cm 2.3cm 3cm},clip]{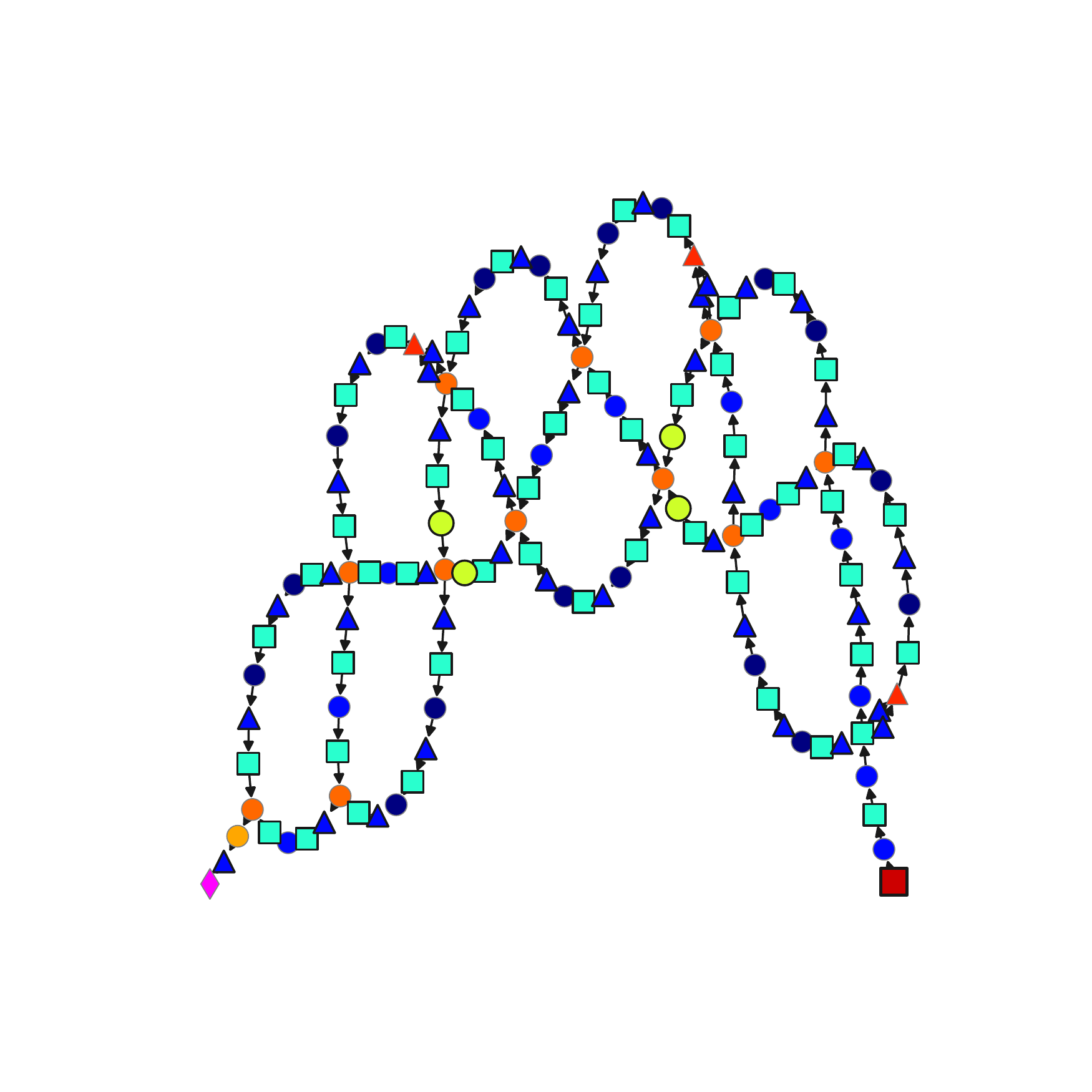}} & 
		{\includegraphics[width=0.22\textwidth,align=c,trim={2.3cm 3cm 2.3cm 3cm},clip]{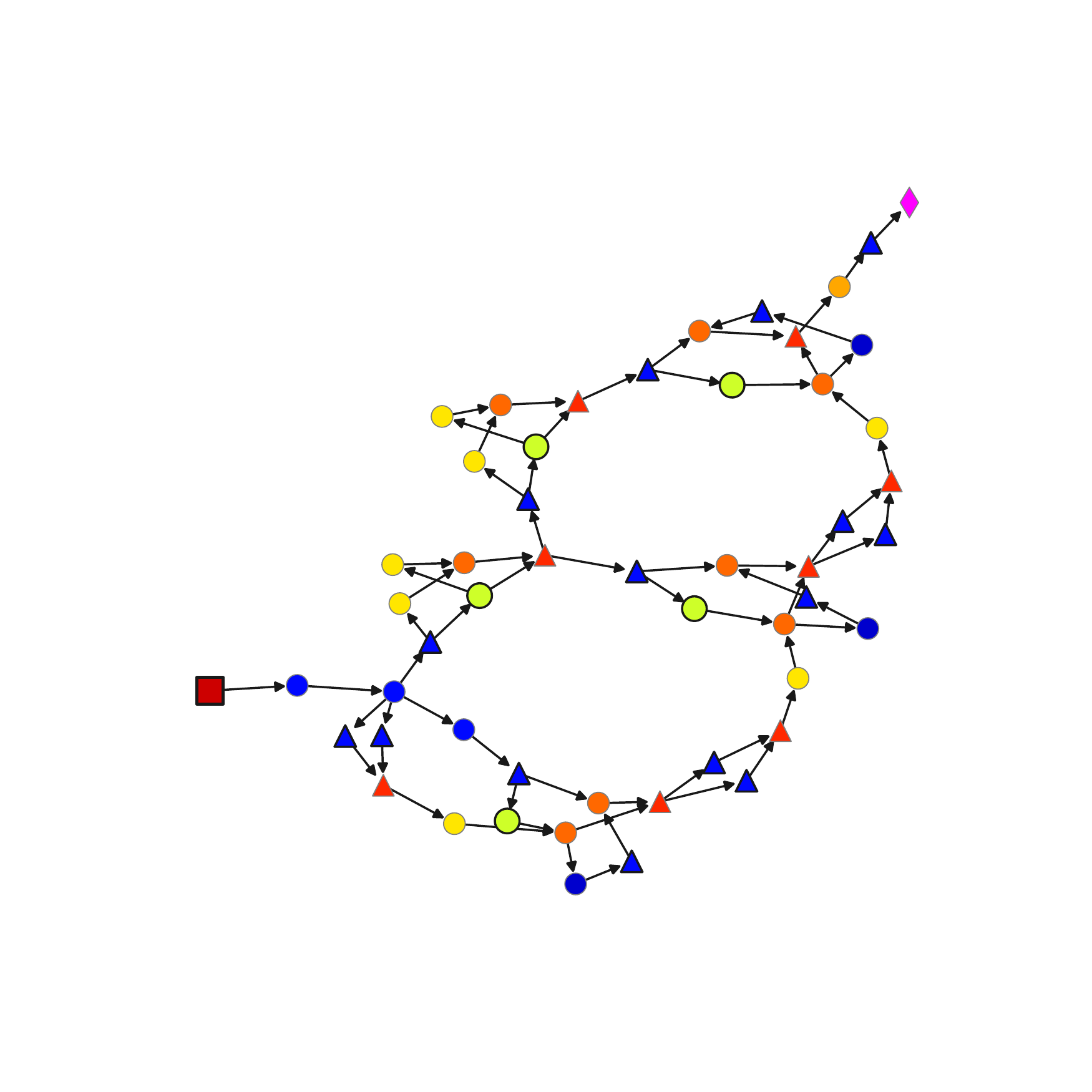}} & {\includegraphics[width=0.22\textwidth,align=c,trim={2.3cm 3cm 2.3cm 3cm},clip]{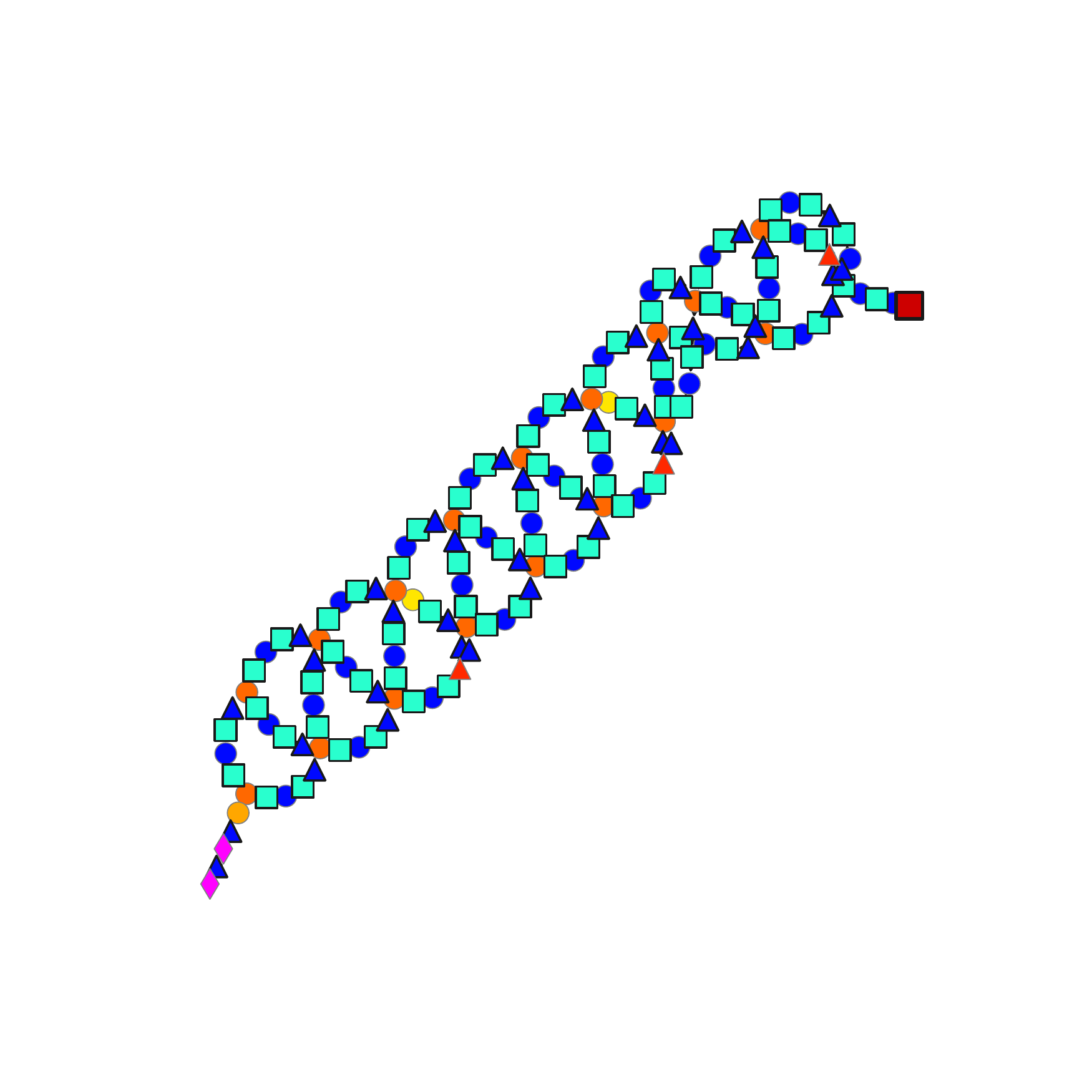}} \\
		\multicolumn{2}{c}{$\ell_2=0.03$} & \multicolumn{2}{c}{$\ell_2=3.4$} \Bstrut \\
		
		{\includegraphics[width=0.22\textwidth,align=c,trim={2.3cm 3cm 2.3cm 3cm},clip]{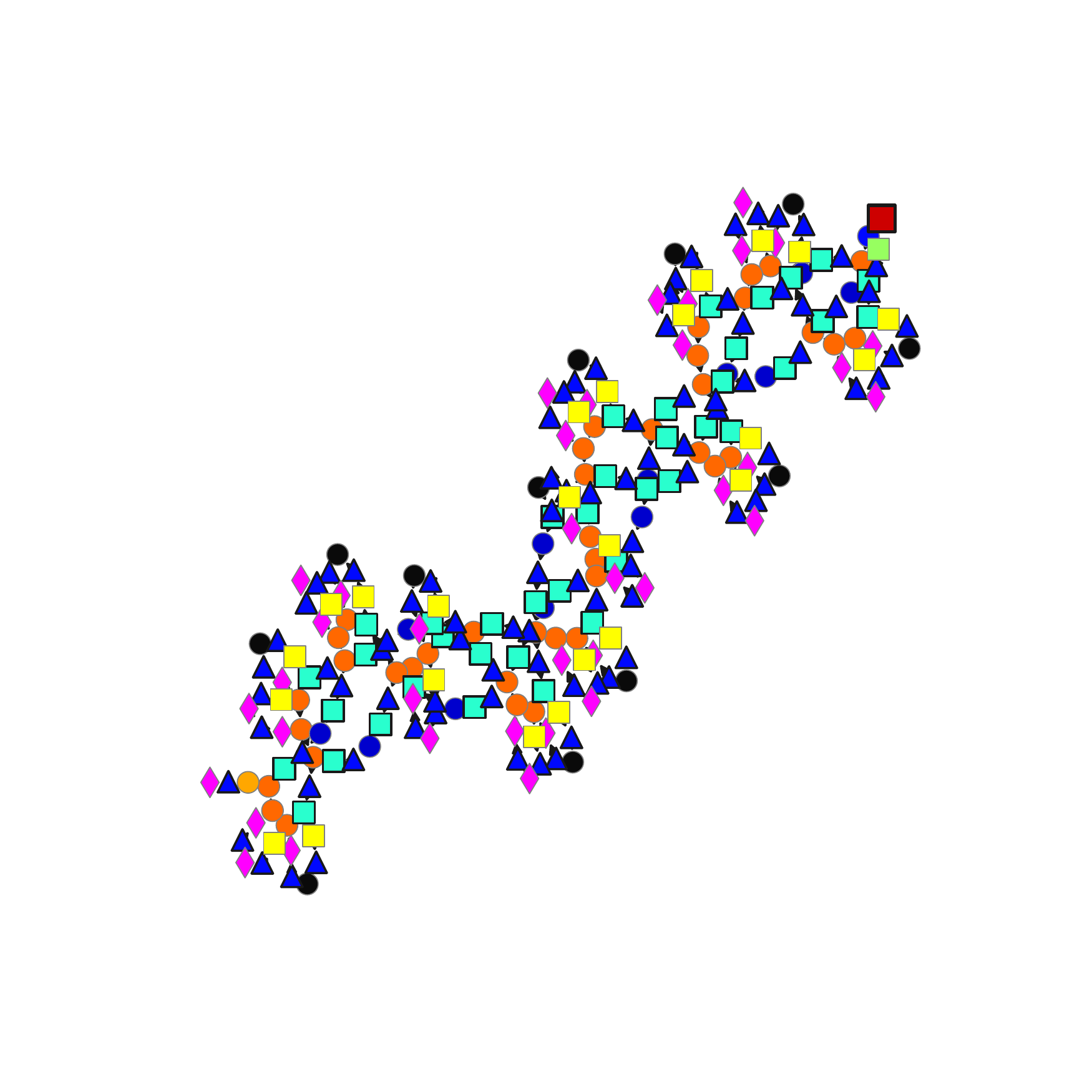}} & {\includegraphics[width=0.22\textwidth,align=c,trim={2.3cm 3cm 2.3cm 3cm},clip]{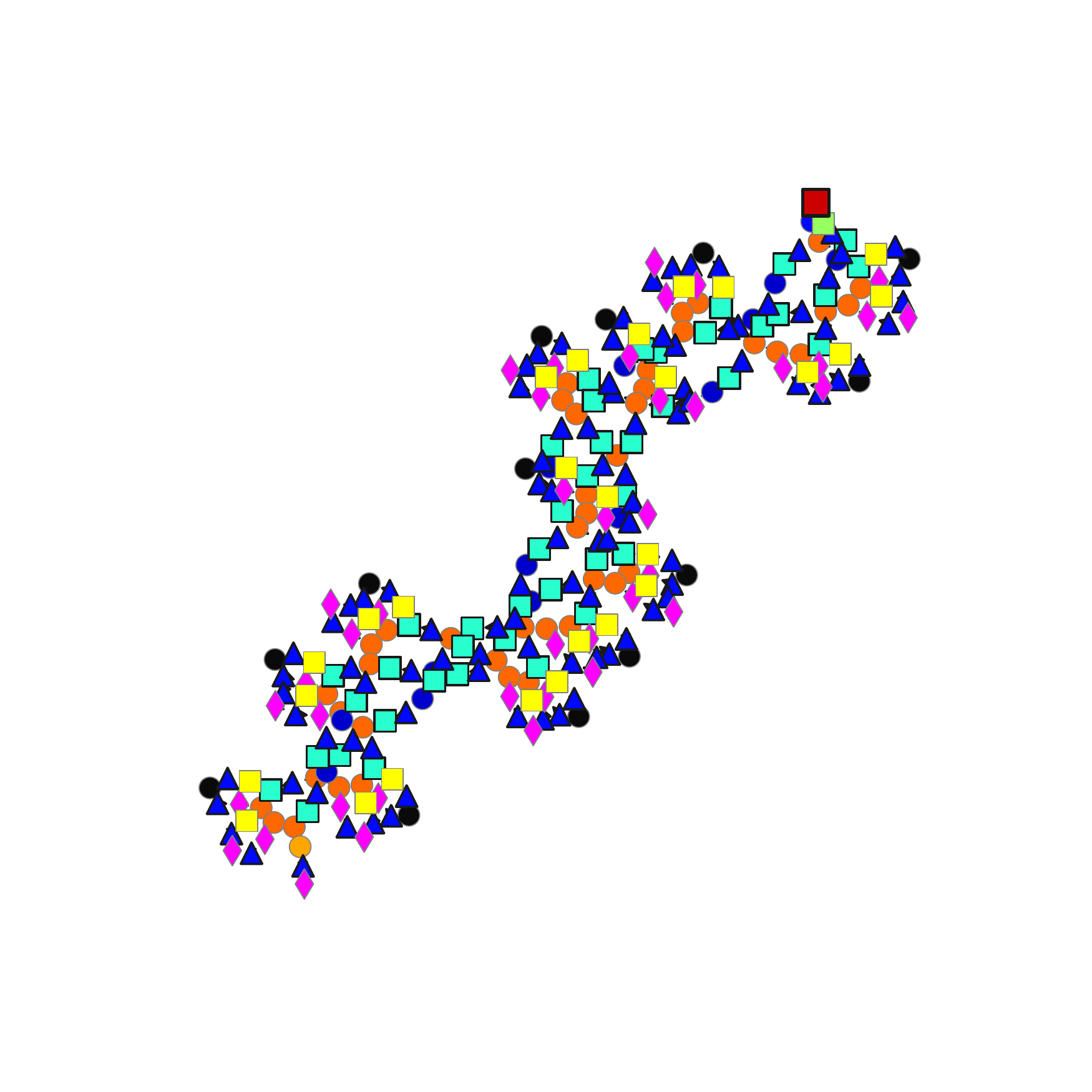}} & 
		{\includegraphics[width=0.22\textwidth,align=c,trim={2.3cm 3cm 2.3cm 3cm},clip]{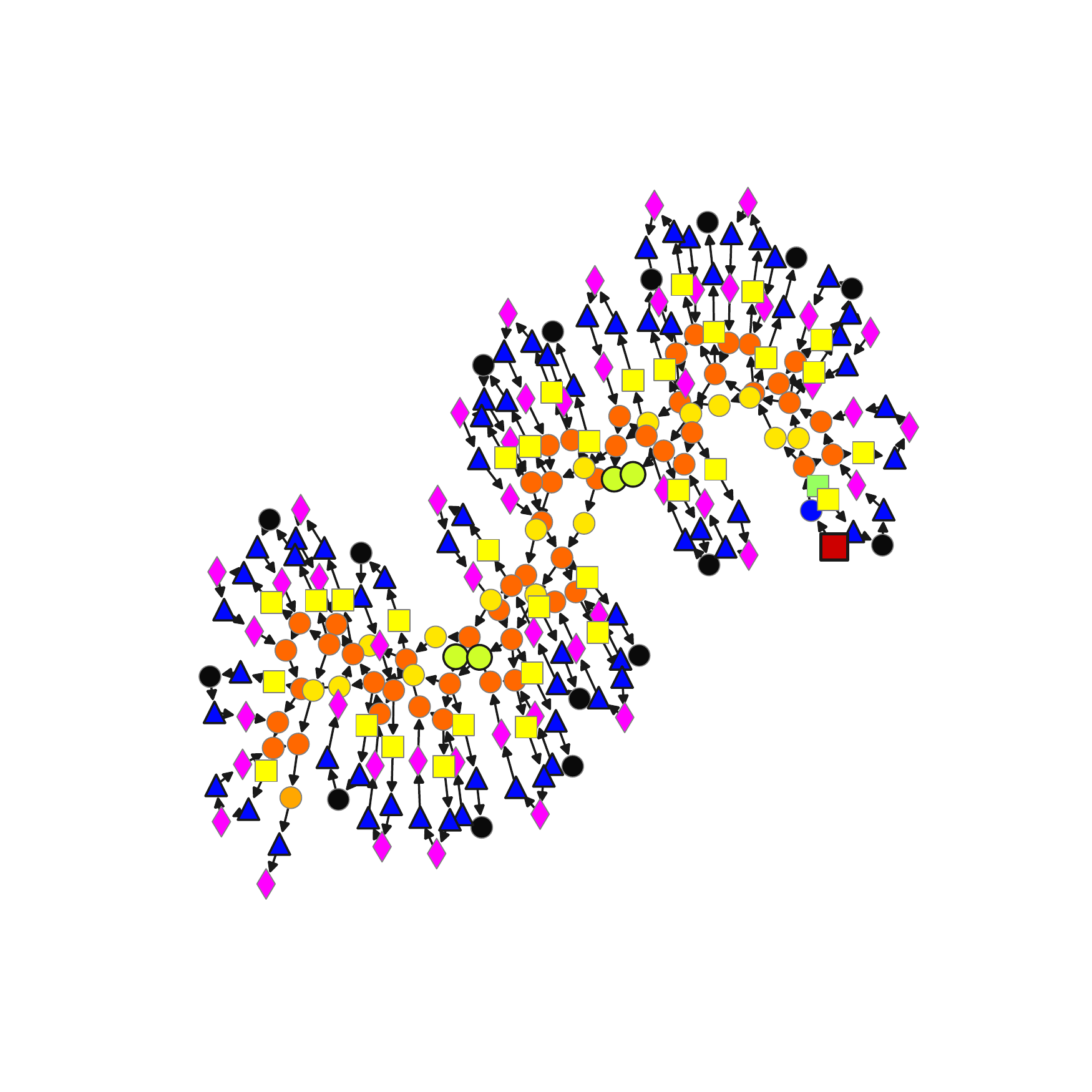}} & {\includegraphics[width=0.22\textwidth,align=c,trim={2.3cm 3cm 2.3cm 3cm},clip]{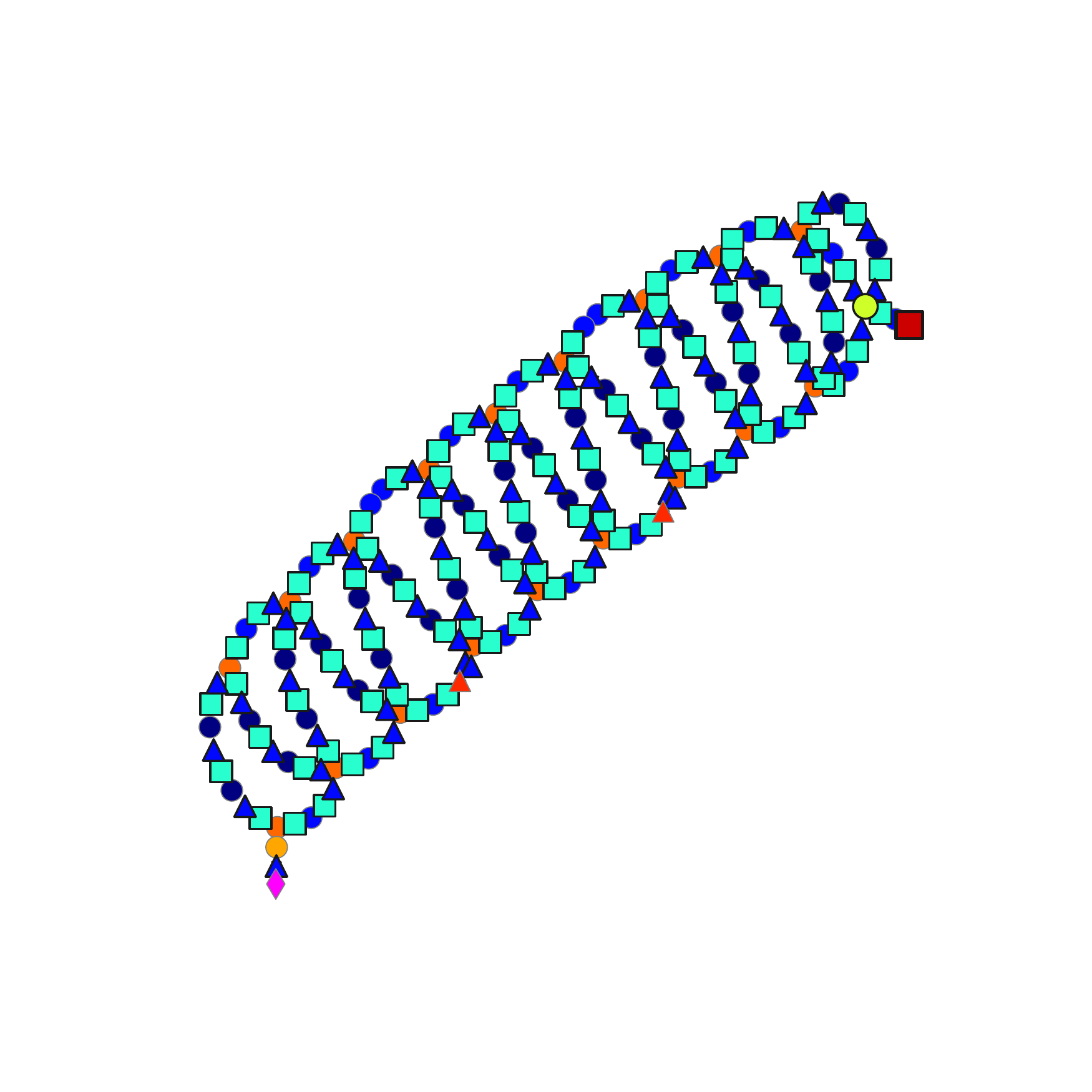}} \\
		\multicolumn{2}{c}{$\ell_2=0.08$} & \multicolumn{2}{c}{$\ell_2=3.6$}
	\end{tabular}
	\caption{Most similar (left) and dissimilar (right) architectures (in terms of the $\ell_2$ distance) in the \iid-\iidtest set based on the graph embeddings obtained by the MLP (top row), \ghnbase (middle row) and \ghnours (bottom row) trained on CIFAR-10.}
	\label{fig:compare}
	\vspace{0pt}
\end{figure}

We further compare the architectures in the \iid-\iidtest set and visualize the most similar and dissimilar ones using our trained models. Based on the visualizations in Fig.~\ref{fig:compare}, MLP as expected is not able to capture graph structures, since it relies only on node features (the most similar architectures shown on the left are quite different). The difference between \ghnbase and \ghnours is hard to understand qualitatively, so we compare them numerically below.

\textbf{Quantitative results.} To quantify the representation power of GHNs, we exploit the fact that, by design, the \ood architectures in our \dataset are more dissimilar from the \iid architectures than the architectures within the \iid distribution. So, a strong GHN should reflect that design principle and map the \ood architectures further from the \iid ones in the embedding space. One way to measure the distance between two feature distributions, such as our graph embeddings, is the Fr\'{e}chet distance (FD)~\cite{dowson1982frechet}. We compute the FD between graph embeddings of 5000 training architectures and five test subsets (in the similar style as in~\cite{liu2019auto2,zilly2019frechet,thompson2020building}): \iid-\iidtest and four \ood subsets (Table~\ref{tab:fd}).
\ghnours maps the \iid-\iidtest architectures close to the training ones, while the \ood ones are further away, reflecting the underlying design characteristics of these subsets. 
On the other hand, both baselines inappropriately map the \deep and \dense architectures as close or even closer to the training ones than \iid-\iidtest, despite the differences in their graph properties (Table~\ref{tab:graphs}, Fig.~\ref{fig:vis_stats}). This indicates \ghnours's stronger representation compared to the \ghnbase and MLP baselines.

\begin{table}[htpb]
	\vspace{0pt}
	\centering
	\definecolor{bad}{rgb}{0.958, 0.788, 0.878}
	\caption{FD between test and training graph embeddings on CIFAR-10. The average assignment cost between two sets of nodes using the Hungarian algorithm is shown for reference.}
	\label{tab:fd}
	\vspace{3pt}
	\small
	\setlength{\tabcolsep}{5pt}
	\begin{tabular}{p{1.2cm}ccccc}
		\toprule
		\textbf{\textsc{Model}} & \textbf{\iid-\iidtest} & \textbf{\wide} & \textbf{\deep} & \textbf{\dense} & \textbf{\bnfree} \\
		\midrule
		Hungarian & 208.7 & 199.9 & 365.6 & 340.7 & 193.1 \\
		MLP & 0.50 & 1.04 & \cellcolor{bad}0.37 & \cellcolor{bad}0.39 & 1.10\Tstrut\\
		\ghnbase & 0.15 & 0.45 & \cellcolor{bad}0.16 & \cellcolor{bad}0.21 & 5.45 \\
		\ghnours & 0.30 & 0.80 & 1.11 & 0.59 & 3.64 \\
		\bottomrule
	\end{tabular}
	\vspace{0pt}
\end{table}

Alternative ways to compare graphs include running expensive graph edit distance (GED) algorithms, infeasible for graphs of our size, or designing task-specific graph kernels~\cite{kriege2020survey,yanardag2015deep,jin2019auto,ma2021deep}. As a more efficient (and less accurate) variant of GED, we employ the Hungarian algorithm. It finds the optimal assignment between two sets of nodes, so its disadvantage is that it ignores the edges. It still can capture the distribution shifts between graphs that can be detected based on the number of nodes and their features, such as for the \deep and \dense subsets (Table~\ref{tab:fd}). It does not differentiate \iidtrain and \bnfree, perhaps, due to the fact that a small portion of architectures in \iidtrain does not have BN. Finally, the inability of the Hungarian algorithm to capture the difference between \iidtrain and \wide can be explained by the fact that we ignore the shape of parameters when running this algorithm.\looseness-1

\subsection{Analysis of Predicted Parameters}

\subsubsection{Diversity}

We analyze how much parameter prediction is sensitive to the input network architecture. For that purpose, we analyze the parameters of 1,000 evaluation architectures (\iid-\iidval and \iid-\iidtest) of \dataset on CIFAR-10. We compare the diversity of the parameters trained with SGD from scratch (He’s initialization+SGD) to the diversity of the parameters predicted by GHNs. To analyze the parameters, we consider all the parameters associated with an operation, which is in general a 4D tensor (i.e. out channels $\PLH$ in channels $\PLH$ height $\PLH$ width). As tensor shapes vary drastically across architectures and layers, it is challenging to find a large set of tensors of the same shape. We found that the shape of 128$\PLH$1$\PLH$3$\PLH$3 is one of the most frequent ones: appearing 760 times across the evaluation architectures. So for a given method of obtaining parameters (i.e. He’s initialization+SGD or GHN), we obtain a set of 760 tensors with 128$\PLH$1$\PLH$3$\PLH$3 = 1152 values in each tensor, i.e. a 760$\PLH$1152 matrix. For each pair of rows in this matrix, we compute the absolute cosine distance, which results in a 760$\PLH$760 matrix with values in range [0,1]. We report the mean value of this matrix in Table~\ref{tab:sensitivity_sparsity}.\looseness-1

\textbf{Results.} The parameters predicted by GHN-2 are more diverse than the ones predicted by GHN-1: average distance is 0.17 compared to 0.07 respectively (Table~\ref{tab:sensitivity_sparsity}). The parameters predicted by MLPs are extremely similar to each other, since the graph structure is not exploited by MLPs. The cosine distance is not exactly zero for MLPs, because two different primitives (group convolution and dilated group convolution) can have the same 128$\PLH$1$\PLH$3$\PLH$3 shape. The parameters predicted by GHN-2 are more similar (low cosine distance) to each other compared to He’s initialization+SGD. Low cosine distances in case of GHNs indicate that GHNs ``store'' good parameters to some degree However, our GHN-2 seems to rely on storing the parameters less than GHN-1 and MLP. Instead, GHN-2 relies more on the input graph to predict more diverse parameters depending on the layer and architecture. So, GHN-2 is more sensitive to the input graph. We believe this is achieved by our enhancements, in particular virtual edges, that allow GHN-2 to better propagate information across nodes of the graph.

\begin{table}[tbhp]
	\caption{Analysis of predicted parameters on CIFAR-10.}
	\label{tab:sensitivity_sparsity}
	\vspace{3pt}
	\small
	\centering
	\begin{tabular}{lcc}
		\toprule
		\textbf{Method of obtaining parameters} & \textbf{Average distance between parameter tensors} &	\textbf{Average sparsity} \\
		\midrule
		He’s init. + SGD &	0.98 & 33\% \\
		MLP	& 0.01	& 16\%\Tstrut\\
		GHN-1 & 0.07 & 20\% \\
		GHN-2 &	0.17 & 39\% \\
		\bottomrule
	\end{tabular}
\end{table}

\subsubsection{Sparsity}

We also analyze the sparsity of predicted parameters using the same 1,000 evaluation architectures. The sparsity can change drastically across the layers, so to fairly compare sparsities we consider the first layer only. We compute the sparsity of parameters $\mathbf{w}$ as the percentage of absolute values satisfying $|\mathbf{w}| < 0.05$. We report the average sparsity of all first-layer parameters in Table~\ref{tab:sensitivity_sparsity}.

\textbf{Results.} The first-layer parameters predicted by GHN-2 are similar to the parameters trained by SGD in terms of sparsity: average sparsity is 39\% compared to 33\% respectively (Table~\ref{tab:sensitivity_sparsity}). Higher sparsity of the parameters predicted by GHN-2 (39\%) compared to the ones of GHN-1 (20\%) may have been achieved due to the proposed parameter normalization method. The parameters predicted by GHN-2 are also more sparse than the ones obtained by SGD. Qualitatively, we found that GHN-2 predicts many values close to 0 in case of convolutional kernels 5$\PLH$5 and larger, which is probably due to the bias towards more frequent 3$\PLH$3 and 1$\PLH$1 shapes during training. Mitigating this bias may improve GHN’s performance.

\subsection{Training Speed of GHNs}
Training GHN-2 with meta-batching takes 0.9 GPU hours per epoch on CIFAR-10 and around 7.4 GPU hours per epoch on ImageNet (Table~\ref{tab:cost}). The training speed is mostly affected by the meta-batch size ($b_m$) and the sequential nature of the GatedGNN. Given that GHNs learn to model 1M architectures, the training is very efficient. The speed of training GHNs with  can be further improved by better parallelization of the meta-batch. Note that GHNs need to be trained only once for a given image dataset. Afterwards, trained GHNs can be used to predict parameters for many arbitrary architectures in less than a second per architecture (see Table~\ref{tab:bench_imagenet} in the main text).

\begin{table}[tbhp]
	\caption{Times of training GHNs on NVIDIA V100-32GB using our code.}
	\label{tab:cost}
	\vspace{3pt}
	\small
	\centering
	\begin{tabular}{lccccc}
		\toprule
		\textbf{\textsc{Model}} & \textbf{\# GPUs} & \multicolumn{2}{c}{\textbf{\textsc{Cifar-10}}} &	\multicolumn{2}{c}{\textbf{\textsc{ImageNet}}} \\
		& & \multicolumn{2}{c}{64 images/batch} &	\multicolumn{2}{c}{256 images/batch}\Bstrut\\
		\midrule
		& & sec/batch & hours/epoch & sec/batch & hours/epoch \\
		Training a single ResNet-50 with SGD & 1 & 0.10  & 0.02 & 0.77 & 1.03\Bstrut\\
		MLP with meta-batch size $b_m=1$ & 1 & 0.32 & 0.06 & 0.67 & 0.90 \\
		\ghnours with meta-batch size $b_m=1$ &	1 & 1.16 & 0.23 & 1.54 & 2.06 \\
		\ghnours with meta-batch size $b_m=8$ & 1 & 7.17 & 1.40 & \multicolumn{2}{c}{out of GPU memory} \\
		\ghnours with meta-batch size $b_m=8$ & 4 &	4.62 & 0.90 & 5.53 & 7.40 \\
		\bottomrule
	\end{tabular}
\end{table}

\section{Additional Related Work\label{apdx:related_work}}

Besides the works discussed in \S~\ref{sec:related}, our work is also loosely related to other parameter prediction methods~\cite{Denil2013-la,bertinetto2016learning,ratzlaff2019hypergan}, analysis of graph structure of neural networks~\cite{you2020graph}, knowledge distillation from multiple teachers~\cite{liu2019knowledge}, compression methods~\cite{cheng2017survey} and optimization-based initialization~\cite{dauphin2019metainit,zhu2021gradinit,das2021data}. \citet{Denil2013-la} train a model that can predict a fraction of network parameters given other parameters requiring to retrain the model for each new architecture.
\citet{bertinetto2016learning} train a model that predicts parameters given a new few-shot task similarly to~\cite{ravi2016optimization,requeima2019fast}, and the model is also tied to a particular architecture.
The HyperGAN~\cite{ratzlaff2019hypergan} allows to generate an ensemble of trained parameters in a computationally efficient way, but as the aforementioned works is constrained to a particular architecture.
Finally, MetaInit~\cite{dauphin2019metainit}, GradInit~\cite{zhu2021gradinit} and Sylvester-based initialization~\cite{das2021data} can initialize arbitrary networks by carefully optimizing their initial parameters, but due to the optimization loop they are generally more computationally expensive compared to predicting parameters using GHNs.
Overall, these prior works did not formulate the task nor proposed the methods of predicting performant \params for diverse and large-scale architectures as ours.\looseness-1

Finally, the construction of our \dataset is related to the works on network design spaces. Generating arbitrary architectures using a graph generative model, e.g.~\cite{yu2019dag,guo2020systematic,you2020graph}, can be one way to create the training dataset $\nets$. Instead, we leverage and extend an existing DARTS framework~\cite{liu2018darts} specializing on neural architectures to generate $\nets$. 
More recent works~\cite{radosavovic2020designing} or other domains~\cite{you2020design} can be considered in future work.

\section{Limitations\label{sec:limit}}

Our work makes a significant step towards reducing the computational burden of iterative optimization methods. However, it is limited in several aspects.

\textbf{Fixed dataset and objective.} Compared to GHNs, one of the crucial advantages of iterative optimizers such as SGD is that they can be easily used to train on new datasets and new loss functions. Future work can thus focus on fine-tuning GHNs on new datasets and objectives or conditioning GHNs on data and hyperparameters akin to SGD.

\textbf{Training speed.} Training \ghnours on larger datasets and with a larger meta-batch ($b_m$) becomes slower. For example, on ImageNet with $b_m=8$ it takes about 7.4 hours per epoch using 4$\PLH$NVIDIA V100-32GB using our PyTorch implementation. So, training of our GHN-2 on ImageNet for 150 epochs took about 50 days. The slow training speed is mainly due to limited parallelization of the meta-batch (i.e. using $b_m$ GPUs should be faster) and the sequential nature of the GatedGNN. \looseness-1

\begin{figure}[htbp]
	\centering
	\includegraphics[width=0.5\textwidth,trim={0.5cm 0 0 0},clip,align=c]{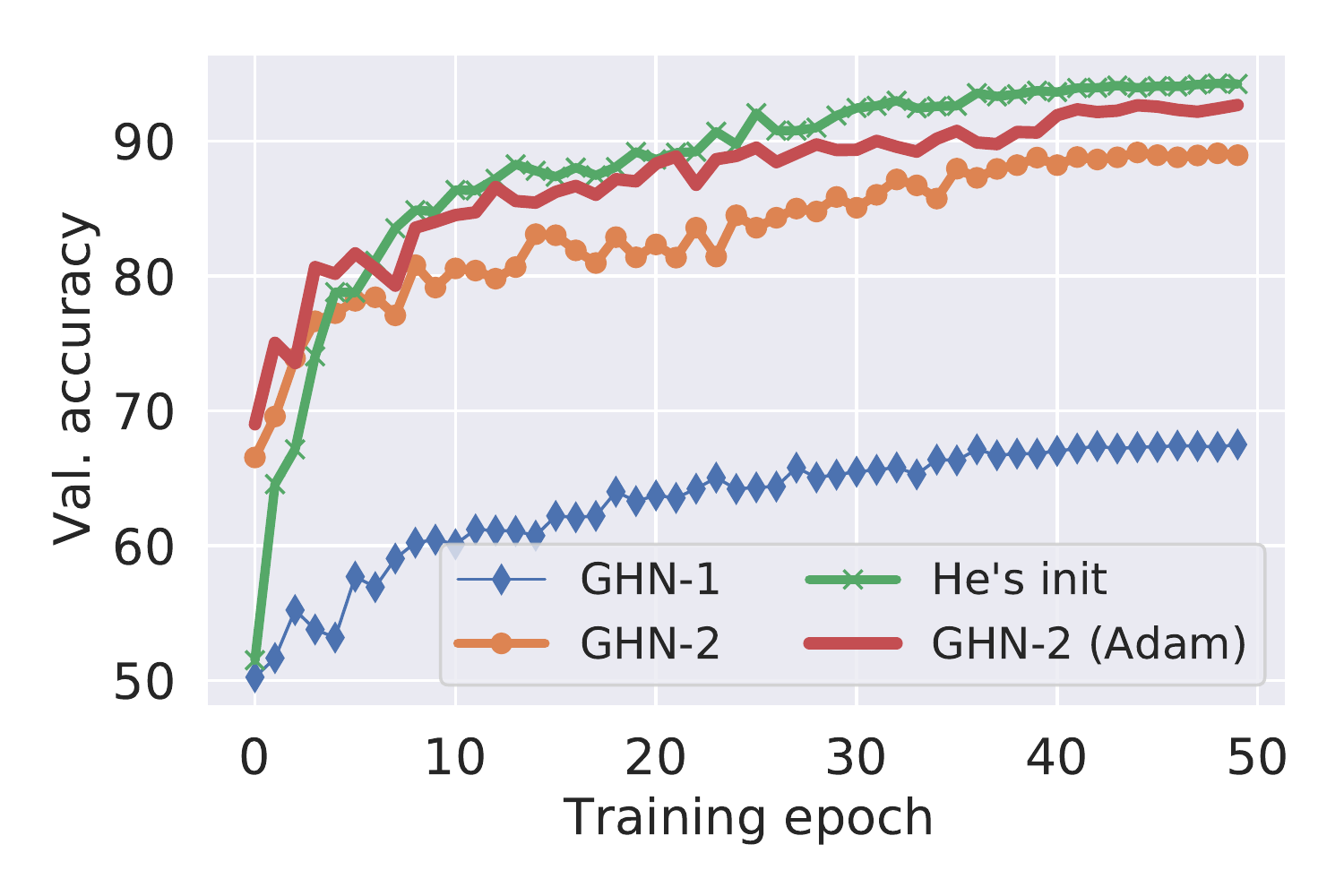}
	\vspace{-5pt}
	\caption{Training ResNet-50 on CIFAR-10 from random and GHN-based initializations.
	}
	\label{fig:fine_tune}
	\vspace{-5pt}
\end{figure}

\textbf{Fine-tuning predicted \params.}
While in \S~\ref{sec:finetune} we showed significant gains of using GHN-2 for initialization on two low-data tasks, we did not find such an initialization beneficial in the case of more data. 
In particular, we compare training curves of ResNet-50 on CIFAR-10 when starting from the predicted parameters versus from the random-based He's initialization~\cite{he2015delving}.
For each initialization case, we tune hyperparameters such as an initial learning rate and weight decay.
Despite ResNet-50 being an OOD architecture, \ghnours-based initialization helps the network to learn faster in the first few epochs compared to He's initialization
(Fig.~\ref{fig:fine_tune}). 
However, after around 5 epochs, He's initialization starts to outperform \ghnours, diminishing its initial benefit. By fine-tuning the predicted parameters with Adam we could slightly improve \ghnours's results. However, the results are still worse than He's initialization in this experiment, which requires further investigation.
Despite this limitation, \ghnours significantly improves on \ghnbase in this experiment similarly to \S~\ref{sec:finetune}.\looseness-1

\section{Societal Impact\label{sec:society}}
One of the negative impacts of training networks using iterative optimization methods is the environmental footprint~\cite{strubell2019energy,cai2019onceforall}. Aiming at improving on the state-of-the-art, practitioners often spend tremendous computational resources, e.g.~for manual/automatic hyperparameter and architecture search requiring rerunning the optimization.
Our work can positively impact society by making a step toward predicting performant parameters in a resource-sustainable fashion. 
On the other hand, compared to random initialization strategies, in parameter prediction a single deterministic model could be used to initialize thousands or more downstream networks. All predicted parameters can 
inherit the same bias or expose any offensive or personally identifiable content present in a source dataset.\looseness-1

% For some reason Latex adds an empty page at the end that I couldn't remove in any easy way
% So I prevent compiling the last page manually based on 
% https://tex.stackexchange.com/questions/96256/compiling-only-a-page-range-or-page-selection
\newcommand{\discardpages}[1]{% \discardpages{<csv list>}
	\xdef\discard@pages{#1}% Store pages to discard
	\AtBeginShipout{% At shipout, decide whether to discard page/not
		\renewcommand*{\do}[1]{% How to handle each page entry in csv list
			\ifnum\value{page}=##1\relax%
			\AtBeginShipoutDiscard% Discard page/not
			\gdef\do####1{}% Do nothing further
			\fi%
		}%
		\expandafter\docsvlist\expandafter{\discard@pages}% Process list of pages to discard
	}%
}
\discardpages{31}

%\vfill
%\clearpage
	
\end{document}